\documentclass[11pt]{article}
\usepackage{b3}

\usepackage{amsmath,amsfonts,bm}

\def\eqref#1{equation~\ref{#1}}

\def\1{\bm{1}}

\DeclareMathAlphabet{\mathsfit}{\encodingdefault}{\sfdefault}{m}{sl}
\SetMathAlphabet{\mathsfit}{bold}{\encodingdefault}{\sfdefault}{bx}{n}

\newif\ifDRAFT
\DRAFTfalse  

\ifDRAFT
  \newcommand{\note}[1]{{\color{blue}\small[#1]}}
  \newcommand{\todo}[1]{{\color{red}\textbf{[TODO: #1]}}}
  \newcommand{\num}[1]{{\color{orange}#1}}
  
\else
  \newcommand{\note}[1]{}
  \newcommand{\todo}[1]{}
  \newcommand{\num}[1]{#1}
  
\fi

\newcommand{\method}{\textsc{Doppel}}
\newcommand{\methodthree}{\textsc{Doppel-3}} 
\newcommand{\methodeight}{\textsc{Doppel-8}} 

\newcommand{\droid}{\textsc{Droid}}
\newcommand{\bridge}{\textsc{Bridge}}
\newcommand{\robomind}{\textsc{RoboMind}}

\newcommand{\obs}{\mathbf{o}}
\newcommand{\act}{\mathbf{a}}
\newcommand{\rend}{\mathbf{r}}
\newcommand{\lat}{\mathbf{z}}
\newcommand{\noise}{\boldsymbol{\epsilon}}

\usepackage{multirow}
\usepackage{subcaption}
\usepackage{tabularx}
\usepackage{array}
\usepackage{placeins}
\usepackage{rotating}

\title{Render the Robot, Generate the World: \\One Video World Model Across Embodiments}

\author{%
  \normalsize
  Yash Jangir$^{1}$ \quad Karan Mirakhor$^{2}$ \quad Renos Zabounidis$^{2}$ \quad Muhammad Zubair Irshad$^{3}$\\[0.3em]
  Sergey Zakharov$^{3}$ \quad Yonatan Bisk$^{2}$ \quad Homanga Bharadhwaj$^{1}$}
\jhuinstitution{%
  $^{1}$Johns Hopkins University \quad $^{2}$Carnegie Mellon University \quad $^{3}$Toyota Research Institute}
\jhucontact{}
\jhulogos{%
  \centering
  \raisebox{-0.5\height}{\includegraphics[height=0.40in,trim=92.9 90.9 92.7 90.7,clip]{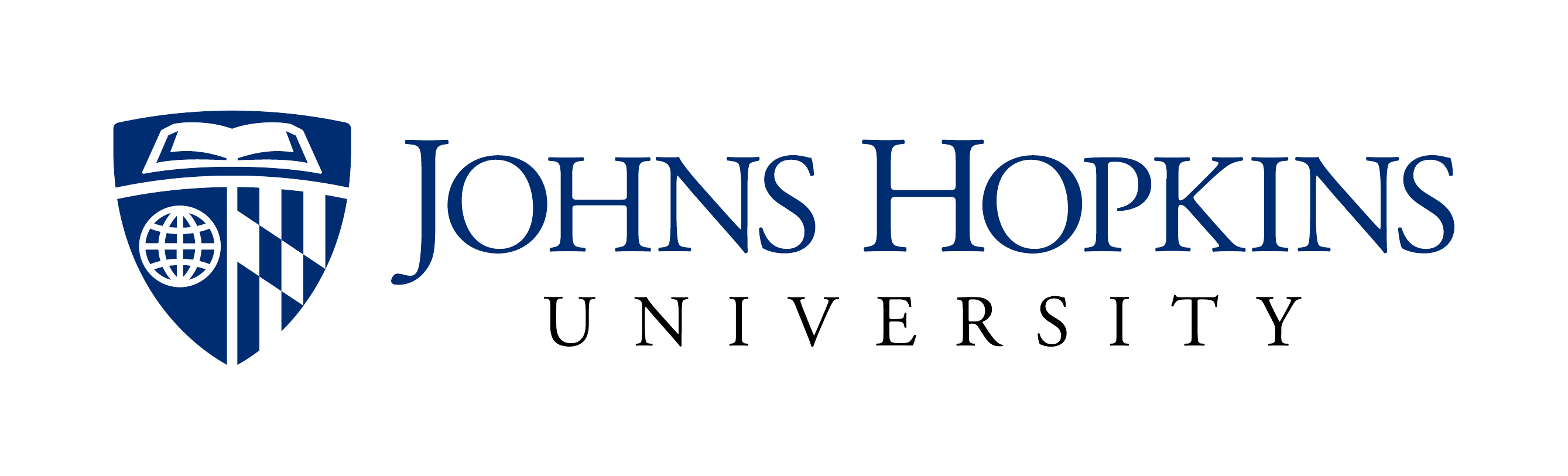}}%
  \hspace{0.32in}{\color{black!18}\rule[-0.22in]{0.6pt}{0.44in}}\hspace{0.32in}%
  \raisebox{-0.5\height}{\includegraphics[height=0.40in,trim=0 47.5 2.5 65.2,clip]{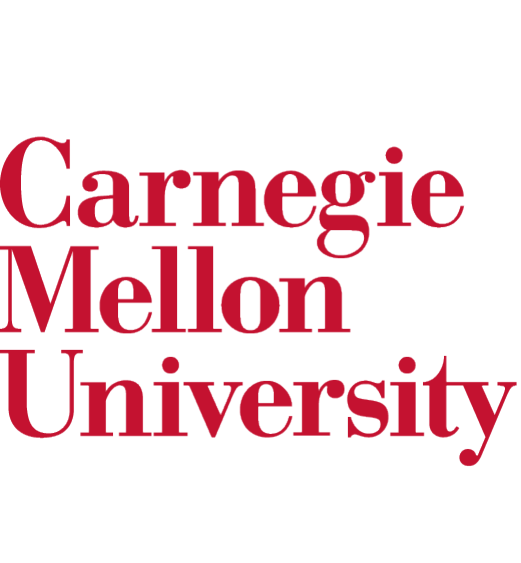}}%
  \hspace{0.32in}{\color{black!18}\rule[-0.22in]{0.6pt}{0.44in}}\hspace{0.32in}%
  \raisebox{-0.5\height}{\includegraphics[height=0.33in]{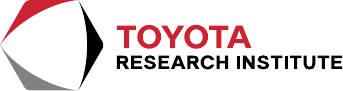}}%
}
\jhuvenue{JHU / CMU / TRI}
\jhurunningtitle{Doppel}
\date{}

\hypersetup{
  pdftitle={Render the Robot, Generate the World: One Video World Model Across Embodiments},
  pdfauthor={Yash Jangir, Karan Mirakhor, Renos Zabounidis, Muhammad Zubair Irshad, Sergey Zakharov, Yonatan Bisk, Homanga Bharadhwaj}
}

\begin{document}

\maketitle

\begin{abstract}

Predicting the consequences of an agent's actions requires modeling both the embodiment's motion and the environment's response. Most video world models attempt to learn both jointly using only numerical action vectors, even though robot geometry and kinematics already provide information needed to specify the intended robot motion. A general-purpose world model should capture these dynamics across multiple embodiments within a single model. Our key idea is to render the robot along a proposed joint-space trajectory and condition the video model directly on these future robot configurations. To this end, we introduce \method{}, an embodiment-agnostic video world model for learning a dynamics model shared across embodiments. \method{} jointly conditions on camera-aligned robot renderings and numerical action trajectories. Using forward kinematics and differentiable rendering, we explicitly provide the model with the robot's intended motion. This allows the shared model to focus its capacity on object motion and interaction-driven changes in the environment. Representing robot motion in image space also provides a common conditioning interface across embodiments with different geometries and action spaces. To enable this approach at scale, we recover missing camera extrinsics from offline robot datasets and align the resulting renderings with recorded observations. Using these aligned renderings, we train a single shared video backbone and conditioning adapter jointly on DROID, RoboMIND, Bridge, ABC, T-Rex, GigaHands, OakInk2, and Ego-Exo4D, spanning Franka, UR5e, WidowX, and bimanual YAM arms, an egocentric humanoid with dexterous hands, and human hands, upper bodies, and full bodies, from about 600 hours of video and without an embodiment label. On the three robot-arm datasets, \method{} improves PSNR over action-only baselines by 1.00 dB on unseen trajectories and 1.20 dB under OOD evaluation while predicting more consistent object motion, and training on all eight embodiments (\methodeight{}) improves robot predictions over the three-robot model (\methodthree{}) on four of six robot evaluation sets.
Project page: \url{https://doppelwm.github.io/}.

\end{abstract}

\begin{figure}[t]
  \centering
  \includegraphics[width=\textwidth]{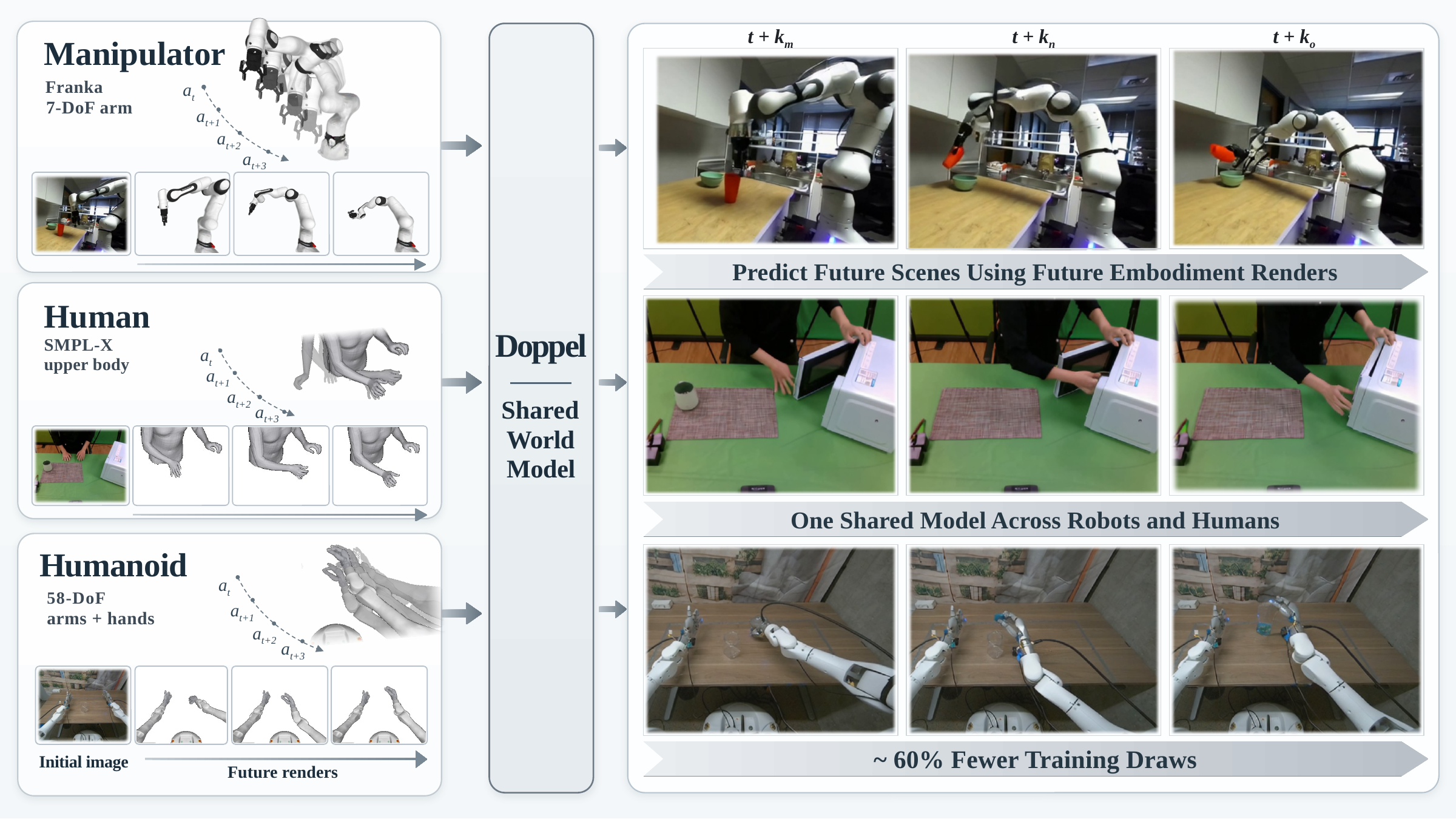}
  \caption{\method{} is a surprisingly simple approach for incorporating future embodiment renderings and improving robot-object interactions in action-conditioned video world models.}
  \label{fig:overview}
\end{figure}
\vspace{-0.2em}

\section{Introduction}
\label{sec:intro}

\begin{flushright}
\vspace*{0.2cm}
\small
\textit{``Civilization advances by extending the number of important operations\\
which we can perform without thinking about them.''}\\ \vspace*{0.2cm}
Alfred North Whitehead, \textit{An Introduction to Mathematics} (1911)
\vspace*{-0.2cm}
\end{flushright}

When we reach for a cup, we have a sense of where our arm is and can anticipate how it will move. Proprioceptive feedback and internal predictions of movement help us estimate the position of our limbs, even when we cannot see them~\citep{wolpert1995internal}. Knowing our own motion, however, does not fully determine what happens to the cup: it may slide, tip over, or stay in our grasp, depending on how we make contact. In everyday physical interactions, we intuitively combine knowledge of our own body with predictions of how the surrounding world will respond. We build on this factorization by using known robot kinematics and camera extrinsics to render the intended robot motion, allowing the world model to focus on modeling the environment's response.

For a robot, its geometry and kinematics provide an explicit description of its body. We study how to incorporate this knowledge in video world models, which \textit{predict future observations from an initial image and a proposed action trajectory.} Recent methods adapt pretrained video generators to this task by conditioning them on numerical action vectors (for example robot joint poses)~\citep{bagchi2026egowm}. The model must then learn both how the actions move the robot in the image and how that motion affects the environment. However, much of the first mapping can already be computed from the robot's geometry, joint configurations, and camera parameters.


Our key idea is to render the robot along the proposed action trajectory and condition the video model on these future frames. We use forward kinematics to compute link poses from joint configurations and the robot URDF, and a learned differentiable rendering model to generate camera-aligned robot renderings from the observed camera viewpoint. The resulting frames show where the arm and gripper would be if the robot followed the provided action trajectory. \textit{In the spirit of Whitehead's observation, we use this known structure to simplify what must be learned}. With the intended robot motion provided explicitly through visual renderings, the video model can focus on predicting its consequences in the environment: whether a grasp would remain stable, how far would a pushed object move, and which way would a cup rotate.


We introduce \textbf{\method{}}, an embodiment-agnostic video world model that uses this conditioning. \method{} combines future robot renderings with the numerical action trajectory. Like a doppelgänger, the rendered robot is a faithful double of the real one; Doppel lets this double show where the robot will move, so the model can focus on how the world responds. Renderings describe the robot's geometry and motion in the image, while the numerical inputs retain information that may be visually ambiguous. Expressing actions this way also enables learning across robots with different joint structures and control spaces. Although Franka, UR5e, and WidowX move differently, their renderings describe the intended motion in a common image space. We therefore train a shared conditioning adapter and video backbone across these embodiments, using robot-specific geometry and normalization only to construct the inputs. This yields a single set of model weights that can represent multiple robot embodiments, avoiding the need to train and maintain a separate world model for each robot.

Realizing this representation across large-scale robot datasets requires accurate alignment between the rendered robot and the observation camera. This alignment depends on the camera pose relative to the robot base, which is often missing from manipulation datasets. We recover these missing extrinsics by optimizing the differentiable renderer to align rendered robot motion with the observations. This allows us to construct camera-aligned conditioning videos across datasets with different robots and camera setups.

Finally, because our formulation explicitly separates the robot's intended motion from its effects on the environment, we evaluate these components separately. We measure prediction quality over robot and moving-object regions, allowing us to distinguish accurate generation of the robot's motion from accurate prediction of the resulting environment dynamics. We train \method{} jointly on \droid{} (Franka), \robomind{} (UR5e), and \bridge{} (WidowX), a model we call \methodthree{}, and evaluate generalization to unseen trajectories and environments. Because the conditioning only needs a rendering of the body, the same model extends beyond robot arms: we also train it on eight embodiments (\methodeight{}), adding a bimanual robot (ABC), a humanoid with dexterous hands (T-Rex), and human hands and bodies (GigaHands, OakInk2, Ego-Exo4D) rendered from fitted body and hand meshes. Our contributions are:

\begin{itemize}[leftmargin=*, labelsep=0.5em, itemsep=2pt, topsep=3pt, parsep=0pt]

\item \textbf{Embodiment-agnostic conditioning.}
We introduce \method{}, a shared video world model conditioned on future robot renderings and numerical actions. This representation describes robot motion in a common image space across embodiments (Sec.~\ref{sec:method}).

\item \textbf{Joint multi-embodiment learning.}
We train one set of weights across three robot arms, outperforming action-only experts, and achieve higher average PSNR and SCS than a shared action-only baseline with 63.8\% fewer training draws. Scaling the same model to eight embodiments (\methodeight{}), including a humanoid and human hands and bodies, with no embodiment label, further improves robot predictions on \droid{} and \robomind{} (Sec.~\ref{sec:results}).

\item \textbf{Interaction-focused evaluation.}
We evaluate robot motion and object dynamics separately using region-specific metrics. Our model improves visual fidelity and motion consistency over action-only and skeleton-conditioned baselines.

\item \textbf{A camera-calibrated corpus.}
We release recovered camera extrinsics for over 45k demonstrations across WidowX, UR5e, and bimanual YAM, along with calibration code for additional datasets and embodiments (Sec.~\ref{sec:method-calib}).

\end{itemize}

\section{Related Work}
\label{sec:related}

\paragraph{Video world models.}
World models predict how an environment evolves under an agent's actions and use these predictions for planning and policy learning~\citep{ha2018worldmodels,hafner2025dreamerv3,wu2022daydreamer}. In robotics, action-conditioned video prediction has benefits in model-predictive control~\citep{finn2017deep,ebert2018visual} and policy improvement~\citep{guo2026vlaw,seo2026stressdream}. Recent large-scale models extend this idea to interactive simulators learned from video~\citep{yang2023unisim,bruce2024genie,agarwal2025cosmos}. Since pretrained video generators~\citep{blattmann2023svd,yang2025cogvideox,wan2025} provide strong visual priors, they can be adapted into action-conditioned world models~\citep{huang2025vid2world}, used to synthesize demonstrations from which actions are inferred for policy learning~\citep{du2023unipi,jang2025dreamgen}, or used as learned simulators for policy evaluation~\citep{quevedo2025worldgym,guo2025ctrlworld}. 

Most robot world models condition on numerical actions per frame~\citep{zhu2024irasim}, with multi-view memory~\citep{guo2025ctrlworld}, or through timestep modulation~\citep{bagchi2026egowm}; others jointly model video and actions~\citep{zhu2025uwm}. These models must learn how actions translate into image-space robot motion, although kinematics and camera geometry largely determine this mapping. Recent work introduces geometry through skeletons~\citep{oscar2026}, 3D robot point flow~\citep{huang2026pointworld}, occupancy-derived visual priors~\citep{yang2026orv}, depth- and normal-aware 4D representations~\citep{zhen2025tesseract}, embodiment masks injected through a ControlNet branch~\citep{chen2026bridgev2w}, projected end-effector pose maps with delta-action embeddings~\citep{jiang2025enerverseac}, robot pointmaps with joint RGB and pointmap generation~\citep{xu2026kinema4d}, egocentric hand-mesh and static-scene renderings~\citep{kim2025dexterous}, and rendered robot motion for demonstration synthesis~\citep{ye2025anchordream,ye2026robodream}. Unlike these, we render the full articulated body along the trajectory and pair it with numerical actions; ORV~\citep{yang2026orv} and EnerVerse-AC~\citep{jiang2025enerverseac} also keep numerical actions, but alongside occupancy renderings or end-effector maps rather than the body itself. Render tokens attend to action tokens, and the result modulates every transformer block per token, in one model shared across robots, humans, and a humanoid without an embodiment label. We can use joint-space trajectories from a policy's action chunks or an off-the-shelf planner.

\paragraph{Multi-embodiment learning.}

Pooling data across robots lets policies share experience across embodiments~\citep{oxe2023,octo2024,kim2024openvla,black2024pi0}, typically through embodiment-specific tokenizers or action heads~\citep{wang2024hpt,doshi2024crossformer}. Other work closes the embodiment gap in image space by swapping one embodiment for another, at test time or in the training data~\citep{chen2024mirage,chen2024roviaug,lepert2025phantom}, or by replacing both with a shared segmentation mask~\citep{lepert2025shadow}. Motion can also be expressed visually, as trajectory sketches~\citep{gu2023rttrajectory}, object flow~\citep{xu2024im2flow2act}, point trajectories~\citep{wu2024draganything,geng2025motionprompting}, projected skeletons~\citep{wang2025visualactions,oscar2026}, or shared particle representations for dynamics~\citep{he2025cross}.

We bring both joint space and image-space embodiment representations together in video world models. Differentiable rendering~\citep{liu2024differentiable} places each robot's motion in a common camera-aligned space, showing shape, gripper geometry, and self-occlusion, while numerical actions retain joint and gripper details that are ambiguous in the image. This lets us train one backbone and adapter jointly across robots, keeping robot-specific assets and normalization and padding actions to a common size. Many offline datasets lack camera extrinsics, so we recover them before training for joint training across embodiments.

\vspace{-0.2em}
\section{Setting and Preliminaries}
\label{sec:problem}

We learn a single visual forward-dynamics model across robots with different geometries and kinematics. Let \(e \in \mathcal{E}\) denote an embodiment with a \(d_e\)-dimensional configuration space. For each embodiment, we use episodes of time-aligned RGB observations \(\obs_{0:T-1}\) and configuration trajectories \(a^e_{0:T-1}\), where \(a^e_t \in \mathbb{R}^{d_e}\) contains the joint positions and gripper state. We also assume access to a URDF \(\mathcal{U}_e\) describing the robot's geometry and kinematics.

Given an initial image \(\obs_0\) and a planned configuration trajectory \(a^e_{1:T-1}\), we predict future images showing how the robot moves and how the environment responds:
\begin{equation}
    p_{\theta}\!\left(
        \obs_{1:T-1}
        \mid
        \obs_0,\,
        a^e_{1:T-1},\,
        \mathcal{U}_e
    \right),
    \label{eq:problem-formulation}
\end{equation}
where we share \(\theta\) across embodiments and model uncertainty in physical properties and object states not visible in the initial image. Since joint vectors differ in size and meaning across robots, we need a common action representation. We evaluate unseen trajectories in familiar environments and shifts to completely new objects, scenes, and environments.

\begin{figure}[t]
  \centering
  \vspace{-3mm}
  
  \includegraphics[width=0.95\textwidth]{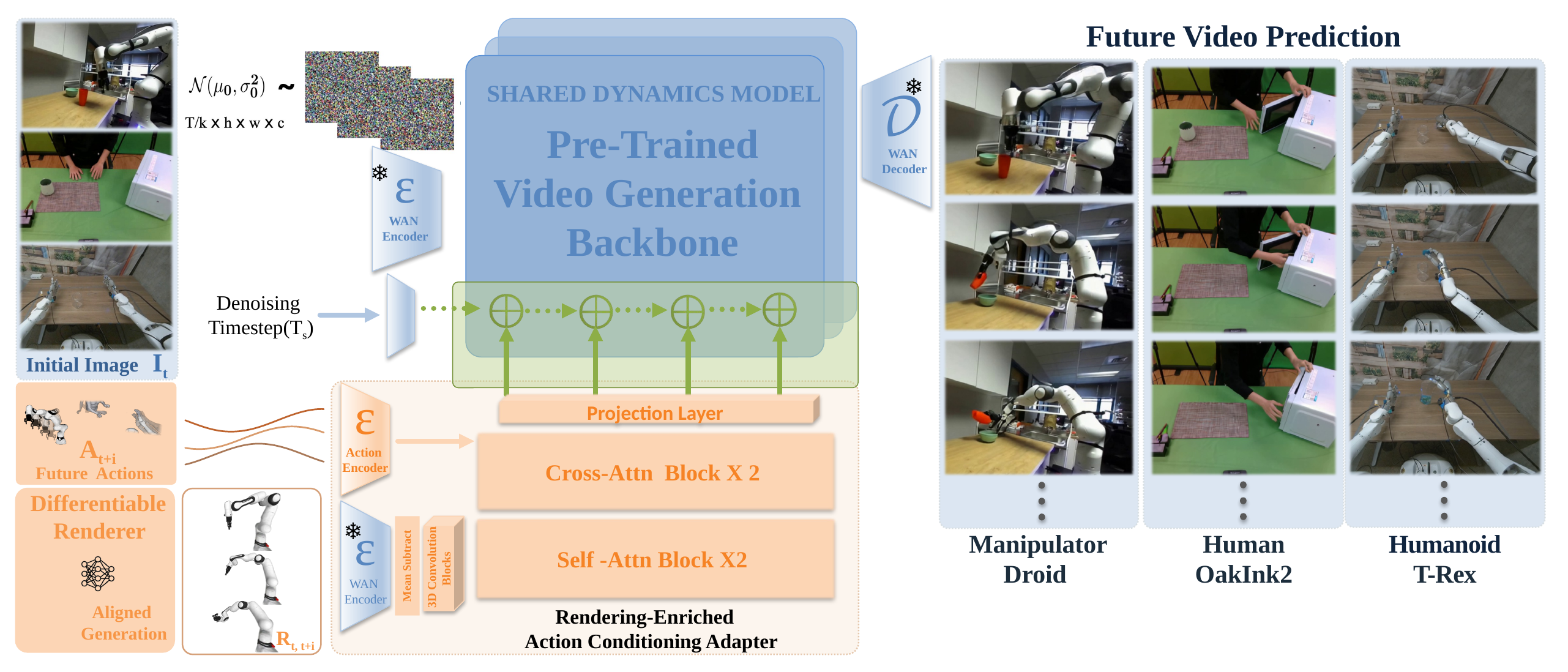}
  \caption{\method{} Architecture. We encode camera-aligned robot renderings with a frozen VAE and fuse their features with numerical actions. We use the resulting representation to modulate the video backbone, which predicts future frames from the initial observation. 
  }
  

  \label{fig:arch}
\end{figure}

\paragraph{Latent video flow matching.}
We build on a pretrained image-to-video rectified-flow transformer~\citep{wan2025}. We use a frozen causal VAE to encode a clip of \(T\) RGB frames into a latent tensor \(\lat \in \mathbb{R}^{C \times T_z \times H_z \times W_z}\), where \(T_z\), \(H_z\), and \(W_z\) are the compressed temporal and spatial dimensions. Given Gaussian noise \(\noise \sim \mathcal{N}(0,I)\), we interpolate between the clean latent and noise at flow timestep \(\tau \in [0,1]\):
\begin{equation}
    \lat_{\tau}
    =
    (1-\tau)\lat+\tau\noise,
    \label{eq:flow-interpolation}
\end{equation}
where \(\tau\) is distinct from video time. We train the model to predict the target velocity \(\noise-\lat\)~\citep{lipman2023flowmatching}.

To condition on the initial observation, we replace the first temporal slice of \(\lat_{\tau}\) with its clean value. We compute the loss over the remaining slices using a mask \(M \in \{0,1\}^{1 \times T_z \times 1 \times 1}\), with \(M_0=0\) and \(M_j=1\) for \(j>0\):
\begin{equation}
    \mathcal{L}_{\mathrm{flow}}
    =
    \mathbb{E}_{\tau,\noise}
    \left[
        \left\|
            M \odot
            \left(
                f_{\theta}(\lat_{\tau},\tau,c)
                -
                (\noise-\lat)
            \right)
        \right\|_2^2
    \right],
    \label{eq:flow}
\end{equation}
where \(c\) is the conditioning provided to the transformer. We pass an empty string to the text encoder throughout training and evaluation. Our predictions therefore depend on the initial image and the robot conditioning described below.

\paragraph{Differentiable robot rendering.}
We use Dr. Robot~\citep{liu2024differentiable} to render the robot at a given configuration. For each embodiment \(e\), we use a set of 3D Gaussians \(\mathcal{G}^0_e\) to represent the robot in a canonical pose. Given a configuration \(a^e_t\), we compute the joint transformations through forward kinematics and deform the Gaussians using an implicit linear blend skinning function \(D_e\). We then apply the pose-conditioned appearance model \(X_e\), which adjusts their orientation, scale, opacity, and spherical-harmonic coefficients:
\begin{equation}
    \mathcal{G}_e(a^e_t)
    =
    X_e\!\left(
        D_e(\mathcal{G}^0_e,a^e_t),
        a^e_t
    \right).
    \label{eq:dr-robot-deformation}
\end{equation}
We render the posed Gaussians with a differentiable rasterizer using the camera intrinsics \(K_i\) and extrinsics \(C_i\):
\begin{equation}
    r^e_t
    =
    \pi_{K_i,C_i}\!\left(
        \mathcal{G}_e(a^e_t)
    \right),
    \label{eq:dr-robot-rendering}
\end{equation}
where \(r^e_t \in \mathbb{R}^{H \times W \times 3}\) is the robot image for episode \(i\) at video time \(t\). We repeat this for the configuration trajectory \(a^e_{0:T-1}\) to obtain a rendering sequence \(r^e_{0:T-1}\) aligned with the observation camera. The renderer is differentiable with respect to both robot configurations and camera parameters. We use its camera gradients to recover missing extrinsics.

\begin{figure}[t]
    \centering
    \vspace{-4mm}
    \includegraphics[width=\linewidth]{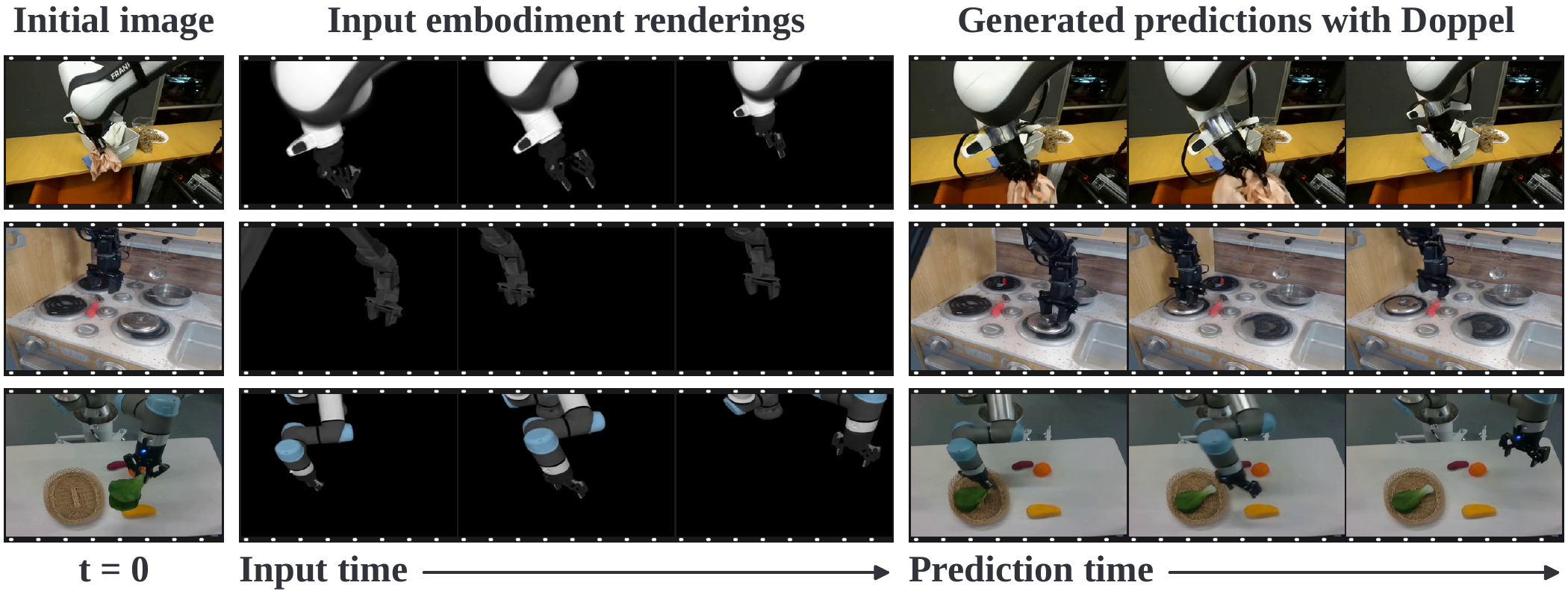}
    \label{fig:ood}
    \vspace{-1em}
    \caption{We show predictions with \method{} on  Franka, WidowX, and UR5e arms in out-of-domain scenes with unseen objects (top to bottom). Each row shows the initial image, future robot renderings used as conditioning, and our predicted frames. We use the same model across all three embodiments.  Our model preserves embodiment-consistent motion while predicting plausible environment responses under substantial distribution shift. }
    \label{fig:ood_results}
\end{figure}

\begin{figure*}[t]
  \centering
  \includegraphics[width=\textwidth]{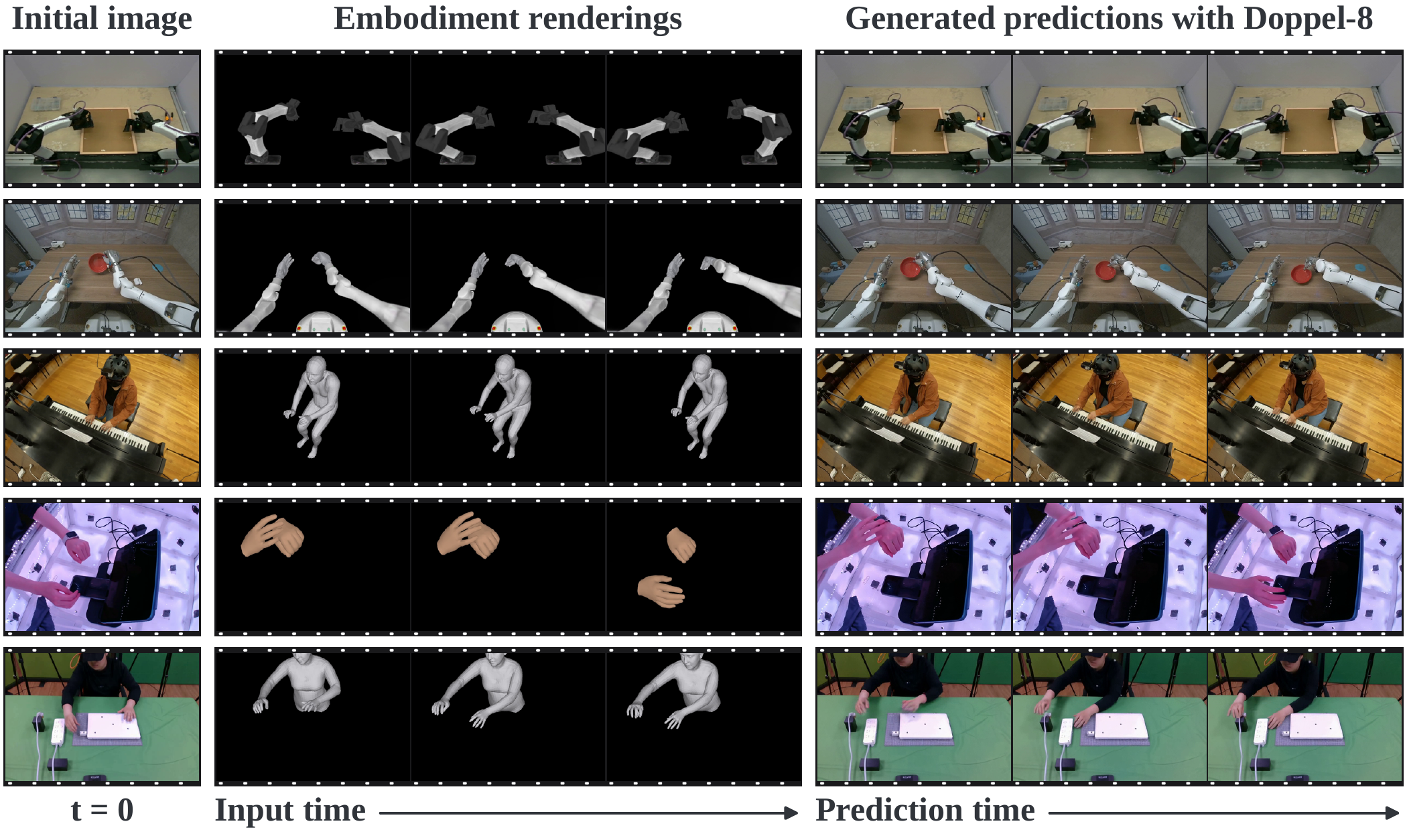}
  \caption{We show predictions with \methodeight{} on held-out test clips from the five added datasets: a bimanual YAM robot (ABC), a humanoid with dexterous hands (T-Rex), a human body (Ego-Exo4D), human hands (GigaHands), and a human upper body (OakInk2), from top to bottom. Each row shows the initial image, future embodiment renderings used as conditioning, and our predicted frames. We use the same model across all five embodiments, with no embodiment label.}
  \label{fig:new-datasets}
\end{figure*}
\vspace{-0.2em}
\section{\method{}}
\label{sec:method}

Our key idea is to provide the robot's intended motion explicitly so the video model can focus on predicting its effects on the environment. We first recover the camera pose relative to the robot base. We then render the robot along the supplied configuration trajectory from this viewpoint. \method{} uses our render-enriched action conditioning adapter to combine these renderings with the numerical joint and gripper trajectories. The renderings show where the robot moves in the image, while the numerical inputs retain configuration information that may be difficult to see. We use this representation to condition \method{}, a single image-to-video model shared across embodiments. 

\subsection{Camera calibration preprocessing}
\label{sec:method-calib}

We need to render the robot from the same viewpoint as the recorded observations. However, the source datasets have missing or inconsistent camera extrinsics. We recover these extrinsics using the calibration procedure of \citet{roboarenainfinity}. We optimize a six-degree-of-freedom camera-to-robot transform by aligning rendered robot trajectories with the recorded video using RGB, optical flow, and DINOv2 features~\citep{oquab2024dinov2}. We keep the robot renderer fixed and optimize only the camera pose using analysis by synthesis approach. 

We calibrate each fixed camera once using a recorded RGB trajectory, the corresponding robot configuration trajectory, the robot's URDF, and the camera intrinsics. We reuse the same differentiable renderer that produces our conditioning videos and recover extrinsics without calibration targets. The recovered camera-to-robot transform is then reused for all demonstrations captured with that camera and remains fixed unless the camera viewpoint changes. We replace the original procedure's optical-flow estimator with RAFT~\citep{teed2020raft} and optimize multiple camera-pose initializations in parallel. We use a coarse pose search for the initialization. After optimization, we select the pose with the lowest alignment loss and keep it fixed while training with \method{}. This procedure removes the need for manual camera calibration by automatically recovering missing camera extrinsics during dataset preprocessing at scale. Qualitative results of the camera calibration are shown in Fig.~\ref{app:fig:camera_overlay}. See App.~\ref{app:camera_extrinsics}.

\subsection{Rendering-Enriched Dense Action Representation}
\label{sec:method-redar}

We combine robot renderings and numerical actions to produce one conditioning vector for each video token. Renderings describe the robot's motion in image space, while numerical actions provide joint and gripper information hidden by occlusion or ambiguous from a single view.

\paragraph{Render and action encoding.}
We encode the rendering sequence \(\rend_{0:T-1}\) with the same frozen causal VAE used for the target video. The encoded renderings have the same latent resolution as the target video. Most of each rendering is empty background, so we center each latent channel by subtracting its mean across frames and spatial locations. We then use a small 3D convolutional encoder to produce features on the backbone's token grid. We add separate positional embeddings for time, height, and width, and use self-attention to let tokens exchange information across the rendering sequence. We normalize the configuration trajectory \(\act^e_{0:T-1}\) using the training statistics for each robot. We pad these vectors so all robots use inputs of the same size, although the joint represented by each coordinate can differ across robots. We apply a shared MLP to each configuration and add a learned embedding for its time index.

\begin{figure}[t]
\centering
\includegraphics[width=0.9\linewidth]{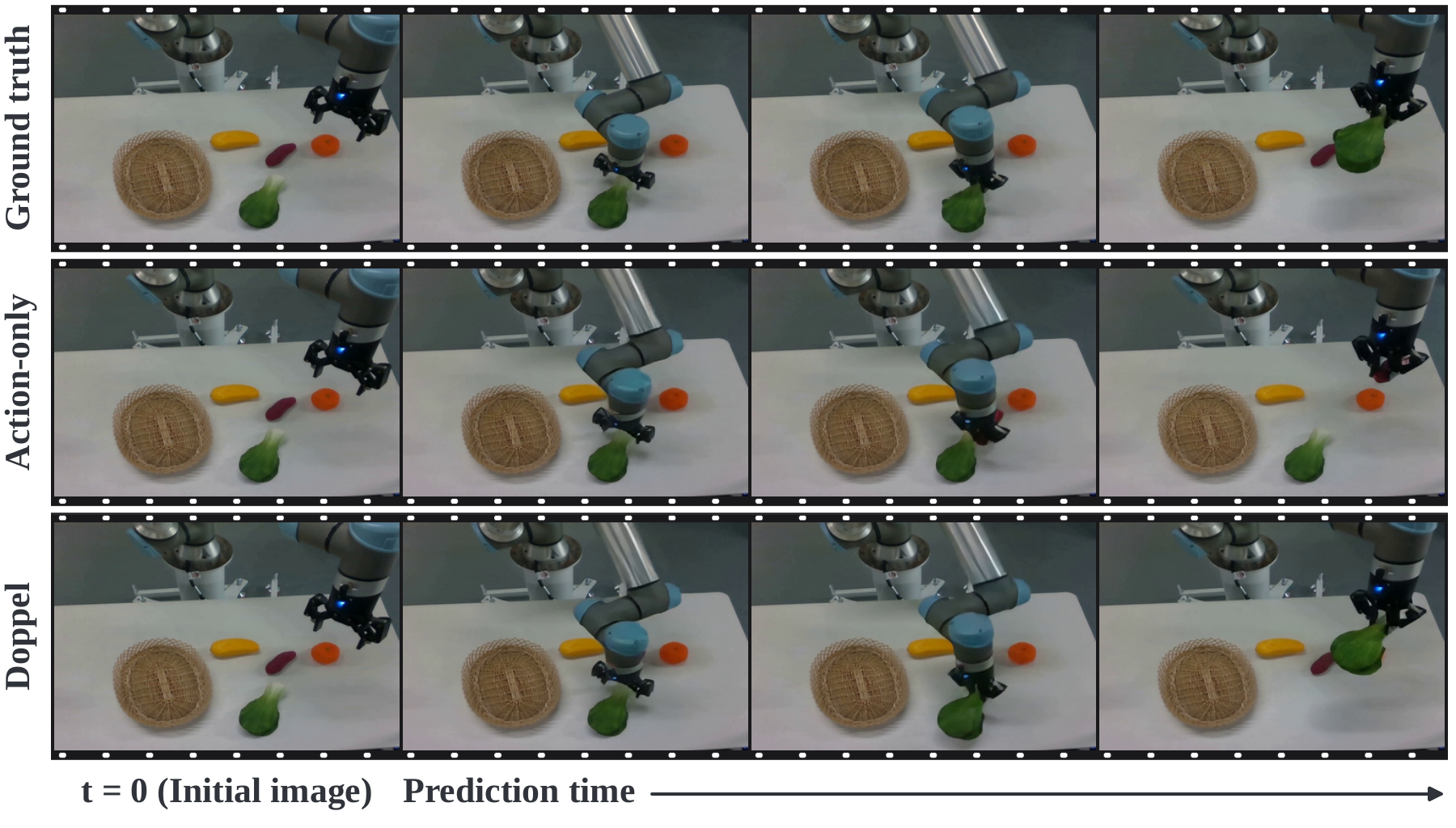}
\vspace{-0.5em}
\caption{
Qualitative comparison across trajectory and OOD generalization settings. Compared to the baseline we observe that \method{} generates much plausible robot-object interactions as intended by the provided action trajectory. Additional qualitative results are available on our project website (Appendix~\ref{app:website}) and a trajectory generalization illustration in Fig. \ref{fig:additional-droid-trajectory}.
}
\label{fig:qualitative_generalization}

\end{figure}

\paragraph{Render-action fusion.}
We let the render tokens attend to the action sequence. In this cross-attention step, render tokens supply the queries and action tokens supply the keys and values. Each output token keeps its location on the video grid and also contains information about the robot's configuration. We denote the complete render--action adapter by \(g_{\phi}\):
\begin{equation}
G
=
g_{\phi}\!\left(
\rend_{0:T-1},
\act^e_{0:T-1}
\right)
\in
\mathbb{R}^{N\times d},
\label{eq:redar-fusion}
\end{equation}
where \(N\) is the number of video tokens and \(d\) is the backbone width. We write the conditioning vector at video token \(j\) as \(g_j\).

\paragraph{Per-token timestep modulation.}
We follow \citet{bagchi2026egowm} and condition the video model through its flow-timestep embedding. We use a separate conditioning vector for each video token so the model receives information about robot motion at different positions and times. For token \(j\), we add \(g_j\) to the embedding \(e(s)\in\mathbb{R}^{d}\) of flow timestep \(s\):
\begin{equation}
\widetilde{e}_j(s)
=
e(s)+g_j.
\label{eq:redar-injection}
\end{equation}
We pass \(\widetilde{e}_j(s)\) through the backbone's existing projection to obtain the shift, scale, and gate parameters for the attention and feed-forward sublayers in every transformer block. We retain the pretrained attention and feed-forward architecture and give the adapter details and tensor dimensions in Appendix~\ref{app:render-action-adapter}.

\paragraph{Sharing across embodiments.}
We train a single render--action adapter and video backbone across all embodiments. Each robot uses its own frozen renderer and action-normalization statistics. We share all trainable parameters and use no embodiment tokens or robot-specific heads. The renderer supplies the geometry and kinematics of each robot, and we use the same world model to learn how its motion changes the scene.

\begin{table}[t]
  \centering
  \caption{
    Training corpus. Validation clips come from held-out trajectories in seen settings; test clips hold out the listed shift. 
  }
\label{tab:data}
  \small
  \setlength{\tabcolsep}{4pt}
  \resizebox{\columnwidth}{!}{%
  \begin{tabular}{@{}llrrrrl@{}}
    \toprule
    Dataset & Embodiment & \shortstack{Train\\episodes} & \shortstack{Train\\clips} & \shortstack{Val.\\clips} & \shortstack{Test\\clips} & Test distribution shift \\
    \midrule
    \droid{} & Franka Panda & \num{16726} & \num{83511} & \num{200} & \num{200} & Unseen lab \\
    \bridge{} & WidowX & \num{12387} & \num{61935} & \num{200} & \num{200} & Unseen kitchen \\
    \robomind{} & UR5e + 2F-85 & \num{14801} & \num{74005} & \num{200} & \num{200} & Unseen objects \\
    \midrule
    ABC & Bimanual YAM & \num{9115} & \num{274523} & \num{200} & \num{200} & Unseen tasks \\
    T-Rex & Vega-1 + Sharpa hands & \num{4784} & \num{58120} & \num{200} & \num{200} & Unseen backgrounds \\
    Ego-Exo4D & Human body (SMPL-H) & \num{2041} & \num{213292} & \num{200} & \num{200} & Unseen tasks \\
    GigaHands & Human hands (MANO) & \num{8491} & \num{31218} & \num{200} & \num{200} & Unseen activities \\
    OakInk2 & Human upper body (SMPL-X) & \num{403} & \num{6998} & \num{200} & \num{200} & Unseen tasks \\
    \midrule
    Three robot datasets & --- & \num{43914} & \num{219451} & \num{600} & \num{600} & --- \\
    All eight datasets & --- & \num{68748} & \num{803602} & \num{1600} & \num{1600} & --- \\
    \bottomrule
  \end{tabular}%
  }
\end{table}

\begin{table*}[t]
\centering
\caption{
Prediction quality on the three robot datasets. \emph{Input} is what the model is conditioned on besides the first frame; \emph{Trained on} is its training data. All our models are evaluated at their final checkpoint and saw a similar amount of each robot dataset during training. 
}
\label{tab:overall-results}
\small
\setlength{\tabcolsep}{4pt}
\resizebox{\textwidth}{!}{%
\begin{tabular}{@{}llllrrrrrrrr@{}}
\toprule
& & & & \multicolumn{4}{c}{Validation: unseen trajectories} & \multicolumn{4}{c}{OOD test} \\
\cmidrule(lr){5-8} \cmidrule(lr){9-12}
Dataset & Method & Input & Trained on & PSNR \(\uparrow\) & SSIM \(\uparrow\) & LPIPS \(\downarrow\) & SCS \(\uparrow\) & PSNR \(\uparrow\) & SSIM \(\uparrow\) & LPIPS \(\downarrow\) & SCS \(\uparrow\) \\
\midrule
\droid{} & Action-only & Actions & This dataset & 22.888 & 0.841 & 0.078 & 0.927 & 20.069 & 0.779 & 0.131 & 0.873 \\
 & OSCAR (fine-tuned) & Skeleton & This dataset & 20.427 & 0.751 & 0.135 & 0.903 & 18.619 & 0.705 & 0.174 & 0.854 \\
 & \method{} & Render + actions & This dataset & \underline{24.357} & \underline{0.864} & \underline{0.065} & \underline{0.940} & \underline{21.276} & 0.802 & 0.113 & \underline{0.902} \\
 & \methodthree{} & Render + actions & 3 robot datasets & 24.297 & 0.862 & 0.066 & \underline{0.940} & 21.250 & \underline{0.803} & \underline{0.111} & 0.899 \\
 & \methodeight{} & Render + actions & All 8 datasets & \textbf{24.819} & \textbf{0.870} & \textbf{0.062} & \textbf{0.943} & \textbf{21.460} & \textbf{0.806} & \textbf{0.108} & \textbf{0.903} \\
\midrule
\bridge{} & Action-only & Actions & This dataset & 26.290 & 0.871 & 0.050 & 0.946 & 20.584 & 0.755 & 0.131 & 0.866 \\
 & OSCAR (fine-tuned) & Skeleton & This dataset & 22.559 & 0.771 & 0.117 & 0.914 & 19.487 & 0.633 & 0.211 & 0.854 \\
 & \method{} & Render + actions & This dataset & \textbf{26.618} & \textbf{0.875} & \textbf{0.048} & \textbf{0.949} & 21.722 & 0.771 & 0.111 & \underline{0.901} \\
 & \methodthree{} & Render + actions & 3 robot datasets & \underline{26.551} & \underline{0.873} & \underline{0.049} & \textbf{0.949} & \textbf{22.474} & \textbf{0.786} & \textbf{0.097} & \textbf{0.905} \\
 & \methodeight{} & Render + actions & All 8 datasets & 26.544 & \underline{0.873} & \underline{0.049} & \underline{0.948} & \underline{22.238} & \underline{0.784} & \underline{0.099} & 0.899 \\
\midrule
\robomind{} & Action-only & Actions & This dataset & 30.474 & 0.923 & 0.038 & 0.940 & 30.098 & 0.918 & \underline{0.039} & 0.932 \\
 & OSCAR (fine-tuned) & Skeleton & This dataset & 27.456 & 0.882 & 0.079 & 0.914 & 27.661 & 0.878 & 0.076 & 0.920 \\
 & \method{} & Render + actions & This dataset & \underline{31.104} & \underline{0.928} & \underline{0.037} & \textbf{0.942} & 30.290 & 0.919 & \underline{0.039} & 0.928 \\
 & \methodthree{} & Render + actions & 3 robot datasets & 31.065 & \underline{0.928} & \underline{0.037} & \underline{0.941} & \underline{30.442} & \underline{0.920} & \textbf{0.038} & \underline{0.935} \\
 & \methodeight{} & Render + actions & All 8 datasets & \textbf{31.302} & \textbf{0.930} & \textbf{0.036} & \textbf{0.942} & \textbf{30.659} & \textbf{0.921} & \textbf{0.038} & \textbf{0.936} \\
\bottomrule
\end{tabular}%
}
\vspace{2pt}
\parbox{\textwidth}{\scriptsize Best in \textbf{bold}, second-best \underline{underlined}; ties share emphasis. 200 clips per split; every method is scored on the same clips.}
\end{table*}

\begin{table}[t]
\centering
\caption{
\methodeight{}, evaluated on the five datasets added beyond the three robot-arm datasets: a bimanual robot (ABC), a humanoid with dexterous hands (T-Rex), human bodies (Ego-Exo4D), human hands (GigaHands), and human upper bodies (OakInk2). These are the same weights as the last row of each block in Table~\ref{tab:overall-results}. Validation holds out trajectories; the test split holds out the shift listed in Table~\ref{tab:data}.
}
\label{tab:other-results}
\small
\setlength{\tabcolsep}{3pt}
\resizebox{\columnwidth}{!}{%
\begin{tabular}{@{}lrrrrrrrr@{}}
\toprule
& \multicolumn{4}{c}{Validation: unseen trajectories} & \multicolumn{4}{c}{Test: held-out shift} \\
\cmidrule(lr){2-5} \cmidrule(lr){6-9}
Dataset & PSNR \(\uparrow\) & SSIM \(\uparrow\) & LPIPS \(\downarrow\) & SCS \(\uparrow\) & PSNR \(\uparrow\) & SSIM \(\uparrow\) & LPIPS \(\downarrow\) & SCS \(\uparrow\) \\
\midrule
ABC & 24.653 & 0.850 & 0.096 & 0.920 & 25.217 & 0.861 & 0.089 & 0.927 \\
T-Rex & 28.327 & 0.871 & 0.050 & 0.961 & 28.346 & 0.867 & 0.051 & 0.956 \\
Ego-Exo4D & 27.935 & 0.895 & 0.045 & 0.966 & 27.611 & 0.892 & 0.044 & 0.965 \\
GigaHands & 25.622 & 0.885 & 0.053 & 0.941 & 25.784 & 0.897 & 0.049 & 0.944 \\
OakInk2 & 29.014 & 0.918 & 0.036 & 0.970 & 29.090 & 0.921 & 0.033 & 0.970 \\
\bottomrule
\end{tabular}%
}
\end{table}

\section{Experimental Setup}
\label{sec:impl}

\paragraph{Datasets and splits.}
We train jointly on \droid{} (Franka), \bridge{} (WidowX), and \robomind{} (UR5e)~\citep{droid2024,walke2023bridgedata,wu2024robomind}. Franka and UR5e use Robotiq 2F-85 grippers. We sample datasets uniformly and balance environment categories, using short contiguous and temporally subsampled full-span clips. We evaluate held-out trajectories in familiar environments and OOD shifts to a new lab (\droid{}), a new kitchen (\bridge{}), and unseen objects and tasks (\robomind{}). We call this model \methodthree{}. For \methodeight{}, we add ABC~\citep{allshire2026abc} (bimanual YAM), T-Rex~\citep{niu2026trex} (Vega-1 humanoid with Sharpa hands), GigaHands~\citep{fu2025gigahands} (human hands), OakInk2~\citep{zhan2024oakink2} (human upper body), and Ego-Exo4D~\citep{grauman2024egoexo4d} (human body). The two new robots are rendered from their kinematic models; people are rendered from fitted MANO~\citep{romero2017mano}, SMPL-X~\citep{pavlakos2019smplx}, and SMPL-H~\citep{romero2017mano} meshes. Their test splits hold out tasks (ABC, OakInk2, Ego-Exo4D), activities (GigaHands), or wall backgrounds (T-Rex). All actions are zero-padded to a common 159-dimensional vector and enter one shared action encoder without an embodiment label, and we sample each dataset in proportion to its training clips. All evaluated embodiments appear in training. We give corpus sizes in Table~\ref{tab:data}, construction details in Appendix~\ref{app:data}, and examples in Appendix~\ref{app:datasetviz}.

\paragraph{Baselines.}
We compare \method{} against both an action-only~\citep{bagchi2026egowm} and geometric-conditioned baselines~\citep{oscar2026}. The action-only baseline follows EgoWM-style~\citep{bagchi2026egowm} timestep-modulated action conditioning using the same Wan~2.1 backbone and corresponding dataset's training corpus. We compare single dataset experts with our multi-embodiment \method{} at approximately matched training exposure per dataset. In separate multi embodiment comparison, both models use same training recipe. OSCAR~\citep{oscar2026} serves as a geometry-conditioned system-level baseline; we fine-tune one model per embodiment on the corresponding training demonstrations using OSCAR's temporal windowing and evaluate on the same splits. Because OSCAR uses a different backbone and training formulation, we retain its native architecture while matching the evaluation resolution and removing text prompts for a fairer comparison. Full training configurations for all baselines are provided in Appendix~\ref{app:baselineimp}.

\paragraph{Evaluation metrics.}
We report whole-frame PSNR, SSIM~\citep{wang2004ssim}, and LPIPS~\citep{zhang2018lpips}, along with object SCS~\citep{bagchi2026egowm}. We compute SCS as the overlap of SAM2 masks propagated from the same initial frame through predicted and reference videos ~\citep{ravi2024sam2}. We also measure PSNR separately over the robot arm, moving objects, and SCS over the arm and moving objects. We exclude the conditioning frame from all metrics and give definitions in Appendix~\ref{app:metrics}.

\paragraph{Implementation details.}
We use 8 H100 GPUs for single-dataset training and 32 GPUs for the multi-dataset runs at a global batch size of 256: \methodthree{} trains for 49.5k steps and \methodeight{} for 189k steps. We give complete training recipes in Appendix Tables~\ref{tab:training-recipe-single}--\ref{tab:training-recipe-emb8}.

\begin{figure}[!htbp]
\centering
\vspace{-3mm}

\includegraphics[width=0.9\linewidth]{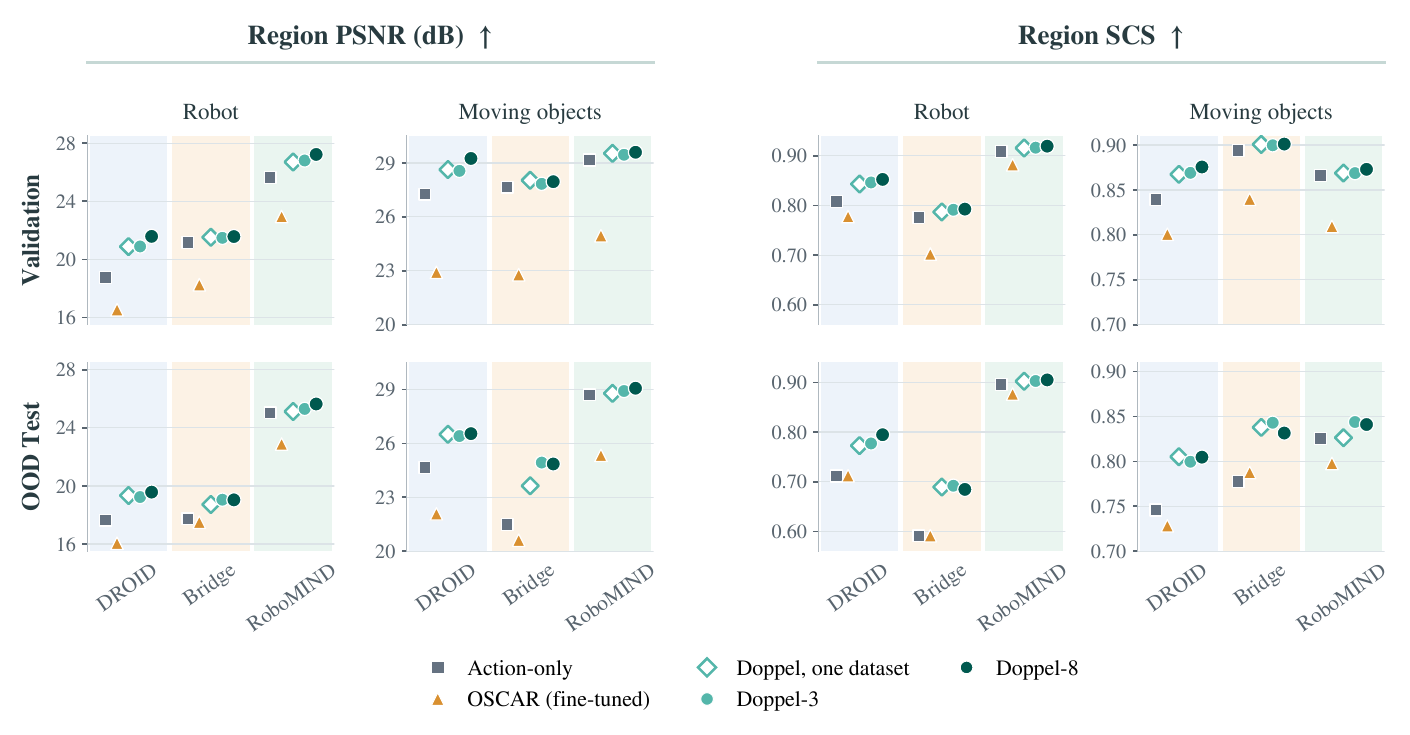}
\caption{Prediction quality by image region (robot and moving objects) on the three robot datasets, for every model at its final checkpoint, on the same clips as Table~\ref{tab:overall-results}. Values are in Appendix Table~\ref{tab:region-results}.
}
\label{fig:region-results}
\end{figure}

\begin{figure}[!htbp]
    \centering
    \includegraphics[width=\linewidth]{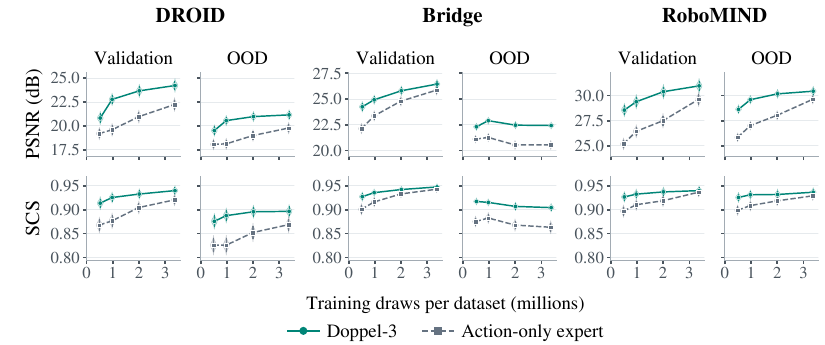}
    
    \caption{Training efficiency of \method{} versus single-dataset action-only experts. \method{} achieves higher validation and OOD prediction quality at comparable training draws.}
    
    \label{fig:trainingeff}
\vspace{-4mm}
    
\end{figure}

\section{Results and Analysis}
\label{sec:results}

\paragraph{Prediction across embodiments.}
We improve whole-frame prediction quality over action-only conditioning and fine-tuned OSCAR~\citep{oscar2026} on all three robot datasets, across validation and OOD splits (Table~\ref{tab:overall-results}). With the same training data, adding the robot rendering to the action input improves PSNR over action-only conditioning by 0.2--1.5\,dB. Our shared models outperform separately trained action-only experts, and \methodeight{} has the best PSNR, SSIM, and LPIPS on \droid{} and \robomind{} in both splits. Compared with \methodthree{} on the same clips, adding the five other datasets raises PSNR by 0.52 and 0.21 dB on \droid{} validation and OOD and by 0.24 and 0.22 dB on \robomind{}, leaves \bridge{} validation unchanged, and lowers \bridge{} OOD by 0.24 dB (Appendix Table~\ref{tab:three-to-eight}). With every model at ${\sim}51$ passes over each robot dataset, \methodeight{} has the highest PSNR on five of the six robot evaluation sets and trails \methodthree{} by 0.05\,dB on \bridge{} OOD (Appendix Table~\ref{tab:matched-51}). On \bridge{} OOD, both multi-dataset models peak before the end of training and then decline while \bridge{} validation keeps improving, which suggests that late training specializes to the seen kitchens. Table~\ref{tab:other-results} reports \methodeight{} on the five added datasets, and Fig.~\ref{fig:new-datasets} shows held-out test examples. We show predictions of \methodthree{} in held-out environments and with unseen objects in Fig.~\ref{fig:ood_results}.

\paragraph{What joint multi-embodiment training contributes.} The five added datasets contain no Franka, UR5e, or WidowX data, yet they improve robot prediction. The model is never told which body it sees: the rendering identifies the embodiment, and the shared action encoder receives zero-padded vectors whose slots mean different joints for different bodies. Overall, \methodeight{} improves PSNR over \methodthree{} on four of the six robot evaluation sets (Table~\ref{tab:overall-results}; Appendix Table~\ref{tab:three-to-eight}), so training on other embodiments helps robot prediction. The exception is \bridge{} OOD, where \methodeight{} is 0.24\,dB below \methodthree{}. We suspect that the models begin to overfit to \bridge{}, the smallest of the three robot datasets, and specialize to its seen kitchens: \bridge{} OOD PSNR peaks early and then drops while \bridge{} validation PSNR keeps rising. In our evaluation, \methodthree{} peaks on \bridge{} OOD at 22.94\,dB after ${\sim}10$ passes over \bridge{} and ends at 22.47\,dB, while its validation PSNR rises from 24.40 to 26.55\,dB; \methodeight{} peaks at 22.57\,dB after ${\sim}30$ passes and ends at 22.24\,dB, while its validation PSNR rises from 26.14 to 26.54\,dB.

\paragraph{Predicting object motion.}
We also improve predictions of the robot and moving objects, measured separately in Fig.~\ref{fig:region-results}. On the \bridge{} OOD split, \methodeight{} raises moving-object PSNR from 21.50 to 24.85 dB and SCS from 0.778 to 0.831 over the action-only baseline. We show more consistent object motion on unseen objects in Fig.~\ref{fig:qualitative_generalization}. We can see that the baseline moves the arm but leaves the object on the table, while we predict the lift. We provide further examples of trajectory generalization in Fig. \ref{fig:additional-droid-trajectory}.

\begin{figure}[!htbp]
    \centering
    \includegraphics[width=\linewidth]{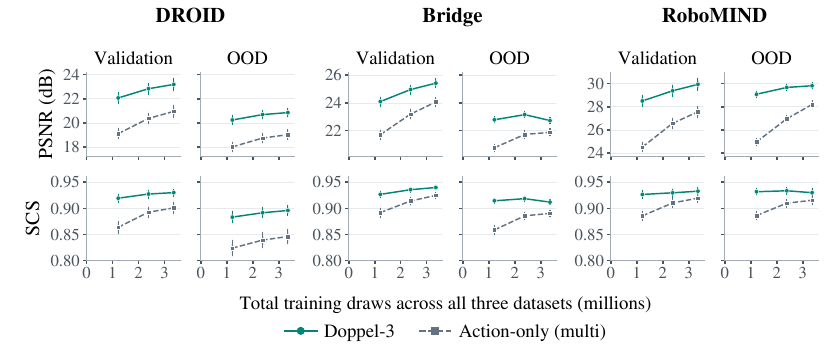}
    \caption{Training efficiency of \methodthree{} versus a three-robot action-only baseline. \method{} achieves higher validation and OOD prediction quality throughout 3.36M training draws.}

    \label{fig:trainingeff_multi}
\end{figure}

\paragraph{Training efficiency.}
We outperform separate action-only experts at comparable numbers of training draws per dataset (Fig.~\ref{fig:trainingeff}). We also compare with a shared action-only model, matching the batch size and optimization schedule and counting total draws across datasets (Fig.~\ref{fig:trainingeff_multi}). Averaged across datasets, we obtain higher validation and OOD PSNR and SCS after 1.216M draws than this baseline obtains after 3.356M draws, using 63.8\% fewer draws. These counts include repeated examples. We provide additional quantitative results in Appendix~\ref{app:results}.

\begin{figure*}[!htbp]
  \centering
  \includegraphics[width=\textwidth]{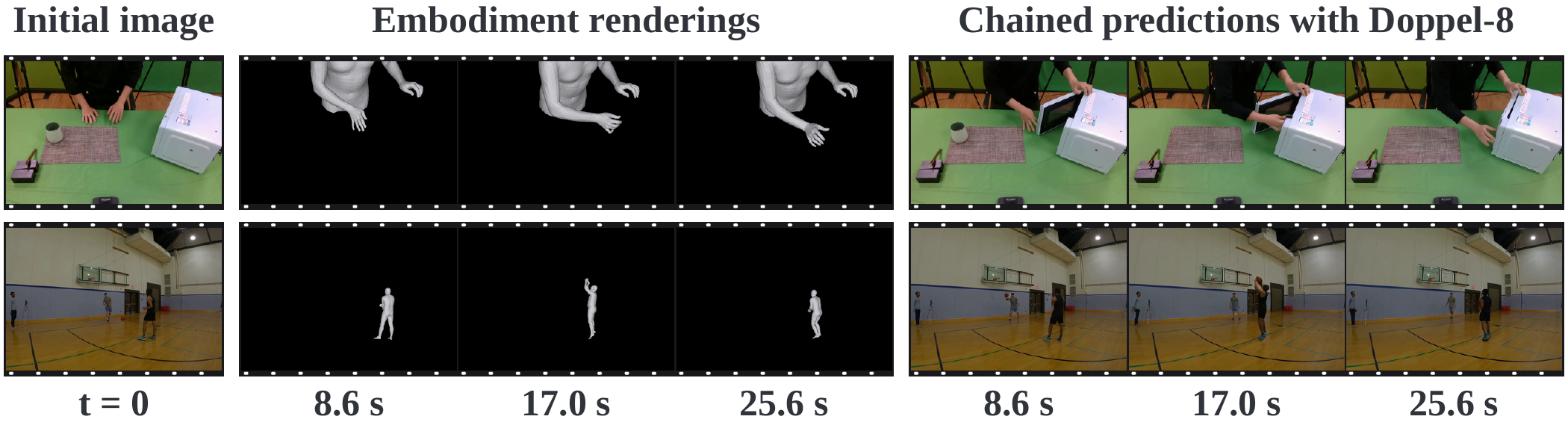}
  \caption{Long-horizon rollouts of \methodeight{} on held-out test episodes from OakInk2 (top) and Ego-Exo4D (bottom). Each episode is predicted for 25.6\,s from its first frame alone by chaining 3.2\,s windows on the model's own predictions. Each row shows the initial image, the embodiment renderings given as conditioning, and our predictions at 8.6, 17.0, and 25.6\,s.}
  \label{fig:long-horizon}
\end{figure*}

\paragraph{Long-horizon rollouts.} We chain 3.2\,s predictions to roll out held-out test episodes for 25.6\,s from a single real frame: each window starts from the model's own last predicted frame and receives that window's rendering and actions, so after the first frame the model never observes the real video. 
Fig.~\ref{fig:long-horizon} shows two long horizon rollouts.


\section{Discussion, Limitations, and Conclusion}
\label{sec:discussion}

We presented \method{}, a shared video world model that uses known robot geometry to predict interactions across embodiments. \method{} combines numerical actions with camera-aligned future robot renderings. This provides the intended robot motion explicitly, reducing what the model must learn about how joint configurations translate into image-space motion. Across \droid{}, \bridge{}, and \robomind{}, we improve whole-frame prediction over action-only and skeleton-conditioned baselines. Our region-specific evaluation also shows better predictions of moving-object appearance and motion, suggesting that explicit robot motion helps the model learn the environment's response. We observe these gains on held-out trajectories and OOD interactions with new objects and scenes. We retain them with one model trained across all three embodiments and outperform the shared action-only baseline with fewer training draws. Extending training to eight embodiments, including human hands and bodies, further improves robot predictions on \droid{} and \robomind{}, suggesting that rendered bodies let a world model learn from data well beyond a single robot. This supports using geometric action representations to share dynamics learning across robots with different bodies and configuration spaces.

We require accurate robot geometry, a URDF, and camera calibration to construct the renderings. Although we recover missing extrinsics in offline datasets, unknown or rapidly changing viewpoints remain a limitation. Camera-aware or camera-invariant representations could reduce this dependence. We also condition on configuration trajectories; handling imperfect geometry or lower-level control commands would require modeling uncertainty in robot motion and execution. Our evaluation covers embodiments seen during training and one camera per trajectory. We leave transfer to unseen embodiments, multi-view prediction, and longer-horizon prediction to future work. Combining pretrained video models with learned 3D representations may further reduce the need for explicit pixel-level alignment.

\newpage
\bibliography{iclr2027_conference_corrected}

@inproceedings{ha2018worldmodels,
  title     = {Recurrent World Models Facilitate Policy Evolution},
  author    = {Ha, David and Schmidhuber, J{\"u}rgen},
  booktitle = {Advances in Neural Information Processing Systems (NeurIPS)},
  volume    = {31},
  pages     = {2450--2462},
  year      = {2018}
}

@article{hafner2025dreamerv3,
  title     = {Mastering diverse control tasks through world models},
  author    = {Hafner, Danijar and Pasukonis, Jurgis and Ba, Jimmy and Lillicrap, Timothy},
  journal   = {Nature},
  volume    = {640},
  number    = {8059},
  pages     = {647--653},
  year      = {2025},
  doi       = {10.1038/s41586-025-08744-2}
}

@inproceedings{finn2017deep,
  title     = {Deep Visual Foresight for Planning Robot Motion},
  author    = {Finn, Chelsea and Levine, Sergey},
  booktitle = {IEEE International Conference on Robotics and Automation (ICRA)},
  pages     = {2786--2793},
  year      = {2017},
  doi       = {10.1109/ICRA.2017.7989324}
}

@article{ebert2018visual,
  title     = {{Visual Foresight}: Model-Based Deep Reinforcement Learning for Vision-Based Robotic Control},
  author    = {Ebert, Frederik and Finn, Chelsea and Dasari, Sudeep and Xie, Annie and Lee, Alex and Levine, Sergey},
  journal   = {arXiv preprint arXiv:1812.00568},
  year      = {2018}
}

@inproceedings{bruce2024genie,
  title     = {{Genie}: Generative Interactive Environments},
  author    = {Bruce, Jake and Dennis, Michael and Edwards, Ashley and Parker-Holder, Jack and Shi, Yuge and Hughes, Edward and Lai, Matthew and Mavalankar, Aditi and Steigerwald, Richie and Apps, Chris and Aytar, Yusuf and Bechtle, Sarah and Behbahani, Feryal and Chan, Stephanie and Heess, Nicolas and Gonzalez, Lucy and Osindero, Simon and Ozair, Sherjil and Reed, Scott and Zhang, Jingwei and Zolna, Konrad and Clune, Jeff and de Freitas, Nando and Singh, Satinder and Rockt{\"a}schel, Tim},
  booktitle = {International Conference on Machine Learning (ICML)},
  volume    = {235},
  pages     = {4603--4623},
  year      = {2024}
}

@article{agarwal2025cosmos,
  title     = {{Cosmos} World Foundation Model Platform for Physical {AI}},
  author    = {{NVIDIA} and Agarwal, Niket and Ali, Arslan and Bala, Maciej and Balaji, Yogesh and Barker, Erik and Cai, Tiffany and Chattopadhyay, Prithvijit and Chen, Yongxin and Cui, Yin and Ding, Yifan and Dworakowski, Daniel and Fan, Jiaojiao and Fenzi, Michele and Ferroni, Francesco and Fidler, Sanja and Fox, Dieter and Ge, Songwei and Ge, Yunhao and Gu, Jinwei and Gururani, Siddharth and He, Ethan and Huang, Jiahui and Huffman, Jacob and Jannaty, Pooya and Jin, Jingyi and Kim, Seung Wook and Kl{\'a}r, Gergely and Lam, Grace and Lan, Shiyi and Leal-Taixe, Laura and Li, Anqi and Li, Zhaoshuo and Lin, Chen-Hsuan and Lin, Tsung-Yi and Ling, Huan and Liu, Ming-Yu and Liu, Xian and Luo, Alice and Ma, Qianli and Mao, Hanzi and Mo, Kaichun and Mousavian, Arsalan and Nah, Seungjun and Niverty, Sriharsha and Page, David and Paschalidou, Despoina and Patel, Zeeshan and Pavao, Lindsey and Ramezanali, Morteza and Reda, Fitsum and Ren, Xiaowei and Sabavat, Vasanth Rao Naik and Schmerling, Ed and Shi, Stella and Stefaniak, Bartosz and Tang, Shitao and Tchapmi, Lyne and Tredak, Przemek and Tseng, Wei-Cheng and Varghese, Jibin and Wang, Hao and Wang, Haoxiang and Wang, Heng and Wang, Ting-Chun and Wei, Fangyin and Wei, Xinyue and Wu, Jay Zhangjie and Xu, Jiashu and Yang, Wei and Yen-Chen, Lin and Zeng, Xiaohui and Zeng, Yu and Zhang, Jing and Zhang, Qinsheng and Zhang, Yuxuan and Zhao, Qingqing and Zolkowski, Artur},
  journal   = {arXiv preprint arXiv:2501.03575},
  year      = {2025}
}

@article{blattmann2023svd,
  title     = {{Stable Video Diffusion}: Scaling Latent Video Diffusion Models to Large Datasets},
  author    = {Blattmann, Andreas and Dockhorn, Tim and Kulal, Sumith and Mendelevitch, Daniel and Kilian, Maciej and Lorenz, Dominik and Levi, Yam and English, Zion and Voleti, Vikram and Letts, Adam and Jampani, Varun and Rombach, Robin},
  journal   = {arXiv preprint arXiv:2311.15127},
  year      = {2023}
}

@inproceedings{yang2025cogvideox,
  title     = {{CogVideoX}: Text-to-Video Diffusion Models with An Expert Transformer},
  author    = {Yang, Zhuoyi and Teng, Jiayan and Zheng, Wendi and Ding, Ming and Huang, Shiyu and Xu, Jiazheng and Yang, Yuanming and Hong, Wenyi and Zhang, Xiaohan and Feng, Guanyu and Yin, Da and Zhang, Yuxuan and Wang, Weihan and Cheng, Yean and Xu, Bin and Gu, Xiaotao and Dong, Yuxiao and Tang, Jie},
  booktitle = {International Conference on Learning Representations (ICLR)},
  pages     = {83048--83077},
  year      = {2025}
}

@inproceedings{huang2025vid2world,
  title     = {{Vid2World}: Crafting Video Diffusion Models to Interactive World Models},
  author    = {Huang, Siqiao and Wu, Jialong and Zhou, Qixing and Miao, Shangchen and Long, Mingsheng},
  booktitle = {International Conference on Learning Representations (ICLR)},
  pages     = {10763--10806},
  year      = {2026}
}

@inproceedings{du2023unipi,
  title     = {Learning Universal Policies via Text-Guided Video Generation},
  author    = {Du, Yilun and Yang, Mengjiao and Dai, Bo and Dai, Hanjun and Nachum, Ofir and Tenenbaum, Joshua B. and Schuurmans, Dale and Abbeel, Pieter},
  booktitle = {Advances in Neural Information Processing Systems (NeurIPS)},
  volume    = {36},
  pages     = {9156--9172},
  year      = {2023}
}

@inproceedings{quevedo2025worldgym,
  title     = {{WorldGym}: World Model as An Environment for Policy Evaluation},
  author    = {Quevedo, Julian and Sharma, Ansh Kumar and Sun, Yixiang and Suryavanshi, Varad and Liang, Percy and Yang, Sherry},
  booktitle = {International Conference on Learning Representations (ICLR)},
  pages     = {78932--78957},
  year      = {2026}
}

@inproceedings{zhu2025uwm,
  title     = {Unified World Models: Coupling Video and Action Diffusion for Pretraining on Large Robotic Datasets},
  author    = {Zhu, Chuning and Yu, Raymond and Feng, Siyuan and Burchfiel, Benjamin and Shah, Paarth and Gupta, Abhishek},
  booktitle = {Robotics: Science and Systems (RSS)},
  year      = {2025},
  doi       = {10.15607/RSS.2025.XXI.015}
}

@inproceedings{black2024pi0,
  title     = {$\pi_0$: A Vision-Language-Action Flow Model for General Robot Control},
  author    = {Black, Kevin and Brown, Noah and Driess, Danny and Esmail, Adnan and Equi, Michael and Finn, Chelsea and Fusai, Niccolo and Groom, Lachy and Hausman, Karol and Ichter, Brian and Jakubczak, Szymon and Jones, Tim and Ke, Liyiming and Levine, Sergey and Li-Bell, Adrian and Mothukuri, Mohith and Nair, Suraj and Pertsch, Karl and Shi, Lucy Xiaoyang and Tanner, James and Vuong, Quan and Walling, Anna and Wang, Haohuan and Zhilinsky, Ury},
  booktitle = {Robotics: Science and Systems (RSS)},
  year      = {2025},
  doi       = {10.15607/RSS.2025.XXI.010}
}

@inproceedings{wang2024hpt,
  title     = {Scaling Proprioceptive-Visual Learning with Heterogeneous Pre-trained Transformers},
  author    = {Wang, Lirui and Chen, Xinlei and Zhao, Jialiang and He, Kaiming},
  booktitle = {Advances in Neural Information Processing Systems (NeurIPS)},
  volume    = {37},
  pages     = {124420--124450},
  year      = {2024}
}

@inproceedings{doshi2024crossformer,
  title     = {Scaling Cross-Embodied Learning: One Policy for Manipulation, Navigation, Locomotion and Aviation},
  author    = {Doshi, Ria and Walke, Homer and Mees, Oier and Dasari, Sudeep and Levine, Sergey},
  booktitle = {Conference on Robot Learning (CoRL)},
  volume    = {270},
  pages     = {496--512},
  year      = {2024}
}

@inproceedings{chen2024mirage,
  title     = {Mirage: Cross-Embodiment Zero-Shot Policy Transfer with Cross-Painting},
  author    = {Chen, Lawrence Yunliang and Hari, Kush and Dharmarajan, Karthik and Xu, Chenfeng and Vuong, Quan and Goldberg, Ken},
  booktitle = {Robotics: Science and Systems (RSS)},
  year      = {2024},
  doi       = {10.15607/RSS.2024.XX.069}
}

@inproceedings{chen2024roviaug,
  title     = {{RoVi-Aug}: Robot and Viewpoint Augmentation for Cross-Embodiment Robot Learning},
  author    = {Chen, Lawrence Yunliang and Xu, Chenfeng and Dharmarajan, Karthik and Irshad, Muhammad Zubair and Cheng, Richard and Keutzer, Kurt and Tomizuka, Masayoshi and Vuong, Quan and Goldberg, Ken},
  booktitle = {Conference on Robot Learning (CoRL)},
  volume    = {270},
  pages     = {209--233},
  year      = {2024}
}

@inproceedings{lepert2025shadow,
  title     = {{SHADOW}: Leveraging Segmentation Masks for Cross-Embodiment Policy Transfer},
  author    = {Lepert, Marion and Doshi, Ria and Bohg, Jeannette},
  booktitle = {Conference on Robot Learning (CoRL)},
  volume    = {270},
  pages     = {3536--3550},
  year      = {2024}
}

@inproceedings{lepert2025phantom,
  title     = {Phantom: Training Robots Without Robots Using Only Human Videos},
  author    = {Lepert, Marion and Fang, Jiaying and Bohg, Jeannette},
  booktitle = {Conference on Robot Learning (CoRL)},
  volume    = {305},
  pages     = {4545--4565},
  year      = {2025}
}

@inproceedings{xu2024im2flow2act,
  title     = {Flow as the Cross-Domain Manipulation Interface},
  author    = {Xu, Mengda and Xu, Zhenjia and Xu, Yinghao and Chi, Cheng and Wetzstein, Gordon and Veloso, Manuela and Song, Shuran},
  booktitle = {Conference on Robot Learning (CoRL)},
  volume    = {270},
  pages     = {2475--2499},
  year      = {2024}
}

@inproceedings{geng2025motionprompting,
  title     = {Motion Prompting: Controlling Video Generation with Motion Trajectories},
  author    = {Geng, Daniel and Herrmann, Charles and Hur, Junhwa and Cole, Forrester and Zhang, Serena and Pfaff, Tobias and Lopez-Guevara, Tatiana and Aytar, Yusuf and Rubinstein, Michael and Sun, Chen and Wang, Oliver and Owens, Andrew and Sun, Deqing},
  booktitle = {IEEE/CVF Conference on Computer Vision and Pattern Recognition (CVPR)},
  pages     = {1--12},
  year      = {2025},
  doi       = {10.1109/CVPR52734.2025.00010}
}

@inproceedings{ye2026robodream,
  title     = {{RoboDream}: Compositional World Models for Scalable Robot Data Synthesis},
  author    = {Ye, Junjie and Xue, Rong and Van Hoorick, Basile and Li, Runhao and Rajaprakash, Harshitha and Tokmakov, Pavel and Irshad, Muhammad Zubair and Guizilini, Vitor and Wang, Yue},
  booktitle = {IEEE/RSJ International Conference on Intelligent Robots and Systems (IROS)},
  year      = {2026}
}

@inproceedings{ye2025anchordream,
  title     = {{AnchorDream}: Repurposing Video Diffusion for Embodiment-Aware Robot Data Synthesis},
  author    = {Ye, Junjie and Xue, Rong and Van Hoorick, Basile and Tokmakov, Pavel and Irshad, Muhammad Zubair and Wang, Yue and Guizilini, Vitor},
  booktitle = {IEEE International Conference on Robotics and Automation (ICRA)},
  year      = {2026},
  pages = {6137--6144},
  doi = {10.1109/ICRA57385.2026.11697505}
}

@inproceedings{guo2026vlaw,
  title     = {{VLAW}: Iterative Co-Improvement of Vision-Language-Action Policy and World Model},
  author    = {Guo, Yanjiang and Lee, Tony and Shi, Lucy Xiaoyang and Chen, Jianyu and Liang, Percy and Finn, Chelsea},
  booktitle = {Proceedings of the 43rd International Conference on Machine Learning (ICML)},
  series    = {Proceedings of Machine Learning Research},
  volume    = {306},
  pages     = {37998--38011},
  year      = {2026}
}

@inproceedings{seo2026stressdream,
  title     = {{StressDream}: Steering Video World Models for Robust Policy Evaluation and Improvement},
  author    = {Seo, Junwon and Veer, Sushant and Tian, Ran and Ding, Wenhao and Sharma, Apoorva and Leung, Karen and Schmerling, Edward and Pavone, Marco and Bajcsy, Andrea},
  booktitle = {Conference on Robot Learning (CoRL)},
  year      = {2026}
}

@inproceedings{zhen2025tesseract,
  title     = {{TesserAct}: Learning {4D} Embodied World Models},
  author    = {Zhen, Haoyu and Sun, Qiao and Zhang, Hongxin and Li, Junyan and Zhou, Siyuan and Du, Yilun and Gan, Chuang},
  booktitle = {IEEE/CVF International Conference on Computer Vision (ICCV)},
  pages     = {5337--5347},
  year      = {2025},
  doi       = {10.1109/ICCV51701.2025.00507}
}

@article{wolpert1995internal,
  title     = {An Internal Model for Sensorimotor Integration},
  author    = {Wolpert, Daniel M. and Ghahramani, Zoubin and Jordan, Michael I.},
  journal   = {Science},
  volume    = {269},
  number    = {5232},
  pages     = {1880--1882},
  year      = {1995},
  doi       = {10.1126/science.7569931}
}

@inproceedings{wu2022daydreamer,
  title     = {{DayDreamer}: World Models for Physical Robot Learning},
  author    = {Wu, Philipp and Escontrela, Alejandro and Hafner, Danijar and Goldberg, Ken and Abbeel, Pieter},
  booktitle = {Conference on Robot Learning (CoRL)},
  volume    = {205},
  pages     = {2226--2240},
  year      = {2022}
}

@inproceedings{yang2023unisim,
  title     = {Learning Interactive Real-World Simulators},
  author    = {Yang, Sherry and Du, Yilun and Ghasemipour, Kamyar and Tompson, Jonathan and Kaelbling, Leslie and Schuurmans, Dale and Abbeel, Pieter},
  booktitle = {International Conference on Learning Representations (ICLR)},
  pages     = {45210--45234},
  year      = {2024}
}

@article{wan2025,
  title     = {{Wan}: Open and Advanced Large-Scale Video Generative Models},
  author    = {Team Wan and Wang, Ang and Ai, Baole and Wen, Bin and Mao, Chaojie and Xie, Chen-Wei and Chen, Di and Yu, Feiwu and Zhao, Haiming and Yang, Jianxiao and Zeng, Jianyuan and Wang, Jiayu and Zhang, Jingfeng and Zhou, Jingren and Wang, Jinkai and Chen, Jixuan and Zhu, Kai and Zhao, Kang and Yan, Keyu and Huang, Lianghua and Feng, Mengyang and Zhang, Ningyi and Li, Pandeng and Wu, Pingyu and Chu, Ruihang and Feng, Ruili and Zhang, Shiwei and Sun, Siyang and Fang, Tao and Wang, Tianxing and Gui, Tianyi and Weng, Tingyu and Shen, Tong and Lin, Wei and Wang, Wei and Wang, Wei and Zhou, Wenmeng and Wang, Wente and Shen, Wenting and Yu, Wenyuan and Shi, Xianzhong and Huang, Xiaoming and Xu, Xin and Kou, Yan and Lv, Yangyu and Li, Yifei and Liu, Yijing and Wang, Yiming and Zhang, Yingya and Huang, Yitong and Li, Yong and Wu, You and Liu, Yu and Pan, Yulin and Zheng, Yun and Hong, Yuntao and Shi, Yupeng and Feng, Yutong and Jiang, Zeyinzi and Han, Zhen and Wu, Zhi-Fan and Liu, Ziyu},
  journal   = {arXiv preprint arXiv:2503.20314},
  year      = {2025}
}

@inproceedings{jang2025dreamgen,
  title     = {{DreamGen}: Unlocking Generalization in Robot Learning through Video World Models},
  author    = {Jang, Joel and Ye, Seonghyeon and Lin, Zongyu and Xiang, Jiannan and Bjorck, Johan and Fang, Yu and Hu, Fengyuan and Huang, Spencer and Kundalia, Kaushil and Lin, Yen-Chen and Magne, Lo\"{i}c and Mandlekar, Ajay and Narayan, Avnish and Tan, You Liang and Wang, Guanzhi and Wang, Jing and Wang, Qi and Xu, Yinzhen and Zeng, Xiaohui and Zheng, Kaiyuan and Zheng, Ruijie and Liu, Ming-Yu and Zettlemoyer, Luke and Fox, Dieter and Kautz, Jan and Reed, Scott and Zhu, Yuke and Fan, Linxi},
  booktitle = {Conference on Robot Learning (CoRL)},
  volume    = {305},
  pages     = {5170--5194},
  year      = {2025}
}

@inproceedings{zhu2024irasim,
  title     = {{IRASim}: A Fine-Grained World Model for Robot Manipulation},
  author    = {Zhu, Fangqi and Wu, Hongtao and Guo, Song and Liu, Yuxiao and Cheang, Chilam and Kong, Tao},
  booktitle = {IEEE/CVF International Conference on Computer Vision (ICCV)},
  pages     = {9834--9844},
  year      = {2025},
  doi       = {10.1109/ICCV51701.2025.00917}
}

@inproceedings{guo2025ctrlworld,
  title     = {{Ctrl-World}: A Controllable Generative World Model for Robot Manipulation},
  author    = {Guo, Yanjiang and Shi, Lucy Xiaoyang and Chen, Jianyu and Finn, Chelsea},
  booktitle = {International Conference on Learning Representations (ICLR)},
  pages     = {6121--6138},
  year      = {2026}
}

@inproceedings{bagchi2026egowm,
  title     = {Walk Through Paintings: Egocentric World Models from {Internet} Priors},
  author    = {Bagchi, Anurag and Bao, Zhipeng and Bharadhwaj, Homanga and Wang, Yu-Xiong and Tokmakov, Pavel and Hebert, Martial},
  booktitle = {European Conference on Computer Vision (ECCV)},
  pages     = {472--489},
  year      = {2026},
  doi       = {10.1007/978-3-032-37098-3_26}
}

@inproceedings{kim2025dexterous,
  title     = {Dexterous World Models},
  author    = {Kim, Byungjun and Kim, Taeksoo and Lee, Junyoung and Joo, Hanbyul},
  booktitle = {IEEE/CVF Conference on Computer Vision and Pattern Recognition (CVPR)},
  pages     = {29663--29673},
  year      = {2026}
}

@inproceedings{oxe2023,
  title     = {Open {X-Embodiment}: Robotic Learning Datasets and {RT-X} Models},
  author    = {O'Neill, Abby and Rehman, Abdul and Maddukuri, Abhiram and Gupta, Abhishek and Padalkar, Abhishek and Lee, Abraham and Pooley, Acorn and Gupta, Agrim and Mandlekar, Ajay and Jain, Ajinkya and Tung, Albert and Bewley, Alex and Herzog, Alex and Irpan, Alex and Khazatsky, Alexander and Rai, Anant and Gupta, Anchit and Wang, Andrew and Singh, Anikait and Garg, Animesh and Kembhavi, Aniruddha and Xie, Annie and Brohan, Anthony and Raffin, Antonin and Sharma, Archit and Yavary, Arefeh and Jain, Arhan and Balakrishna, Ashwin and Wahid, Ayzaan and Burgess-Limerick, Ben and Kim, Beomjoon and Sch{\"o}lkopf, Bernhard and Wulfe, Blake and Ichter, Brian and Lu, Cewu and Xu, Charles and Le, Charlotte and Finn, Chelsea and Wang, Chen and Xu, Chenfeng and Chi, Cheng and Huang, Chenguang and Chan, Christine and Agia, Christopher and Pan, Chuer and Fu, Chuyuan and Devin, Coline and Xu, Danfei and Morton, Daniel and Driess, Danny and Chen, Daphne and Pathak, Deepak and Shah, Dhruv and B{\"u}chler, Dieter and Jayaraman, Dinesh and Kalashnikov, Dmitry and Sadigh, Dorsa and Johns, Edward and Foster, Ethan and Liu, Fangchen and Ceola, Federico and Xia, Fei and Zhao, Feiyu and Stulp, Freek and Zhou, Gaoyue and Sukhatme, Gaurav S. and Salhotra, Gautam and Yan, Ge and Feng, Gilbert and Schiavi, Giulio and Berseth, Glen and Kahn, Gregory and Wang, Guanzhi and Su, Hao and Fang, Hao-Shu and Shi, Haochen and Bao, Henghui and Ben Amor, Heni and Christensen, Henrik I and Furuta, Hiroki and Walke, Homer and Fang, Hongjie and Ha, Huy and Mordatch, Igor and Radosavovic, Ilija and Leal, Isabel and Liang, Jacky and Abou-Chakra, Jad and Kim, Jaehyung and Drake, Jaimyn and Peters, Jan and Schneider, Jan and Hsu, Jasmine and Bohg, Jeannette and Bingham, Jeffrey and Wu, Jeffrey and Gao, Jensen and Hu, Jiaheng and Wu, Jiajun and Wu, Jialin and Sun, Jiankai and Luo, Jianlan and Gu, Jiayuan and Tan, Jie and Oh, Jihoon and Wu, Jimmy and Lu, Jingpei and Yang, Jingyun and Malik, Jitendra and Silv{\'e}rio, Jo{\~a}o and Hejna, Joey and Booher, Jonathan and Tompson, Jonathan and Yang, Jonathan and Salvador, Jordi and Lim, Joseph J. and Han, Junhyek and Wang, Kaiyuan and Rao, Kanishka and Pertsch, Karl and Hausman, Karol and Go, Keegan and Gopalakrishnan, Keerthana and Goldberg, Ken and Byrne, Kendra and Oslund, Kenneth and Kawaharazuka, Kento and Black, Kevin and Lin, Kevin and Zhang, Kevin and Ehsani, Kiana and Lekkala, Kiran and Ellis, Kirsty and Rana, Krishan and Srinivasan, Krishnan and Fang, Kuan and Singh, Kunal Pratap and Zeng, Kuo-Hao and Hatch, Kyle and Hsu, Kyle and Itti, Laurent and Chen, Lawrence Yunliang and Pinto, Lerrel and Fei-Fei, Li and Tan, Liam and Fan, Linxi Jim and Ott, Lionel and Lee, Lisa and Weihs, Luca and Chen, Magnum and Lepert, Marion and Memmel, Marius and Tomizuka, Masayoshi and Itkina, Masha and Castro, Mateo Guaman and Spero, Max and Du, Maximilian and Ahn, Michael and Yip, Michael C. and Zhang, Mingtong and Ding, Mingyu and Heo, Minho and Srirama, Mohan Kumar and Sharma, Mohit and Kim, Moo Jin and Kanazawa, Naoaki and Hansen, Nicklas and Heess, Nicolas and Joshi, Nikhil J and Suenderhauf, Niko and Liu, Ning and Di Palo, Norman and Shafiullah, Nur Muhammad Mahi and Mees, Oier and Kroemer, Oliver and Bastani, Osbert and Sanketi, Pannag R and Miller, Patrick Tree and Yin, Patrick and Wohlhart, Paul and Xu, Peng and Fagan, Peter David and Mitrano, Peter and Sermanet, Pierre and Abbeel, Pieter and Sundaresan, Priya and Chen, Qiuyu and Vuong, Quan and Rafailov, Rafael and Tian, Ran and Doshi, Ria and Mart{\'\i}n-Mart{\'\i}n, Roberto and Baijal, Rohan and Scalise, Rosario and Hendrix, Rose and Lin, Roy and Qian, Runjia and Zhang, Ruohan and Mendonca, Russell and Shah, Rutav and Hoque, Ryan and Julian, Ryan and Bustamante, Samuel and Kirmani, Sean and Levine, Sergey and Lin, Shan and Moore, Sherry and Bahl, Shikhar and Dass, Shivin and Sonawani, Shubham and Song, Shuran and Xu, Sichun and Haldar, Siddhant and Karamcheti, Siddharth and Adebola, Simeon and Guist, Simon and Nasiriany, Soroush and Schaal, Stefan and Welker, Stefan and Tian, Stephen and Ramamoorthy, Subramanian and Dasari, Sudeep and Belkhale, Suneel and Park, Sungjae and Nair, Suraj and Mirchandani, Suvir and Osa, Takayuki and Gupta, Tanmay and Harada, Tatsuya and Matsushima, Tatsuya and Xiao, Ted and Kollar, Thomas and Yu, Tianhe and Ding, Tianli and Davchev, Todor and Zhao, Tony Z. and Armstrong, Travis and Darrell, Trevor and Chung, Trinity and Jain, Vidhi and Vanhoucke, Vincent and Zhan, Wei and Zhou, Wenxuan and Burgard, Wolfram and Chen, Xi and Wang, Xiaolong and Zhu, Xinghao and Geng, Xinyang and Liu, Xiyuan and Liangwei, Xu and Li, Xuanlin and Lu, Yao and Ma, Yecheng Jason and Kim, Yejin and Chebotar, Yevgen and Zhou, Yifan and Zhu, Yifeng and Wu, Yilin and Xu, Ying and Wang, Yixuan and Bisk, Yonatan and Cho, Yoonyoung and Lee, Youngwoon and Cui, Yuchen and Cao, Yue and Wu, Yueh-Hua and Tang, Yujin and Zhu, Yuke and Zhang, Yunchu and Jiang, Yunfan and Li, Yunshuang and Li, Yunzhu and Iwasawa, Yusuke and Matsuo, Yutaka and Ma, Zehan and Xu, Zhuo and Cui, Zichen Jeff and Zhang, Zichen and Lin, Zipeng},
  booktitle = {IEEE International Conference on Robotics and Automation (ICRA)},
  pages     = {6892--6903},
  year      = {2024},
  doi       = {10.1109/ICRA57147.2024.10611477}
}

@inproceedings{octo2024,
  title     = {{Octo}: An Open-Source Generalist Robot Policy},
  author    = {Ghosh, Dibya and Walke, Homer and Pertsch, Karl and Black, Kevin and Mees, Oier and Dasari, Sudeep and Hejna, Joey and Kreiman, Tobias and Xu, Charles and Luo, Jianlan and Tan, You Liang and Chen, Lawrence Yunliang and Sanketi, Pannag and Vuong, Quan and Xiao, Ted and Sadigh, Dorsa and Finn, Chelsea and Levine, Sergey},
  booktitle = {Robotics: Science and Systems (RSS)},
  year      = {2024},
  doi       = {10.15607/RSS.2024.XX.090}
}

@inproceedings{kim2024openvla,
  title     = {{OpenVLA}: An Open-Source Vision-Language-Action Model},
  author    = {Kim, Moo Jin and Pertsch, Karl and Karamcheti, Siddharth and Xiao, Ted and Balakrishna, Ashwin and Nair, Suraj and Rafailov, Rafael and Foster, Ethan and Sanketi, Pannag and Vuong, Quan and Kollar, Thomas and Burchfiel, Benjamin and Tedrake, Russ and Sadigh, Dorsa and Levine, Sergey and Liang, Percy and Finn, Chelsea},
  booktitle = {Conference on Robot Learning (CoRL)},
  volume    = {270},
  pages     = {2679--2713},
  year      = {2024}
}

@inproceedings{he2025cross,
  title     = {Scaling Cross-Embodiment World Models for Dexterous Manipulation},
  author    = {He, Zihao and Ai, Bo and Mu, Tongzhou and Liu, Yulin and Wan, Weikang and Fu, Jiawei and Du, Yilun and Christensen, Henrik I. and Su, Hao},
  booktitle = {IEEE/RSJ International Conference on Intelligent Robots and Systems (IROS)},
  year      = {2026}
}

@inproceedings{gu2023rttrajectory,
  title     = {{RT-Trajectory}: Robotic Task Generalization via Hindsight Trajectory Sketches},
  author    = {Gu, Jiayuan and Kirmani, Sean and Wohlhart, Paul and Lu, Yao and Gonzalez Arenas, Montserrat and Rao, Kanishka and Yu, Wenhao and Fu, Chuyuan and Gopalakrishnan, Keerthana and Xu, Zhuo and Sundaresan, Priya and Xu, Peng and Su, Hao and Hausman, Karol and Finn, Chelsea and Vuong, Quan and Xiao, Ted},
  booktitle = {International Conference on Learning Representations (ICLR)},
  pages     = {44918--44941},
  year      = {2024}
}

@inproceedings{wu2024draganything,
  title     = {{DragAnything}: Motion Control for Anything Using Entity Representation},
  author    = {Wu, Weijia and Li, Zhuang and Gu, Yuchao and Zhao, Rui and He, Yefei and Zhang, David Junhao and Shou, Mike Zheng and Li, Yan and Gao, Tingting and Zhang, Di},
  booktitle = {European Conference on Computer Vision (ECCV)},
  pages     = {331--348},
  year      = {2024},
  doi       = {10.1007/978-3-031-72670-5_19}
}

@inproceedings{wang2025visualactions,
  title     = {Precise Action-to-Video Generation Through Visual Action Prompts},
  author    = {Wang, Yuang and Wen, Chao and Guo, Haoyu and Peng, Sida and Qin, Minghan and Bao, Hujun and Zhou, Xiaowei and Hu, Ruizhen},
  booktitle = {IEEE/CVF International Conference on Computer Vision (ICCV)},
  pages     = {12713--12724},
  year      = {2025},
  doi       = {10.1109/ICCV51701.2025.01181}
}

@inproceedings{teed2020raft,
  title     = {{RAFT}: Recurrent All-Pairs Field Transforms for Optical Flow},
  author    = {Teed, Zachary and Deng, Jia},
  booktitle = {European Conference on Computer Vision (ECCV)},
  pages     = {402--419},
  year      = {2020},
  doi       = {10.1007/978-3-030-58536-5_24}
}

@inproceedings{liu2024differentiable,
  title     = {Differentiable Robot Rendering},
  author    = {Liu, Ruoshi and Canberk, Alper and Song, Shuran and Vondrick, Carl},
  booktitle = {Conference on Robot Learning (CoRL)},
  volume    = {270},
  pages     = {117--129},
  year      = {2024}
}

@inproceedings{roboarenainfinity,
  title     = {{RobotArena} $\infty$: Scalable Robot Benchmarking via Real-to-Sim Translation},
  author    = {Jangir, Yash and Zhang, Yidi and Lo, Pang-Chi and Yamazaki, Kashu and Zhang, Chenyu and Tu, Kuan-Hsun and Ke, Tsung-Wei and Ke, Lei and Bisk, Yonatan and Fragkiadaki, Katerina},
  booktitle = {International Conference on Learning Representations (ICLR)},
  pages     = {25017--25043},
  year      = {2026}
}

@inproceedings{droid2024,
  title     = {{DROID}: A Large-Scale In-The-Wild Robot Manipulation Dataset},
  author    = {Alexander Khazatsky and Karl Pertsch and Suraj Nair and Ashwin Balakrishna and Sudeep Dasari and Siddharth Karamcheti and Soroush Nasiriany and Mohan Kumar Srirama and Lawrence Yunliang Chen and Kirsty Ellis and Peter David Fagan and Joey Hejna and Masha Itkina and Marion Lepert and Yecheng Jason Ma and Patrick Tree Miller and Jimmy Wu and Suneel Belkhale and Shivin Dass and Huy Ha and Arhan Jain and Abraham Lee and Youngwoon Lee and Marius Memmel and Sungjae Park and Ilija Radosavovic and Kaiyuan Wang and Albert Zhan and Kevin Black and Cheng Chi and Kyle Beltran Hatch and Shan Lin and Jingpei Lu and Jean Mercat and Abdul Rehman and Pannag R Sanketi and Archit Sharma and Cody Simpson and Quan Vuong and Homer Rich Walke and Blake Wulfe and Ted Xiao and Jonathan Heewon Yang and Arefeh Yavary and Tony Z. Zhao and Christopher Agia and Rohan Baijal and Mateo Guaman Castro and Daphne Chen and Qiuyu Chen and Trinity Chung and Jaimyn Drake and Ethan Paul Foster and Jensen Gao and Vitor Guizilini and David Antonio Herrera and Minho Heo and Kyle Hsu and Jiaheng Hu and Muhammad Zubair Irshad and Donovon Jackson and Charlotte Le and Yunshuang Li and Kevin Lin and Roy Lin and Zehan Ma and Abhiram Maddukuri and Suvir Mirchandani and Daniel Morton and Tony Nguyen and Abigail O'Neill and Rosario Scalise and Derick Seale and Victor Son and Stephen Tian and Emi Tran and Andrew E. Wang and Yilin Wu and Annie Xie and Jingyun Yang and Patrick Yin and Yunchu Zhang and Osbert Bastani and Glen Berseth and Jeannette Bohg and Ken Goldberg and Abhinav Gupta and Abhishek Gupta and Dinesh Jayaraman and Joseph J Lim and Jitendra Malik and Roberto Martín-Martín and Subramanian Ramamoorthy and Dorsa Sadigh and Shuran Song and Jiajun Wu and Michael C. Yip and Yuke Zhu and Thomas Kollar and Sergey Levine and Chelsea Finn},
  booktitle = {Robotics: Science and Systems (RSS)},
  year      = {2024},
  doi       = {10.15607/RSS.2024.XX.120}
}

@inproceedings{walke2023bridgedata,
  title     = {{BridgeData} {V2}: A Dataset for Robot Learning at Scale},
  author    = {Walke, Homer and Black, Kevin and Lee, Abraham and Kim, Moo Jin and Du, Max and Zheng, Chongyi and Zhao, Tony and Hansen-Estruch, Philippe and Vuong, Quan and He, Andre and Myers, Vivek and Fang, Kuan and Finn, Chelsea and Levine, Sergey},
  booktitle = {Conference on Robot Learning (CoRL)},
  volume    = {229},
  pages     = {1723--1736},
  year      = {2023}
}

@inproceedings{wu2024robomind,
  title     = {{RoboMIND}: Benchmark on Multi-embodiment Intelligence Normative Data for Robot Manipulation},
  author    = {Wu, Kun and Hou, Chengkai and Liu, Jiaming and Che, Zhengping and Ju, Xiaozhu and Yang, Zhuqin and Li, Meng and Zhao, Yinuo and Xu, Zhiyuan and Yang, Guang and Fan, Shichao and Wang, Xinhua and Liao, Fei and Zhao, Zhen and Li, Guangyu and Jin, Zhao and Wang, Lecheng and Mao, Jilei and Liu, Ning and Ren, Pei and Zhang, Qiang and Lyu, Yaoxu and Liu, Mengzhen and He, Jingyang and Luo, Yulin and Gao, Zeyu and Li, Chenxuan and Gu, Chenyang and Fu, Yankai and Wu, Di and Wang, Xingyu and Chen, Sixiang and Wang, Zhenyu and An, Pengju and Qian, Siyuan and Zhang, Shanghang and Tang, Jian},
  booktitle = {Robotics: Science and Systems (RSS)},
  year      = {2025},
  doi       = {10.15607/RSS.2025.XXI.152}
}

@article{oscar2026,
  title     = {{OSCAR}: Omni-Embodiment Action-Conditioned World Model for Robotics},
  author    = {Wu, Zhuoyuan and Gao, Jun},
  journal   = {arXiv preprint arXiv:2606.04463},
  year      = {2026}
}

@inproceedings{zhang2018lpips,
  title     = {The Unreasonable Effectiveness of Deep Features as a Perceptual Metric},
  author    = {Zhang, Richard and Isola, Phillip and Efros, Alexei A. and Shechtman, Eli and Wang, Oliver},
  booktitle = {IEEE/CVF Conference on Computer Vision and Pattern Recognition (CVPR)},
  pages     = {586--595},
  year      = {2018},
  doi       = {10.1109/CVPR.2018.00068}
}

@inproceedings{ravi2024sam2,
  title     = {{SAM} 2: Segment Anything in Images and Videos},
  author    = {Ravi, Nikhila and Gabeur, Valentin and Hu, Yuan-Ting and Hu, Ronghang and Ryali, Chaitanya and Ma, Tengyu and Khedr, Haitham and R{\"a}dle, Roman and Rolland, Chloe and Gustafson, Laura and Mintun, Eric and Pan, Junting and Alwala, Kalyan Vasudev and Carion, Nicolas and Wu, Chao-Yuan and Girshick, Ross and Doll{\'a}r, Piotr and Feichtenhofer, Christoph},
  booktitle = {International Conference on Learning Representations (ICLR)},
  pages     = {28085--28128},
  year      = {2025}
}

@inproceedings{yang2026orv,
  title     = {{ORV}: {4D} Occupancy-centric Robot Video Generation},
  author    = {Yang, Xiuyu and Li, Bohan and Xu, Shaocong and Wang, Nan and Ye, Chongjie and Chen, Zhaoxi and Qin, Minghan and Ding, Yikang and Zhu, Zheng and Jin, Xin and Zhao, Hang and Zhao, Hao},
  booktitle = {IEEE/CVF Conference on Computer Vision and Pattern Recognition (CVPR)},
  pages     = {1053--1066},
  year      = {2026}
}

@inproceedings{huang2026pointworld,
  title     = {{PointWorld}: Scaling {3D} World Models for In-The-Wild Robotic Manipulation},
  author    = {Huang, Wenlong and Chao, Yu-Wei and Mousavian, Arsalan and Liu, Ming-Yu and Fox, Dieter and Mo, Kaichun and Fei-Fei, Li},
  booktitle = {IEEE/CVF Conference on Computer Vision and Pattern Recognition (CVPR)},
  pages     = {20765--20779},
  year      = {2026}
}

@inproceedings{Wang_2025_CVPR,
  title     = {{MoGe}: Unlocking Accurate Monocular Geometry Estimation for Open-Domain Images with Optimal Training Supervision},
  author    = {Wang, Ruicheng and Xu, Sicheng and Dai, Cassie and Xiang, Jianfeng and Deng, Yu and Tong, Xin and Yang, Jiaolong},
  booktitle = {IEEE/CVF Conference on Computer Vision and Pattern Recognition (CVPR)},
  pages     = {5261--5271},
  year      = {2025},
  doi       = {10.1109/CVPR52734.2025.00496}
}

@inproceedings{lipman2023flowmatching,
  title     = {Flow Matching for Generative Modeling},
  author    = {Lipman, Yaron and Chen, Ricky T. Q. and Ben-Hamu, Heli and Nickel, Maximilian and Le, Matt},
  booktitle = {International Conference on Learning Representations (ICLR)},
  year      = {2023}
}

@inproceedings{fu2025gigahands,
  title={{GigaHands}: A Massive Annotated Dataset of Bimanual Hand Activities},
  author={Fu, Rao and Zhang, Dingxi and Jiang, Alex and Fu, Wanjia and Funk, Austin and Ritchie, Daniel and Sridhar, Srinath},
  booktitle = {IEEE/CVF Conference on Computer Vision and Pattern Recognition (CVPR)},
  year={2025},
  pages = {17461--17474}
}

@inproceedings{zhan2024oakink2,
  title={{OakInk2}: A Dataset of Bimanual Hands-Object Manipulation in Complex Task Completion},
  author={Zhan, Xinyu and Yang, Lixin and Zhao, Yifei and Mao, Kangrui and Xu, Hanlin and Lin, Zenan and Li, Kailin and Lu, Cewu},
  booktitle = {IEEE/CVF Conference on Computer Vision and Pattern Recognition (CVPR)},
  year={2024},
  pages = {445--456}
}

@inproceedings{grauman2024egoexo4d,
  title     = {{Ego-Exo4D}: Understanding Skilled Human Activity from First- and Third-Person Perspectives},
  author    = {Grauman, Kristen and Westbury, Andrew and Torresani, Lorenzo and Kitani, Kris and Malik, Jitendra and Afouras, Triantafyllos and Ashutosh, Kumar and Baiyya, Vijay and Bansal, Siddhant and Boote, Bikram and Byrne, Eugene and Chavis, Zach and Chen, Joya and Cheng, Feng and Chu, Fu-Jen and Crane, Sean and Dasgupta, Avijit and Dong, Jing and Escobar, Maria and Forigua, Cristhian and Gebreselasie, Abrham and Haresh, Sanjay and Huang, Jing and Islam, Md Mohaiminul and Jain, Suyog and Khirodkar, Rawal and Kukreja, Devansh and Liang, Kevin J and Liu, Jia-Wei and Majumder, Sagnik and Mao, Yongsen and Martin, Miguel and Mavroudi, Effrosyni and Nagarajan, Tushar and Ragusa, Francesco and Ramakrishnan, Santhosh Kumar and Seminara, Luigi and Somayazulu, Arjun and Song, Yale and Su, Shan and Xue, Zihui and Zhang, Edward and Zhang, Jinxu and Castillo, Angela and Chen, Changan and Fu, Xinzhu and Furuta, Ryosuke and Gonzalez, Cristina and Gupta, Prince and Hu, Jiabo and Huang, Yifei and Huang, Yiming and Khoo, Weslie and Kumar, Anush and Kuo, Robert and Lakhavani, Sach and Liu, Miao and Luo, Mi and Luo, Zhengyi and Meredith, Brighid and Miller, Austin and Oguntola, Oluwatumininu and Pan, Xiaqing and Peng, Penny and Pramanick, Shraman and Ramazanova, Merey and Ryan, Fiona and Shan, Wei and Somasundaram, Kiran and Song, Chenan and Southerland, Audrey and Tateno, Masatoshi and Wang, Huiyu and Wang, Yuchen and Yagi, Takuma and Yan, Mingfei and Yang, Xitong and Yu, Zecheng and Zha, Shengxin Cindy and Zhao, Chen and Zhao, Ziwei and Zhu, Zhifan and Zhuo, Jeff and Arbelaez, Pablo and Bertasius, Gedas and Damen, Dima and Engel, Jakob and Farinella, Giovanni Maria and Furnari, Antonino and Ghanem, Bernard and Hoffman, Judy and Jawahar, C.V. and Newcombe, Richard and Park, Hyun Soo and Rehg, James M. and Sato, Yoichi and Savva, Manolis and Shi, Jianbo and Shou, Mike Zheng and Wray, Michael},
  booktitle = {IEEE/CVF Conference on Computer Vision and Pattern Recognition (CVPR)},
  pages     = {19383--19400},
  year      = {2024}
}

@article{maddukuri2026egoexo4dhm,
  title     = {{Ego-Exo4D} Human Meshes Dataset: {4D} Human Motion Reconstruction for Ego-Exo Captures},
  author    = {Maddukuri, Abhiram and Pavlakos, Georgios},
  journal   = {arXiv preprint arXiv:2609.30187},
  year      = {2026}
}

@article{romero2017mano,
  title={Embodied Hands: Modeling and Capturing Hands and Bodies Together},
  author={Romero, Javier and Tzionas, Dimitrios and Black, Michael J.},
  journal={ACM Transactions on Graphics},
  volume={36},
  number={6},
  pages={245:1--245:17},
  year={2017}
}

@inproceedings{pavlakos2019smplx,
  title = {Expressive Body Capture: {3D} Hands, Face, and Body from a Single Image},
  author={Pavlakos, Georgios and Choutas, Vasileios and Ghorbani, Nima and Bolkart, Timo and Osman, Ahmed A. A. and Tzionas, Dimitrios and Black, Michael J.},
  booktitle = {IEEE/CVF Conference on Computer Vision and Pattern Recognition (CVPR)},
  year={2019},
  pages = {10967--10977}
}

@inproceedings{niu2026trex,
  title     = {{T-Rex}: Tactile-Reactive Dexterous Manipulation},
  author    = {Dantong Niu and Zhuoyang Liu and Zekai Wang and Boning Shao and Zhao-Heng Yin and Anirudh Pai and Yuvan Sharma and Stefano Saravalle and Ruijie Zheng and Jing Wang and Ryan Punamiya and Mengda Xu and Yuqi Xie and Yunfan Jiang and Letian Fu and Konstantinos Kallidromitis and Matteo Gioia and Junyi Zhang and Jiaxin Ge and Haiwen Feng and Fabio Galasso and Wei Zhan and David M. Chan and Yutong Bai and Roei Herzig and Jiahui Lei and Li Fei-Fei and Ken Goldberg and Jitendra Malik and Pieter Abbeel and Yuke Zhu and Danfei Xu and Linxi Fan and Trevor Darrell},
  booktitle = {Conference on Robot Learning (CoRL)},
  year      = {2026}
}

@inproceedings{allshire2026abc,
  title     = {Scalable Behavior Cloning with Open Data, Training, and Evaluation},
  author    = {Allshire, Arthur and Singh, Himanshu Gaurav and Singh, Ritvik and Rashid, Adam and Choi, Hongsuk and McAllister, David and Yu, Justin and Chen, Yiyuan and Huang, Huang and Abbeel, Pieter and Chen, Xi and Duan, Rocky and Isola, Phillip and Malik, Jitendra and Shentu, Fred and Shi, Guanya and Wu, Philipp and Kanazawa, Angjoo},
  booktitle = {Conference on Robot Learning (CoRL)},
  year      = {2026}
}

@article{chen2026bridgev2w,
  title     = {{BridgeV2W}: Bridging Video Generation Models to Embodied World Models via Embodiment Masks},
  author    = {Chen, Yixiang and Li, Peiyan and Yang, Jiabing and He, Keji and Wu, Xiangnan and Xu, Yuan and Wang, Kai and Liu, Jing and Liu, Nianfeng and Huang, Yan and Wang, Liang},
  journal   = {arXiv preprint arXiv:2602.03793},
  year      = {2026}
}

@article{jiang2025enerverseac,
  title     = {{EnerVerse-AC}: Envisioning Embodied Environments with Action Condition},
  author    = {Jiang, Yuxin and Chen, Shengcong and Huang, Siyuan and Chen, Liliang and Zhou, Pengfei and Liao, Yue and He, Xindong and Liu, Chiming and Li, Hongsheng and Yao, Maoqing and Ren, Guanghui},
  journal   = {arXiv preprint arXiv:2505.09723},
  year      = {2025}
}

@inproceedings{xu2026kinema4d,
  title     = {{Kinema4D}: Kinematic {4D} World Modeling for Spatiotemporal Embodied Simulation},
  author    = {Xu, Mutian and Zhang, Tianbao and Liu, Tianqi and Chen, Zhaoxi and Han, Xiaoguang and Liu, Ziwei},
  booktitle = {IEEE/CVF Conference on Computer Vision and Pattern Recognition (CVPR) Workshops},
  pages     = {4495--4500},
  year      = {2026}
}

@article{oquab2024dinov2,
  title     = {{DINOv2}: Learning Robust Visual Features without Supervision},
  author    = {Oquab, Maxime and Darcet, Timoth{\'e}e and Moutakanni, Th{\'e}o and Vo, Huy V. and Szafraniec, Marc and Khalidov, Vasil and Fernandez, Pierre and Haziza, Daniel and Massa, Francisco and El-Nouby, Alaaeldin and Assran, Mido and Ballas, Nicolas and Galuba, Wojciech and Howes, Russell and Huang, Po-Yao and Li, Shang-Wen and Misra, Ishan and Rabbat, Michael and Sharma, Vasu and Synnaeve, Gabriel and Xu, Hu and Jegou, Herve and Mairal, Julien and Labatut, Patrick and Joulin, Armand and Bojanowski, Piotr},
  journal   = {Transactions on Machine Learning Research},
  year      = {2024}
}

@article{nvidia2025worldsim,
  title     = {World Simulation with Video Foundation Models for Physical {AI}},
  author    = {NVIDIA and Arslan Ali and Junjie Bai and Maciej Bala and Yogesh Balaji and Aaron Blakeman and Tiffany Cai and Jiaxin Cao and Tianshi Cao and Elizabeth Cha and Yu-Wei Chao and Prithvijit Chattopadhyay and Mike Chen and Yongxin Chen and Yu Chen and Shuai Cheng and Yin Cui and Jenna Diamond and Yifan Ding and Jiaojiao Fan and Linxi Fan and Liang Feng and Francesco Ferroni and Sanja Fidler and Xiao Fu and Ruiyuan Gao and Yunhao Ge and Jinwei Gu and Aryaman Gupta and Siddharth Gururani and Imad El Hanafi and Ali Hassani and Zekun Hao and Jacob Huffman and Joel Jang and Pooya Jannaty and Jan Kautz and Grace Lam and Xuan Li and Zhaoshuo Li and Maosheng Liao and Chen-Hsuan Lin and Tsung-Yi Lin and Yen-Chen Lin and Huan Ling and Ming-Yu Liu and Xian Liu and Yifan Lu and Alice Luo and Qianli Ma and Hanzi Mao and Kaichun Mo and Seungjun Nah and Yashraj Narang and Abhijeet Panaskar and Lindsey Pavao and Trung Pham and Morteza Ramezanali and Fitsum Reda and Scott Reed and Xuanchi Ren and Haonan Shao and Yue Shen and Stella Shi and Shuran Song and Bartosz Stefaniak and Shangkun Sun and Shitao Tang and Sameena Tasmeen and Lyne Tchapmi and Wei-Cheng Tseng and Jibin Varghese and Andrew Z. Wang and Hao Wang and Haoxiang Wang and Heng Wang and Ting-Chun Wang and Fangyin Wei and Jiashu Xu and Dinghao Yang and Xiaodong Yang and Haotian Ye and Seonghyeon Ye and Xiaohui Zeng and Jing Zhang and Qinsheng Zhang and Kaiwen Zheng and Andrew Zhu and Yuke Zhu},
  journal   = {arXiv preprint arXiv:2511.00062},
  year      = {2025}
}

@inproceedings{fang2024rh20t,
  title     = {{RH20T}: A Comprehensive Robotic Dataset for Learning Diverse Skills in One-Shot},
  author    = {Hao-Shu Fang and Hongjie Fang and Zhenyu Tang and Jirong Liu and Chenxi Wang and Junbo Wang and Haoyi Zhu and Cewu Lu},
  booktitle = {IEEE International Conference on Robotics and Automation (ICRA)},
  pages     = {653--660},
  year      = {2024}
}

@article{tian2025interndataa1,
  title     = {{InternData-A1}: Pioneering High-Fidelity Synthetic Data for Pre-training Generalist Policy},
  author    = {Yang Tian and Yuyin Yang and Yiman Xie and Zetao Cai and Xu Shi and Ning Gao and Hangxu Liu and Xuekun Jiang and Zherui Qiu and Feng Yuan and Yaping Li and Ping Wang and Junhao Cai and Jia Zeng and Hao Dong and Jiangmiao Pang},
  journal   = {arXiv preprint arXiv:2511.16651},
  year      = {2025}
}

@inproceedings{agibot2025colosseo,
  title     = {{AgiBot World Colosseo}: A Large-scale Manipulation Platform for Scalable and Intelligent Embodied Systems},
  author    = {{AgiBot World Contributors} and Qingwen Bu and Jisong Cai and Li Chen and Xiuqi Cui and Yan Ding and Siyuan Feng and Shenyuan Gao and Xindong He and Xuan Hu and Xu Huang and Shu Jiang and Yuxin Jiang and Cheng Jing and Hongyang Li and Jialu Li and Chiming Liu and Yi Liu and Yuxiang Lu and Jianlan Luo and Ping Luo and Yao Mu and Yuehan Niu and Yixuan Pan and Jiangmiao Pang and Yu Qiao and Guanghui Ren and Cheng Ruan and Jiaqi Shan and Yongjian Shen and Chengshi Shi and Mingkang Shi and Modi Shi and Chonghao Sima and Jianheng Song and Huijie Wang and Wenhao Wang and Dafeng Wei and Chengen Xie and Guo Xu and Junchi Yan and Cunbiao Yang and Lei Yang and Shukai Yang and Maoqing Yao and Jia Zeng and Chi Zhang and Qinglin Zhang and Bin Zhao and Chengyue Zhao and Jiaqi Zhao and Jianchao Zhu},
  booktitle = {IEEE/RSJ International Conference on Intelligent Robots and Systems (IROS)},
  pages     = {3549--3556},
  year      = {2025}
}

@article{takanami2025airoamoma,
  title     = {{AIRoA MoMa} Dataset: A Large-Scale Hierarchical Dataset for Mobile Manipulation},
  author    = {Ryosuke Takanami and Petr Khrapchenkov and Shu Morikuni and Jumpei Arima and Yuta Takaba and Shunsuke Maeda and Takuya Okubo and Genki Sano and Satoshi Sekioka and Aoi Kadoya and Motonari Kambara and Naoya Nishiura and Haruto Suzuki and Takanori Yoshimoto and Koya Sakamoto and Shinnosuke Ono and Hu Yang and Daichi Yashima and Aoi Horo and Tomohiro Motoda and Kensuke Chiyoma and Hiroshi Ito and Koki Fukuda and Akihito Goto and Kazumi Morinaga and Yuya Ikeda and Riko Kawada and Masaki Yoshikawa and Norio Kosuge and Yuki Noguchi and Kei Ota and Tatsuya Matsushima and Yusuke Iwasawa and Yutaka Matsuo and Tetsuya Ogata},
  journal   = {arXiv preprint arXiv:2509.25032},
  year      = {2025}
}

@inproceedings{hoque2026egodex,
  title     = {{EgoDex}: Learning Dexterous Manipulation from Large-Scale Egocentric Video},
  author    = {Ryan Hoque and Peide Huang and David J. Yoon and Mouli Sivapurapu and Jian Zhang},
  booktitle = {International Conference on Learning Representations (ICLR)},
  year      = {2026},
  pages = {4218--4237}
}

@article{damen2022rescaling,
  title     = {Rescaling Egocentric Vision: Collection, Pipeline and Challenges for {EPIC-KITCHENS-100}},
  author    = {Dima Damen and Hazel Doughty and Giovanni Maria Farinella and Antonino Furnari and Evangelos Kazakos and Jian Ma and Davide Moltisanti and Jonathan Munro and Toby Perrett and Will Price and Michael Wray},
  journal   = {International Journal of Computer Vision},
  volume    = {130},
  number    = {1},
  pages     = {33--55},
  year      = {2022}
}

@article{wang2004ssim,
  title     = {Image Quality Assessment: From Error Visibility to Structural Similarity},
  author    = {Wang, Zhou and Bovik, A. C. and Sheikh, H. R. and Simoncelli, E. P.},
  journal   = {IEEE Transactions on Image Processing},
  volume    = {13},
  number    = {4},
  pages     = {600--612},
  year      = {2004}
}

@inproceedings{loshchilov2019adamw,
  title     = {Decoupled Weight Decay Regularization},
  author    = {Loshchilov, Ilya and Hutter, Frank},
  booktitle = {International Conference on Learning Representations (ICLR)},
  year      = {2019}
}
\bibliographystyle{unsrtnat}

\newpage 
\appendix
\section{Appendix}

\subsection{Website and Video}
\label{app:website}
Additional qualitative results are available on our project webpage, including extended video comparisons across all eight datasets, long-horizon rollouts, examples of trajectory and environment generalization, and side-by-side comparisons with our baselines. We encourage readers to view the webpage for the full video results, which are difficult to represent comprehensively in static paper figures.

\noindent\url{https://doppelwm.github.io/}

\subsection{Rendering Pipeline}
\label{app:rendering}

We render the robot along the supplied joint and gripper trajectory using an
articulated Gaussian robot model and the target camera calibration. The
renderer is fixed during world-model training. RGB observations, robot
configurations, and rendered frames are constructed using the same sampled
source-frame indices, ensuring temporal alignment across all conditioning
signals. When constructing training windows, we additionally verify that
cached renderings correspond to the same temporal sampling grid.

\begin{figure*}[t]
    \centering
    \includegraphics[width=\textwidth]{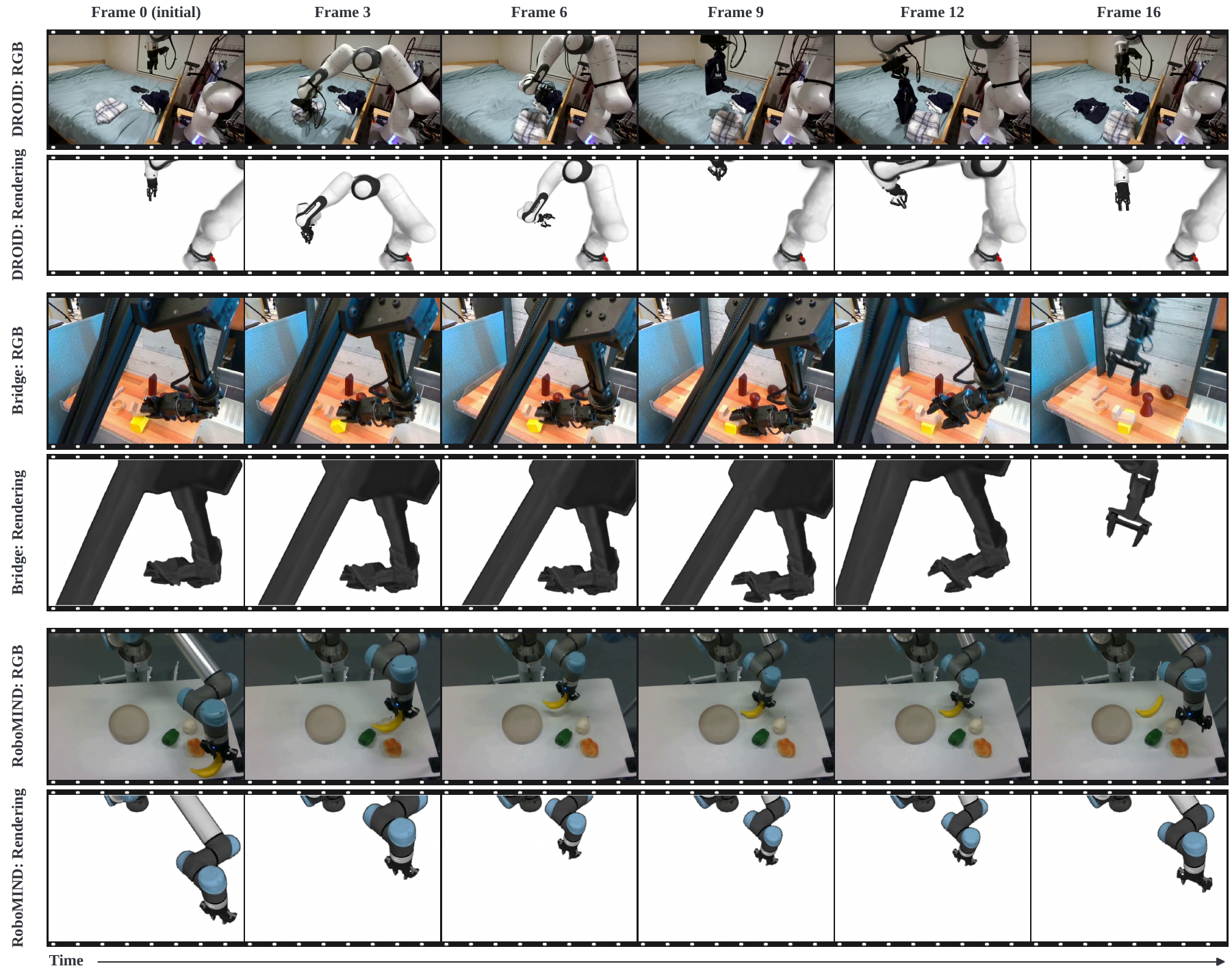}
    \caption{
        Alignment between the observed and rendered robot trajectories.
        The top row shows frames from the real RGB trajectory, and the
        bottom row shows the corresponding target-view robot renderings.
        The renderings express the commanded embodiment motion in the same
        image coordinates as the observations, providing the dense visual
        component of \method{}.
    }
    \label{fig:rendering-vs-real}
\end{figure*}

\begin{figure*}[t]
    \centering
    \includegraphics[width=\textwidth]{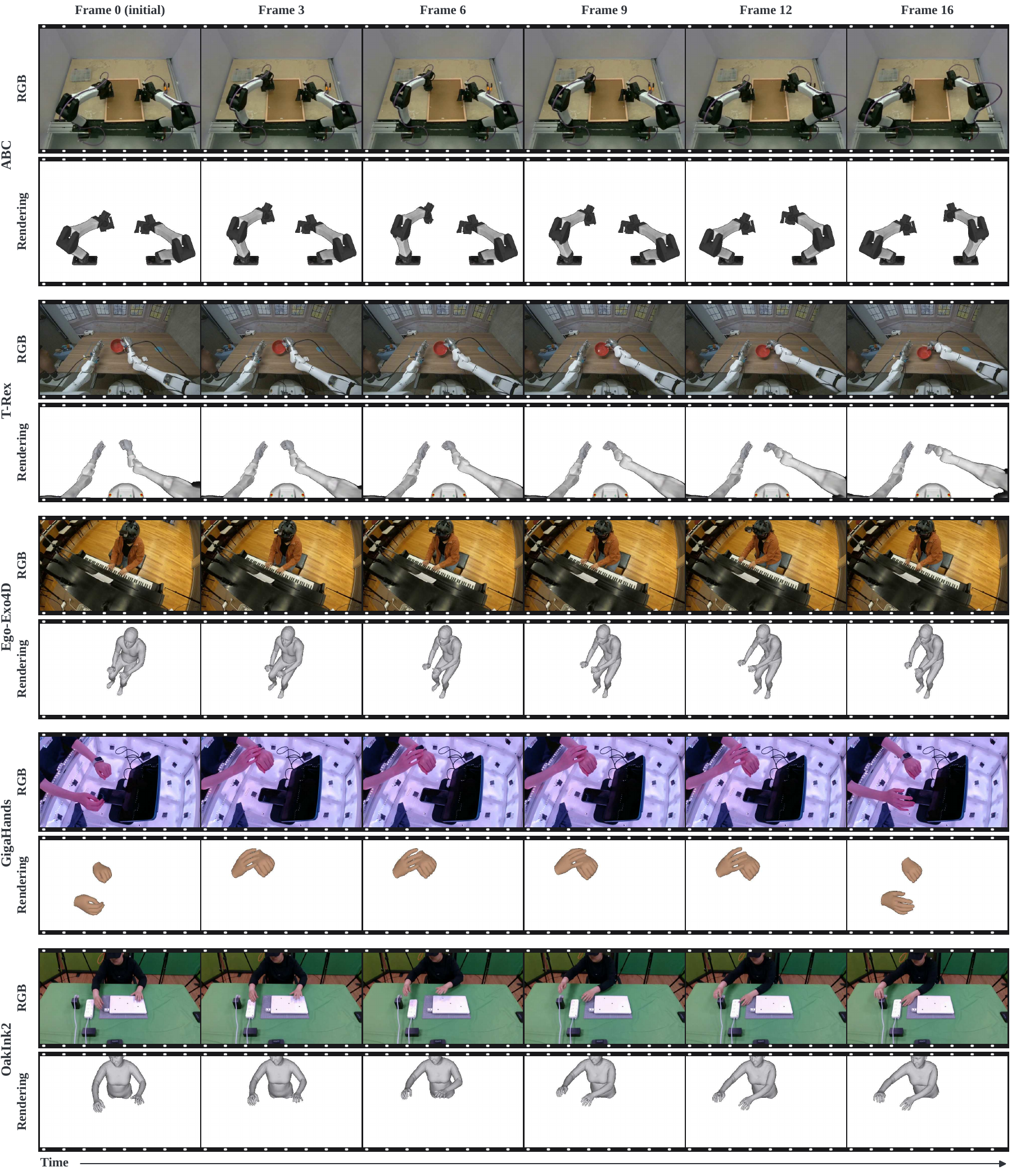}
    \caption{
        Alignment between the observed and rendered trajectories for the five added embodiments: a bimanual YAM robot (ABC), a humanoid with dexterous hands (T-Rex), a human body (Ego-Exo4D), human hands (GigaHands), and a human upper body (OakInk2). For each dataset, the top row shows the real RGB frames and the bottom row the corresponding camera-aligned renderings.
    }
    \label{fig:rendering-vs-real-others}
\end{figure*}

The robot-only renderings encode the embodiment's geometry and motion directly
in image coordinates; they contain no future scene or object information.
Figures~\ref{fig:rendering-vs-real} and~\ref{fig:rendering-vs-real-others} show representative alignment between the
rendered robot trajectories and the corresponding RGB observations across all
three datasets.

For each embodiment, we use the differentiable robot-rendering framework of
\citet{liu2024differentiable}. Starting from the robot's URDF, we train an articulated
Gaussian model using the original renderer architecture and training
procedure provided by the authors. We train a separate renderer for each robot
embodiment and keep its parameters frozen when training \method{}.

\paragraph{Additional embodiments.} The bimanual YAM arms in ABC are rendered with a DrRobot model trained for the YAM in the same way. The Vega-1 humanoid and its Sharpa hands in T-Rex are rendered from the robot's URDF meshes with the dataset's released head-camera calibration. People are rendered from fitted body and hand meshes: MANO hands for GigaHands~\citep{romero2017mano}, SMPL-X for OakInk2~\citep{pavlakos2019smplx}, and SMPL-H for Ego-Exo4D, using the meshes of \citet{maddukuri2026egoexo4dhm}; each mesh is rasterized through the dataset's camera model, including lens distortion. All renderings use a black background and contain only the embodiment.

\subsection{Dataset Construction and Evaluation Splits}
\label{app:data}

\paragraph{Training data.}
We construct training clips from demonstrations in \droid{}, \robomind{}, and \bridge{}. Within each dataset, we balance sampling across environment categories. From each demonstration, we extract up to five clips: four contiguous windows capturing local interaction dynamics and one full-span window covering the entire demonstration at a coarser temporal resolution. Demonstrations that are too short contribute fewer clips. Table~\ref{tab:data} reports the resulting numbers of demonstrations and training clips. During three-robot training, we sample uniformly across the three datasets, such that each embodiment contributes approximately one third of the optimization batches.

\paragraph{Additional embodiments.} For ABC, T-Rex, GigaHands, OakInk2, and Ego-Exo4D, we cut each video into back-to-back 3.2\,s windows at 5\,fps, plus equidistant segment clips and one whole-video clip; for Ego-Exo4D we use two exocentric cameras per take. Whole-video clips of ABC and Ego-Exo4D are evaluated but not used for training. Validation holds out a random 5\% of episodes. The test splits hold out: 19 of ABC's tasks; two of T-Rex's twelve wall backdrops (400 of 5,464 episodes), never seen in training or validation; one task per Ego-Exo4D activity (Mikan layup, chain cleaning, latte making, advanced salsa, CPR, piano scales, a bouldering route, and penalty kicks); five GigaHands activities (tea, stone, laptop, phone, crochet); and 51 OakInk2 tasks across its four scenes. Table~\ref{tab:corpus-hours} gives the amount of video per split.

\begin{table}[t]
  \centering
  \caption{Hours of video per dataset and split, from the full length of each source episode; each episode counts once, and Ego-Exo4D is counted per take (training uses up to two exocentric cameras per take, 176.3\,h of camera video).}
  \label{tab:corpus-hours}
  \small
  \begin{tabular}{@{}lrrrr@{}}
    \toprule
    Dataset & Train & Val. & Test & Total \\
    \midrule
    \droid{} & 98.9 & 5.4 & 11.6 & 115.9 \\
    \bridge{} & 26.4 & 1.4 & 3.2 & 31.1 \\
    \robomind{} & 62.1 & 3.2 & 6.8 & 72.1 \\
    ABC & 242.6 & 12.7 & 31.4 & 286.7 \\
    T-Rex & 44.2 & 2.4 & 3.9 & 50.5 \\
    Ego-Exo4D & 88.3 & 5.1 & 10.4 & 103.8 \\
    GigaHands & 25.4 & 1.2 & 2.1 & 28.6 \\
    OakInk2 & 5.2 & 1.2 & 1.5 & 7.9 \\
    \midrule
    All eight & 593.1 & 32.6 & 70.8 & 696.6 \\
    \bottomrule
  \end{tabular}
\end{table} \methodeight{} samples each dataset in proportion to its training clips, which gives about 60 passes over every dataset.

\paragraph{Evaluation splits.}
We evaluate generalization along two distinct axes. The validation split measures \emph{trajectory generalization}: for each embodiment, we reserve \num{200} clips from held-out demonstrations collected in environments that remain represented in the training set. These comprise \num{100} contiguous-window clips and \num{100} full-span clips. This split tests whether the model can predict previously unseen robot trajectories within familiar environment distributions.

The test split measures \emph{environment generalization} under three complementary distribution shifts. For \droid{}, we hold out an entire laboratory; no scene, camera, operator, or task from this laboratory appears in the training set. For \bridge{}, whose demonstrations are collected in toy-kitchen environments, we hold out one complete kitchen environment. For \robomind{}, we hold out four object-centric tasks and evaluate interactions involving object instances absent from the training set. These splits therefore test generalization to a completely new laboratory, an unseen instance of a semantically familiar environment, and unseen objects and tasks, respectively.

Each dataset contributes \num{200} test clips, evenly divided between \num{100} contiguous-window clips and \num{100} full-span clips. We report validation and test results separately because predicting a novel trajectory in a familiar environment and predicting interaction dynamics under an environmental distribution shift capture different generalization capabilities.

\subsection{Adapter architecture details}
\label{app:render-action-adapter}

This section provides the architectural details of the render--action adapter \(g_{\phi}\) introduced in Sec.~\ref{sec:method-redar}. The adapter contains about \(30.5\)M parameters and is shared across all embodiments.

\paragraph{Render normalization.}
Robot-only renderings are sparse, with most spatial locations
containing empty background. We first encode the rendering sequence \(\rend_{0:T-1}\) using the same frozen causal VAE of Wan 2.1:
\begin{equation}
z^{\mathrm{rend}}
=
E_{\mathrm{vae}}\!\left(r_{0:T-1}\right).
\label{eq:app-render-latents}
\end{equation}
Before applying the trainable render encoder, we subtract the per-sample, per-channel spatiotemporal mean:
\begin{equation}
\bar{z}^{\mathrm{rend}}_{c,t,h,w}
=
z^{\mathrm{rend}}_{c,t,h,w}
-
\frac{1}{T_z H_z W_z}
\sum_{t',h',w'}
z^{\mathrm{rend}}_{c,t',h',w'}.
\label{eq:app-render-centering}
\end{equation}
This prevents the empty-backgrounds from making the renders tokens very similar. If this mean subtractions is skipped the adapter can collapse to zero weights. We compute this operation in \(\mathrm{fp32}\), since the robot signal can be small relative to the shared latent response of the empty rendering background.

\paragraph{Render encoder.}
The centered render latent is processed by three 3D convolutions with GELU activations:
\begin{enumerate}
\item \(\operatorname{Conv3D}(16,256,1\times3\times3)\)
\item \(\operatorname{Conv3D}(256,512,3\times3\times3)\)
\item \(\operatorname{Conv3D}(512,512,1\times3\times3)\)
\end{enumerate}
The output grid is flattened into \(2025\) tokens of width \(512\). We then add factorized learned spatiotemporal positional embeddings,
\begin{equation}
p_{t,h,w}
=
p^{\mathrm{time}}_t
+
p^{\mathrm{height}}_h
+
p^{\mathrm{width}}_w,
\label{eq:app-render-position}
\end{equation}
using temporal, height, and width embedding tables of maximum sizes \(32\), \(32\), and \(64\), respectively.

Two pre-normalization self-attention blocks contextualize the render tokens. Each block uses eight attention heads, an FFN expansion factor of four, and residual connections around both the attention and FFN sublayers. This global mixing allows information from the spatially sparse robot region to reach tokens outside the convolutional encoder's local receptive field.

\paragraph{Action encoder.}
Each frame is associated with a configuration vector: up to seven joint values and one gripper value for the robot arms, and up to 159 values for the other embodiments. We zero-pad every vector to 159 dimensions. We normalize each embodiment using its own training-set statistics and zero-pad configurations with fewer degrees of freedom to the shared 159-dimensional layout. Padding provides a common input shape but does not establish correspondence between the joint dimensions of different embodiments.

A shared framewise MLP maps each action vector to width \(512\):
\begin{equation}
h^{\mathrm{act}}_t
=
W_2\,\operatorname{GELU}\!\left(W_1 a^e_t\right)
+
p^{\mathrm{act}}_t.
\label{eq:app-action-encoding}
\end{equation}
where \(W_1:\mathbb{R}^{159}\rightarrow\mathbb{R}^{512}\), \(W_2:\mathbb{R}^{512}\rightarrow\mathbb{R}^{512}\), and \(p^{\mathrm{act}}_t\) is selected from a learned temporal position table supporting up to \(128\) action frames.

\paragraph{Render--action fusion.}
We fuse the encoded streams using two pre-normalization cross-attention blocks. The \(2025\) render tokens serve as queries, while the framewise action tokens serve as keys and values:
\begin{equation}
H_{\mathrm{fused}}
=
\operatorname{CrossAttn}\!\left(
H_{\mathrm{rend}},
H_{\mathrm{act}},
H_{\mathrm{act}}
\right).
\label{eq:app-redar-cross-attention}
\end{equation}
Each block uses eight attention heads, an FFN expansion factor of four, and residual connections around both sublayers. A final projection maps each fused token to the Wan hidden dimension:
\begin{equation}
G
=
\operatorname{Linear}_{2048\rightarrow5120}\!\left(
\operatorname{GELU}\!\left(
\operatorname{Linear}_{512\rightarrow2048}\!\left(
\operatorname{LayerNorm}\!\left(H_{\mathrm{fused}}\right)
\right)
\right)
\right).
\label{eq:app-redar-projection}
\end{equation}
The resulting tensor has shape \(G\in\mathbb{R}^{2025\times5120}\). Neither the output projection nor its preceding layers are zero-initialized.

\paragraph{Backbone injection.}
Let \(g_j\in\mathbb{R}^{5120}\) denote the adapter output for video token \(j\), and let \(e(s)\) denote the embedded flow timestep. We form the token-specific timestep embedding
\begin{equation}
\widetilde{e}_j(s)
=
e(s)+g_j.
\label{eq:app-redar-injection}
\end{equation}
The backbone's existing \(\operatorname{time\_proj}\) maps this embedding to six modulation vectors per token: the shift, scale, and gate applied to the attention and FFN sublayers. The same token-specific embedding is used in all \(40\) transformer blocks and determines the shift and scale of the final output normalization. Thus, rendering-enriched actions enters exclusively through the pretrained timestep-modulation pathway and does not modify the backbone's attention or FFN architecture.

During training, we set the adapter contribution \(G\) to zero with probability \(0.1\), while retaining the initial-frame conditioning. This conditioning dropout regularizes the adapter and provides an action-unconditioned branch for evaluation controls.

\subsection{Baseline Re-implementation}
\label{app:baselineimp}

\paragraph{Action-only baseline.}
Our action-only baseline follows the conditioning mechanism of EgoWM~\citep{bagchi2026egowm}. Starting from the same Wan~2.1 image-to-video backbone used by \method{}, we remove the rendering encoder and render--action fusion modules entirely. The numerical robot configuration at each timestep is normalized using the same per-embodiment statistics as \method{}, projected with an MLP followed by a 1D temporal convolution, and injected through the backbone's timestep-modulation pathway. Thus, the baseline receives the same numerical robot trajectory as \method{} but no camera-aligned robot renderings.

All remaining components are kept identical to the corresponding \method{} configuration, including backbone initialization, training clips, clip construction, optimization schedule, conditioning dropout, and training budget. All models are evaluated on exactly the same validation and test splits. The complete single-embodiment training recipe is given in Table~\ref{tab:training-recipe-single}. This baseline therefore provides a controlled comparison between numerical action-only conditioning and \method{} conditioning.

\paragraph{OSCAR.}
For OSCAR~\citep{oscar2026}, we retain the released architecture and its 2D kinematic-skeleton conditioning rather than modifying its conditioning formulation. We initialize from the released OSCAR checkpoint and fine-tune a separate model for each of \droid{}, \bridge{}, and \robomind{} using the corresponding training demonstrations and selected start
positions, with OSCAR's 81-frame temporal windows. We do not report zero-shot results on \bridge{} or \robomind{}, since these datasets are not part of OSCAR's original training corpus.

To align OSCAR more closely with our evaluation setting, we fine-tune it at the same $432\times240$ resolution used by \method{} and use an empty text prompt throughout training and evaluation. We preserve its native skeleton-conditioning mechanism and otherwise follow the fine-tuning configuration summarized in Table~\ref{tab:oscar-recipe}. All models are evaluated on the same validation and test clips. Because OSCAR differs from \method{} in backbone, pretraining data, model size, conditioning representation, and optimization history, we do not attempt to equalize architecture-level compute and treat OSCAR as a system-level baseline rather than a controlled ablation of \method{}.

\begin{table}[t]
\centering
\caption{
OSCAR configurations used in our experiments.
The fine-tuned models are initialized from the released OSCAR checkpoint~\citep{oscar2026}.
}
\label{tab:oscar-recipe}
\small
\setlength{\tabcolsep}{8pt}
\renewcommand{\arraystretch}{1.35}

\begin{tabularx}{\linewidth}{
@{}l
>{\raggedright\arraybackslash}X
>{\raggedright\arraybackslash}X@{}}
\toprule
& \textbf{Released OSCAR} & \textbf{OSCAR (Fine-tuned)} \\
\midrule

Initialization
& Cosmos-Predict2.5~\citep{nvidia2025worldsim}, 2B
& Released OSCAR checkpoint \\

Conditioning
& 2D skeleton + text
& 2D skeleton + empty text prompt \\

Training data
& \num{180657} episodes:
\num{94830} robot episodes
(RH20T~\citep{fang2024rh20t}, InternData-A1~\citep{tian2025interndataa1}, DROID~\citep{droid2024}~\num{21904},
AgiBot-Beta~\citep{agibot2025colosseo}, AIROA-MoMa~\citep{takanami2025airoamoma})
and \num{85827} human episodes
(EgoDex~\citep{hoque2026egodex}, EPIC-Kitchens~\citep{damen2022rescaling})
& Our training split (Table~\ref{tab:data}),
using 81-frame windows at the selected start positions, with overlapping windows permitted\\

Clip length
& 81 frames, native frame rate
& 81 frames, native frame rate \\

Resolution
& $640\times480$
& $432\times240$ \\

Optimizer
& AdamW
& AdamW, $\beta=(0.9,0.99)$,
weight decay $10^{-3}$ \\

Learning rate
& $3\times10^{-5}$
& $3\times10^{-5}$ with 500-step warmup
and cosine decay to $3\times10^{-6}$ \\

Batch size
& 16
& 16 \\

Training steps
& 15k robot-only steps, followed by robot + human training
& \num{10000} \\

Hardware
& $1\times$ GH200
& $8\times$ A100 \\

Sampling
& 35 steps, guidance scale 6
& 35 steps, guidance scale 6 \\

\bottomrule
\end{tabularx}
\end{table}

\subsection{Dataset Visualizations}
\label{app:datasetviz}

\begin{figure*}[t]
    \centering
    \includegraphics[width=\textwidth]{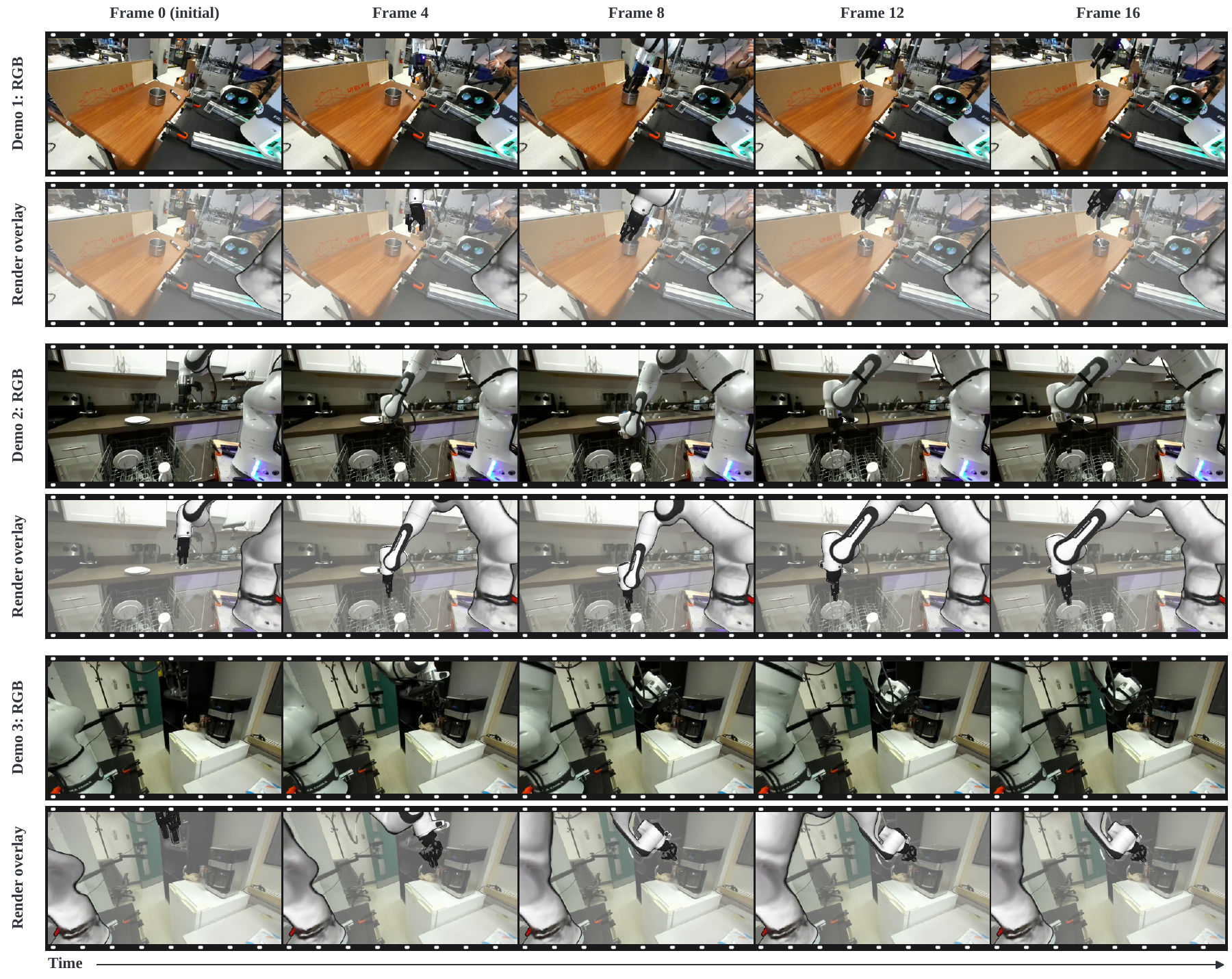}
    \caption{
    Representative examples from \droid{} with corresponding
    camera-aligned robot renderings.
    }
    \label{fig:droid-visualization}
\end{figure*}

\begin{figure*}[t]
    \centering
    \includegraphics[width=\textwidth]{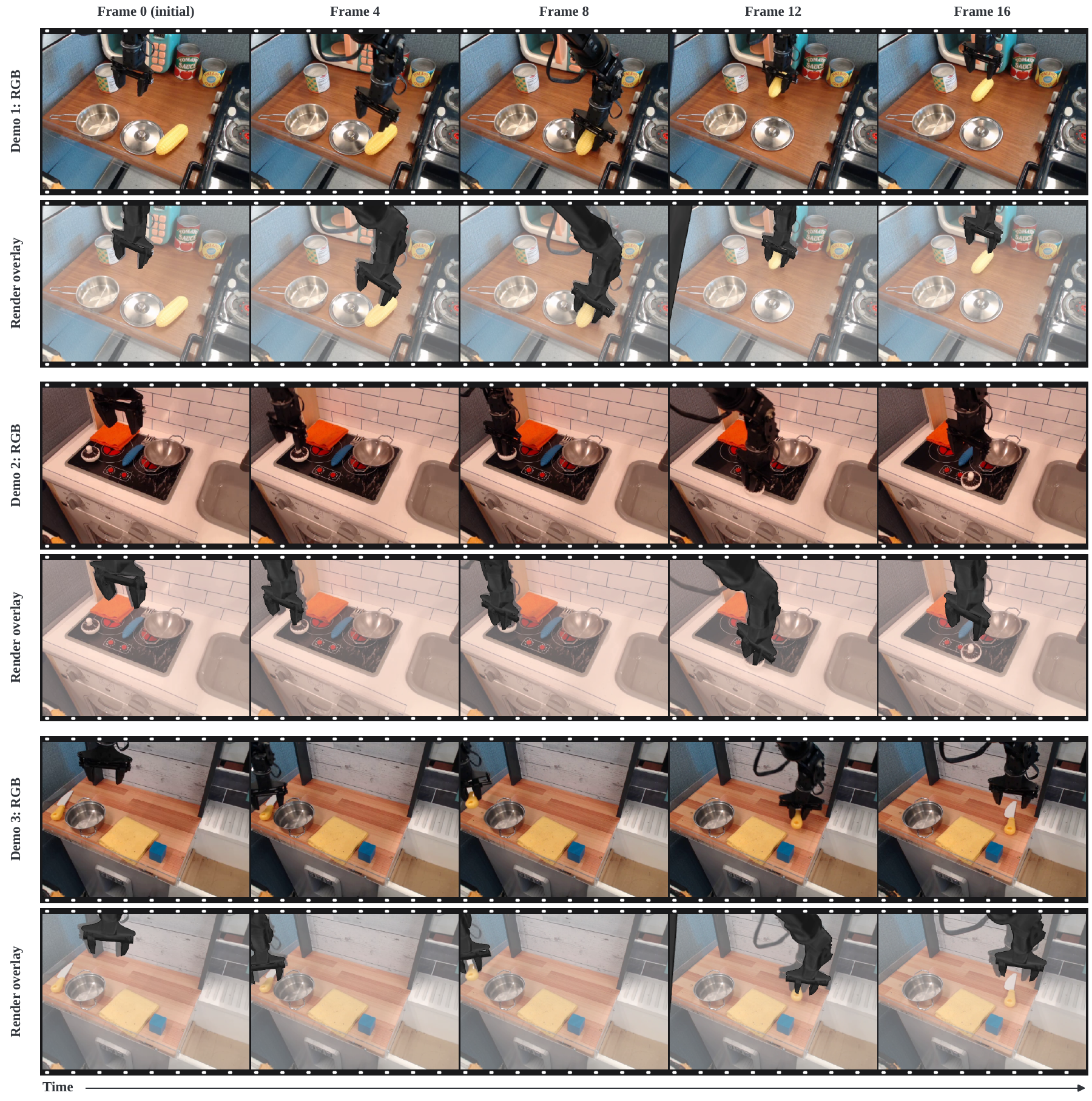}
    \caption{
    Representative examples from \bridge{} with corresponding
    camera-aligned robot renderings.
    }
    \label{fig:bridge-visualization}
\end{figure*}

\begin{figure*}[t]
    \centering
    \includegraphics[width=\textwidth]{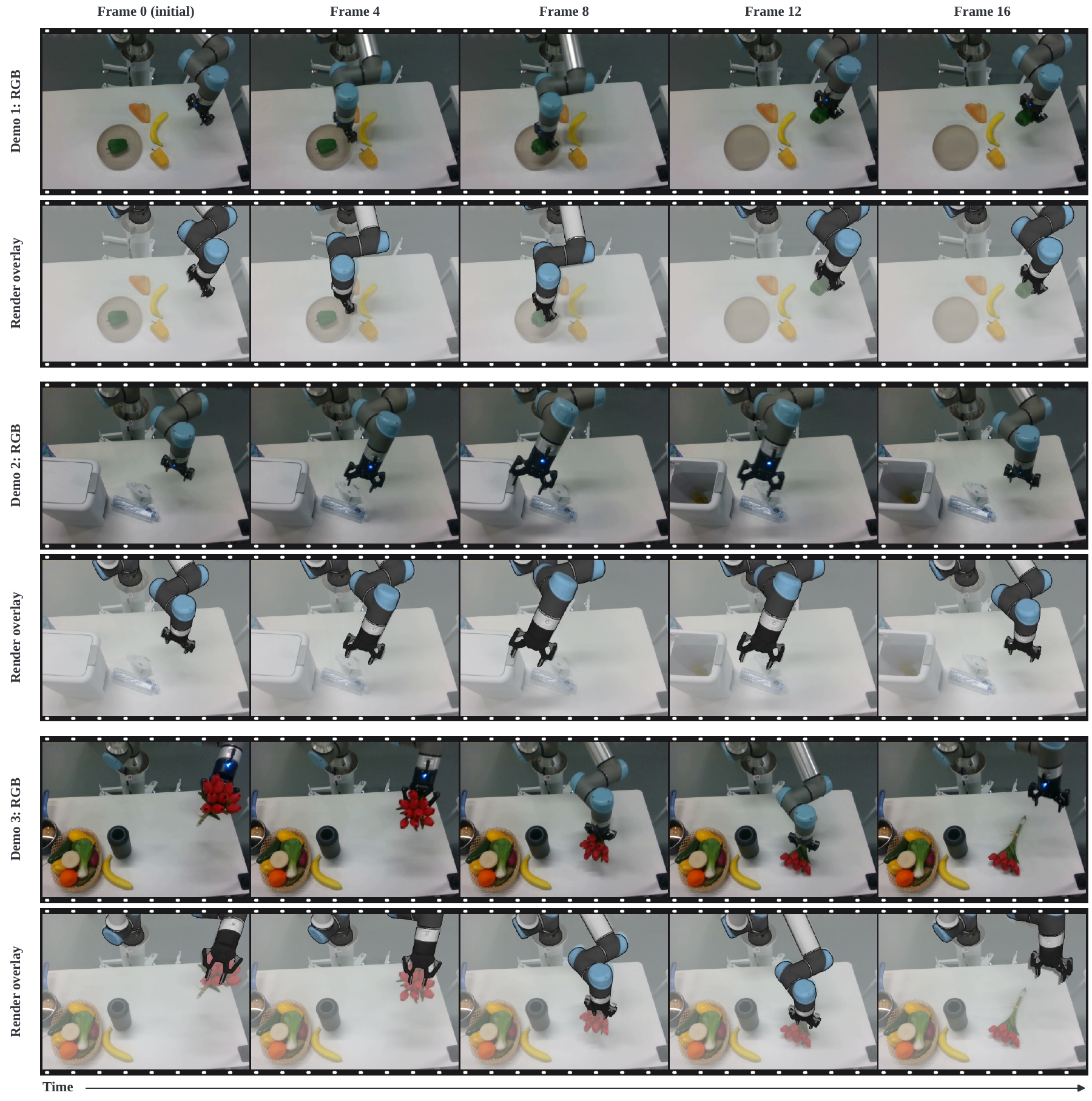}
    \caption{
    Representative examples from \robomind{} with corresponding
    camera-aligned robot renderings.
    }
    \label{fig:robomind-visualization}
\end{figure*}

\begin{figure*}[t]
    \centering
    \includegraphics[width=\textwidth]{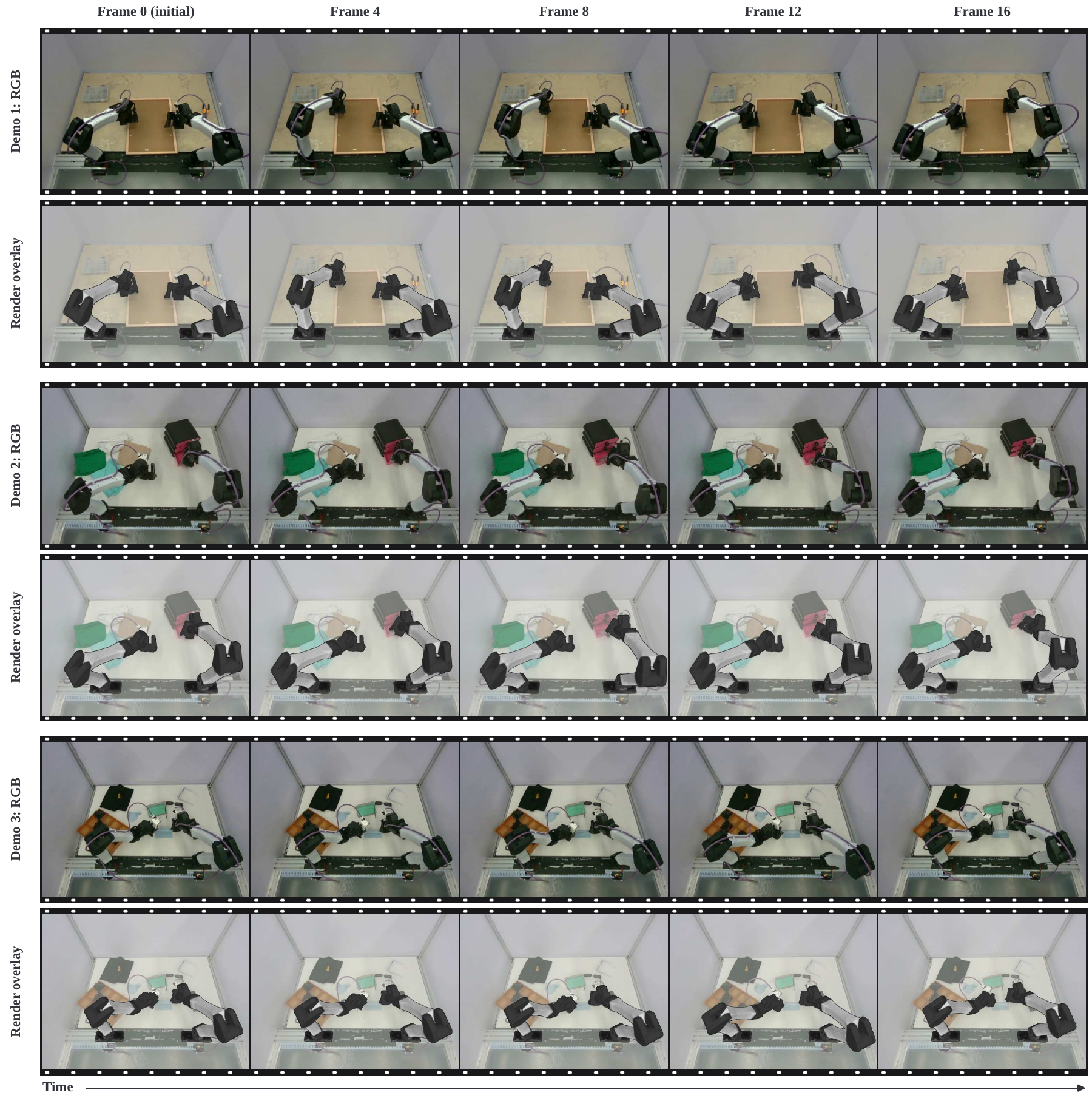}
    \caption{
    Representative examples from ABC (bimanual YAM) with corresponding
    camera-aligned renderings.
    }
    \label{fig:abc-visualization}
\end{figure*}

\begin{figure*}[t]
    \centering
    \includegraphics[width=\textwidth]{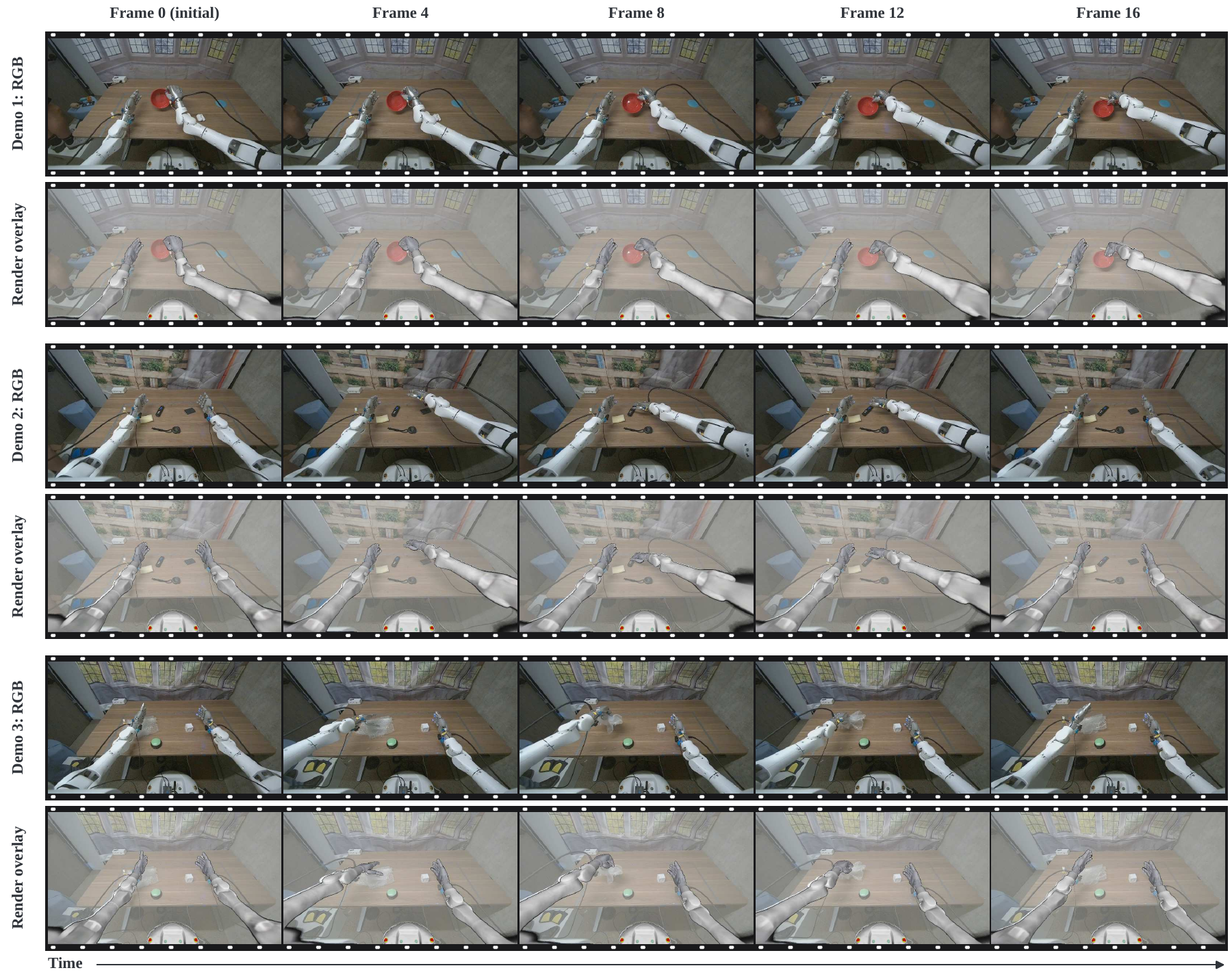}
    \caption{
    Representative examples from T-Rex (humanoid with dexterous hands) with corresponding
    camera-aligned renderings.
    }
    \label{fig:trex-visualization}
\end{figure*}

\begin{figure*}[t]
    \centering
    \includegraphics[width=\textwidth]{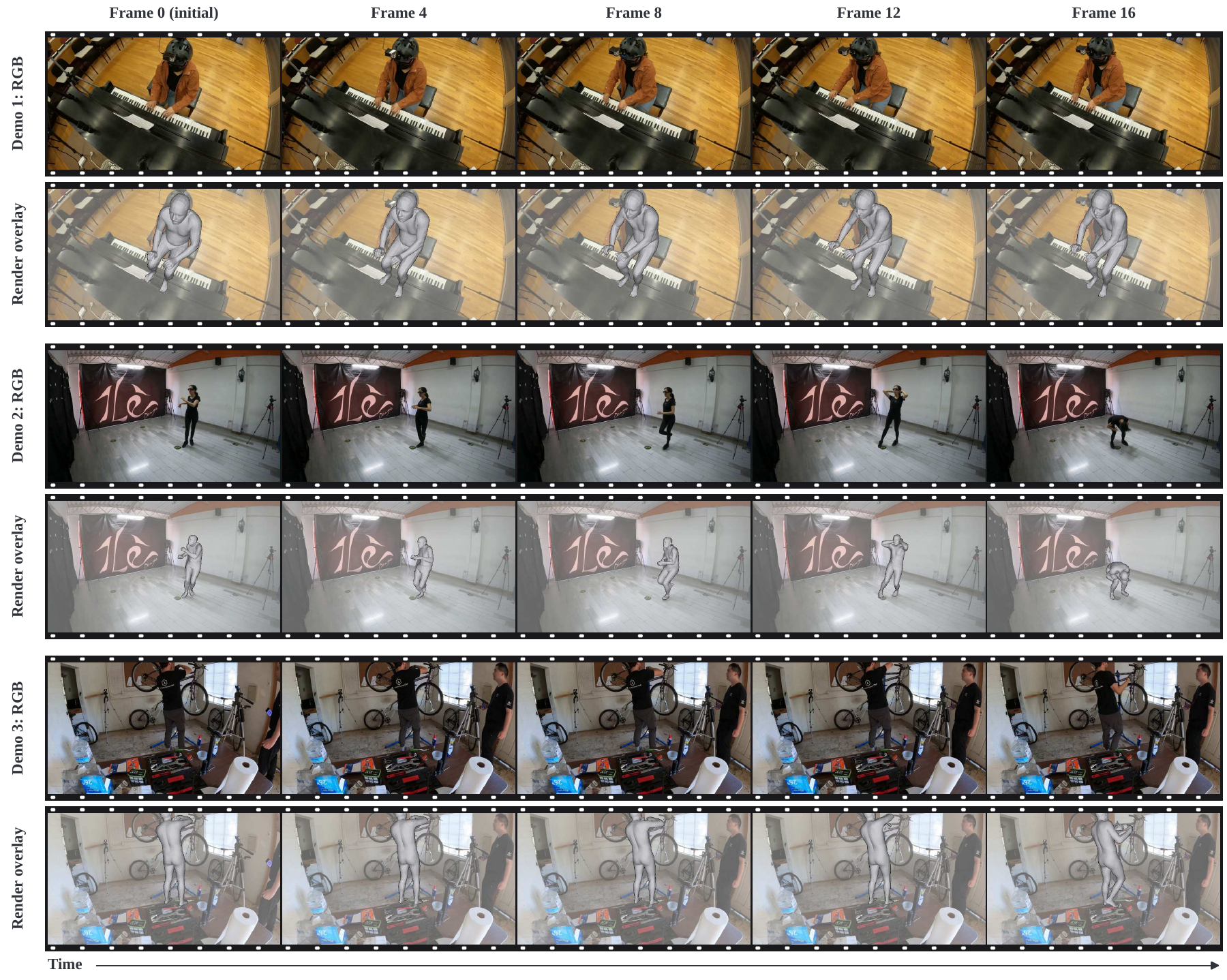}
    \caption{
    Representative examples from Ego-Exo4D (human body) with corresponding
    camera-aligned renderings.
    }
    \label{fig:egoexo4d-visualization}
\end{figure*}

\begin{figure*}[t]
    \centering
    \includegraphics[width=\textwidth]{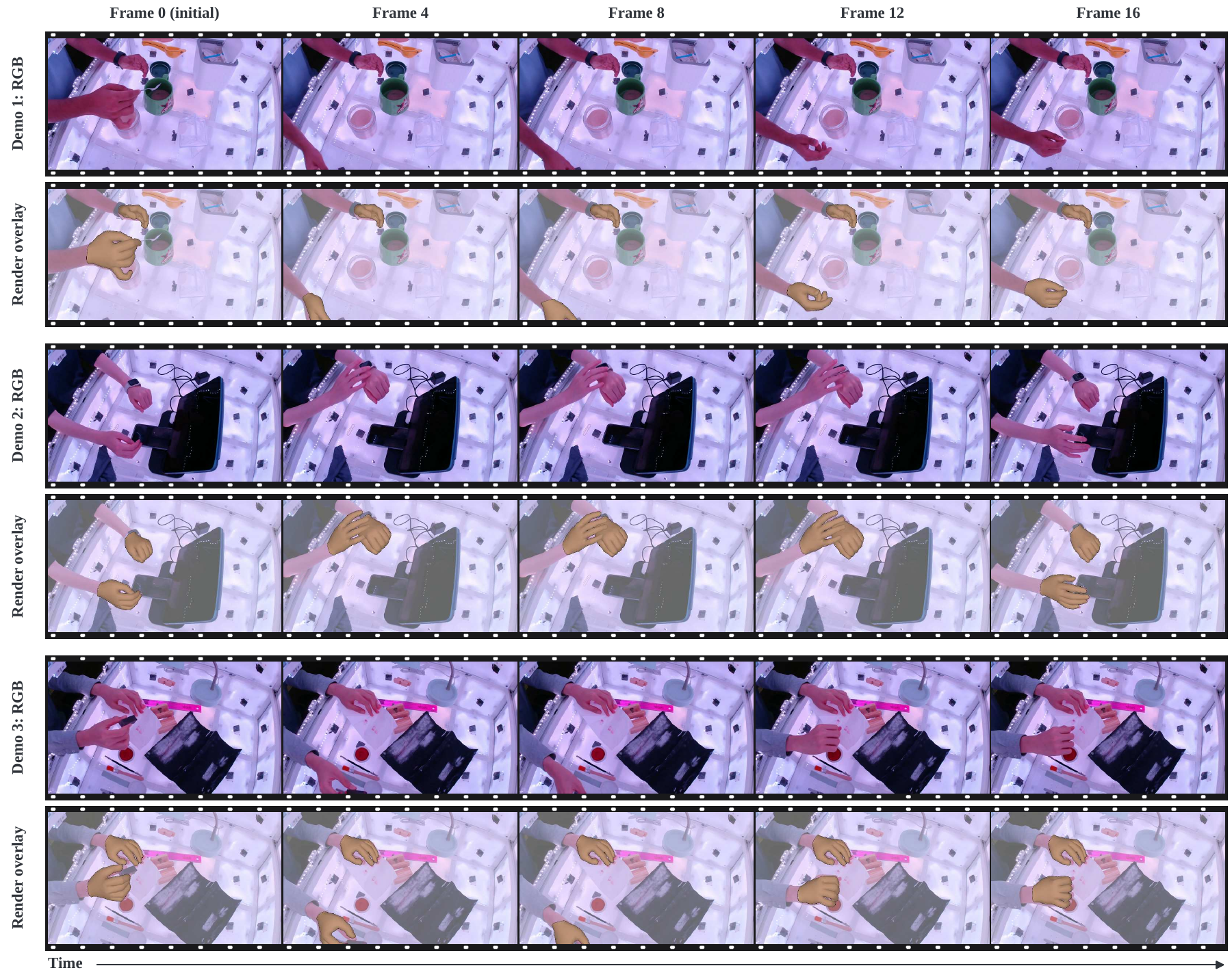}
    \caption{
    Representative examples from GigaHands (human hands) with corresponding
    camera-aligned renderings.
    }
    \label{fig:gigahands-visualization}
\end{figure*}

\begin{figure*}[t]
    \centering
    \includegraphics[width=\textwidth]{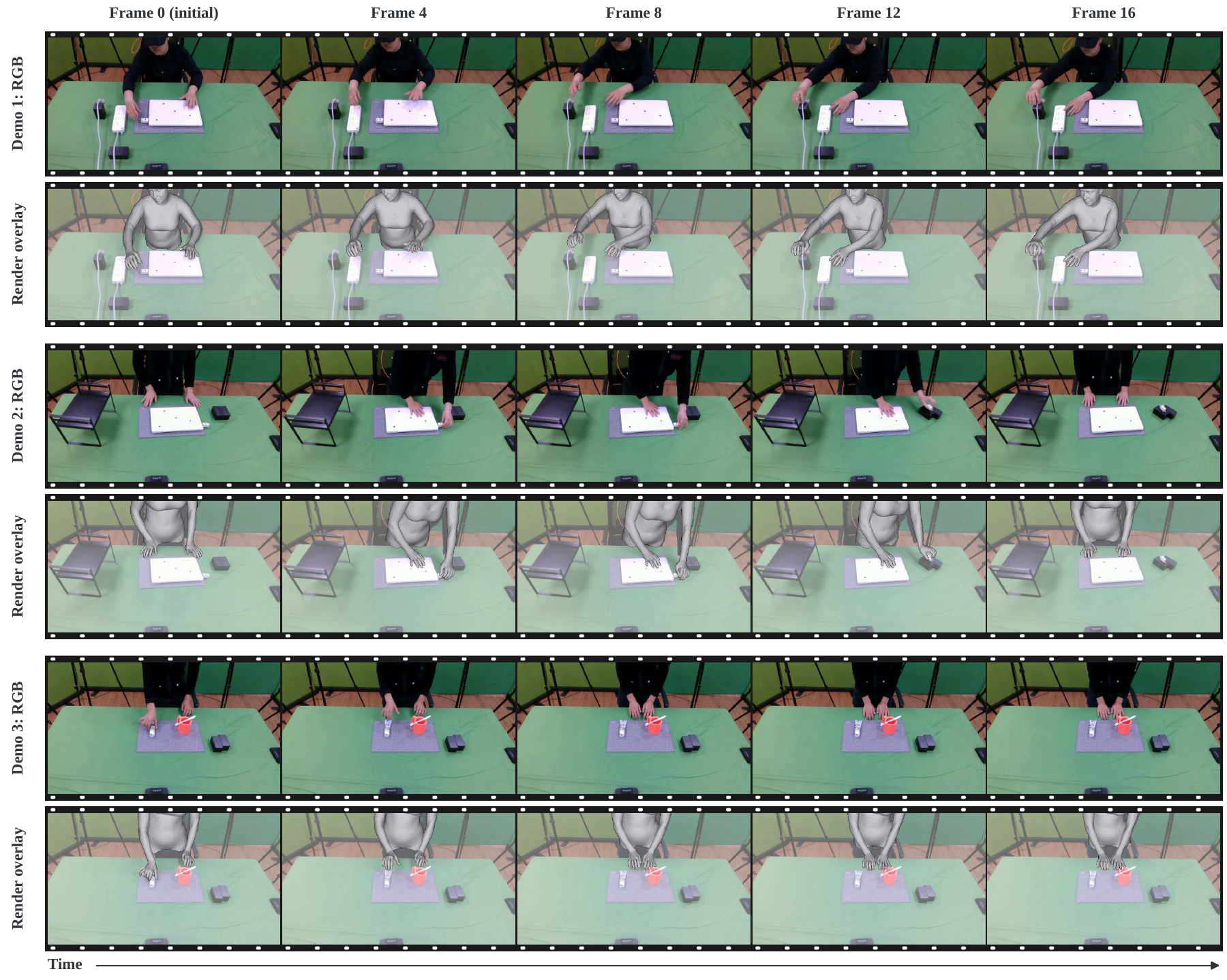}
    \caption{
    Representative examples from OakInk2 (human upper body) with corresponding
    camera-aligned renderings.
    }
    \label{fig:oakink2-visualization}
\end{figure*}

We show representative RGB observations and corresponding camera-aligned
robot renderings from \droid{}, \bridge{}, and \robomind{} in
Figures~\ref{fig:droid-visualization},
\ref{fig:bridge-visualization}, and
\ref{fig:robomind-visualization}, respectively, and from the five added
datasets in Figures~\ref{fig:abc-visualization}--\ref{fig:oakink2-visualization}.
These examples illustrate the diversity of embodiments, environments, and
viewpoints across the eight datasets.

\subsection{Implementation Details}
\label{app:implementation}

We provide the training configurations for all our models in  Tables~\ref{tab:training-recipe-single}, ~\ref{tab:training-recipe-multi-single} and ~\ref{tab:training-recipe-multi}. To ensure that the multi-embodiment model observes a comparable number of draws per dataset as the corresponding single-dataset action-only and baseline runs, we use multi-node training with a larger global batch size. Unless otherwise noted, the multi-embodiment models use the same architecture and optimization settings as the single-embodiment models.

\begin{table*}[t]
  \centering
  \caption{Training configuration for the single-embodiment models.}
  \label{tab:training-recipe-single}
  \footnotesize
  \renewcommand{\arraystretch}{1.08}
  \setlength{\tabcolsep}{5pt}
  \begin{tabularx}{\textwidth}{
    @{}
    >{\bfseries\raggedright\arraybackslash}p{0.18\textwidth}
    >{\raggedright\arraybackslash}X
    @{}
  }
    \toprule

    Backbone
      & Wan 2.1 I2V 14B (16.4B parameters). We train all 40 transformer
        blocks together with the output and timestep projections,
        yielding 16.3B trainable backbone parameters. \\

    Adapters
      & Render--action adapter: 30.5M parameters; action-only adapter:
        4.2M parameters. \\

    Data
      & One dataset per model:
        see Table~\ref{tab:data}. \\

    Input
      & 17 frames at \(240\times432\). Contiguous windows use \(0.2\) s
        temporal spacing, while full-span windows cover the complete
        demonstration. The text prompt is always empty. \\

    Optimizer
      & AdamW~\citep{loshchilov2019adamw} with weight decay \(10^{-3}\) and gradient clipping at \(1.0\).
        Training uses bf16, fp32 master weights, FSDP, and activation
        checkpointing. \\

    Learning rates
      & \(5\times10^{-6}\) for the backbone and \(2\times10^{-5}\) for
        the conditioning adapter. \\

    Schedule
      & Cosine decay with 500 warm-up steps and a minimum learning-rate
        ratio of \(0.05\). \\

    Batch size
      & 8 examples per GPU on 8 H100-80GB GPUs, giving a global batch
        size of 64. \\


    Weight EMA
      & Decay of \(0.9975\), updated every 10 steps after a 500-step
        warm-up.  \\

    Conditioning dropout
      & Probability \(0.1\) per training step. \\

    Objective
      & Latent flow matching~\citep{lipman2023flowmatching}. \\

    \bottomrule
  \end{tabularx}
\end{table*}

\begin{table*}[t]
  \centering
  \caption{
    Training configuration for the single-node multi-embodiment model.
    Settings not listed are identical to
    Table~\ref{tab:training-recipe-single}.
  }
  \label{tab:training-recipe-multi-single}
  \footnotesize
  \renewcommand{\arraystretch}{1.08}
  \setlength{\tabcolsep}{5pt}
  \begin{tabularx}{\textwidth}{
    @{}
    >{\bfseries\raggedright\arraybackslash}p{0.18\textwidth}
    >{\raggedright\arraybackslash}X
    @{}
  }
    \toprule

    Data
      & Joint training on \droid{}, \robomind{}, and \bridge{}.
        Datasets are sampled with equal probability, and clips are
        sampled uniformly within each dataset. \\

    Hardware
      & 8 H100-80GB GPUs on one node, with one FSDP rank per GPU. \\

    Batch size
      & 8 examples per GPU, giving a global batch size of 64. \\

    Learning rates
      & Backbone: \(5\times10^{-6}\); adapter: \(2\times10^{-5}\). \\

    Schedule
      & Cosine decay with 500 warm-up steps and a minimum learning-rate
        ratio of \(0.05\). \\

    EMA
      & Decay of \(0.9975\), updated every 10 steps after a 500-step
        warm-up. \\

    \bottomrule
  \end{tabularx}
\end{table*}

\begin{table*}[t]
  \centering
  \caption{
    Training configuration for the four-node multi-embodiment render--action model. Settings not listed are identical to
    Table~\ref{tab:training-recipe-single}.
  }
  \label{tab:training-recipe-multi}
  \footnotesize
  \renewcommand{\arraystretch}{1.08}
  \setlength{\tabcolsep}{5pt}
  \begin{tabularx}{\textwidth}{
    @{}
    >{\bfseries\raggedright\arraybackslash}p{0.18\textwidth}
    >{\raggedright\arraybackslash}X
    @{}
  }
    \toprule

    Data
      & Joint training on \droid{}, \robomind{}, and \bridge{}.
        Datasets are sampled with equal probability, and clips are
        sampled uniformly within each dataset. \\

    Hardware
      & 32 H100-80GB GPUs across four nodes, with one FSDP rank per GPU. \\

    Batch size
      & 8 examples per GPU, giving a global batch size of 256. \\

    Learning rates
      & Backbone: \(1\times10^{-5}\); adapter: \(4\times10^{-5}\). \\

    Schedule
      & Cosine decay with 250 warm-up steps and a minimum learning-rate
        ratio of \(0.05\). \\

    Weight EMA
      & Decay of \(0.995\), updated every 10 steps after a 250-step
        warm-up. \\

    \bottomrule
  \end{tabularx}
\end{table*}

\begin{table*}[t]
  \centering
  \caption{
    Training configuration for \methodeight{}. Settings not listed are identical to
    Table~\ref{tab:training-recipe-multi}.
  }
  \label{tab:training-recipe-emb8}
  \footnotesize
  \renewcommand{\arraystretch}{1.08}
  \setlength{\tabcolsep}{5pt}
  \begin{tabularx}{\textwidth}{
    @{}
    >{\bfseries\raggedright\arraybackslash}p{0.18\textwidth}
    >{\raggedright\arraybackslash}X
    @{}
  }
    \toprule
    Data
      & Joint training on the eight datasets of Table~\ref{tab:data} (803,602 training clips, 593 hours of training video, Table~\ref{tab:corpus-hours}).
        Datasets are sampled in proportion to their training clips, and clips uniformly within each dataset. \\
    Actions
      & Every action vector is zero-padded to 159 dimensions and passed to one shared action encoder; there is no
        embodiment label or embodiment-specific parameter. \\
    Hardware
      & 32 GPUs across four nodes, one FSDP rank per GPU: H100-80GB. \\
    \bottomrule
  \end{tabularx}
\end{table*}

\subsection{Evaluation Metrics}
\label{app:metrics}

We evaluate generated videos against the reference videos at
\(240\times432\) resolution. All metrics exclude the conditioning frame
and are computed over the sixteen predicted frames. Unless otherwise
specified, scores are first averaged over time within each clip and then
averaged across clips with equal weight. Undefined scores are omitted.

\paragraph{Whole-frame visual fidelity.}
We report peak signal-to-noise ratio (PSNR), structural similarity
(SSIM)~\citep{wang2004ssim}, and learned perceptual image patch similarity (LPIPS).
PSNR measures pixel-level reconstruction fidelity, SSIM measures local
agreement in luminance, contrast, and structure, and LPIPS measures
perceptual discrepancy using pretrained AlexNet features. Each metric is
computed independently for every predicted frame and averaged over the
prediction horizon. Higher PSNR and SSIM are better, while lower LPIPS
is better.

\paragraph{Region-specific reconstruction.}
To separate robot reconstruction from changes in the surrounding scene,
we compute PSNR over the robot and moving non-robot objects. We initialize the robot mask from the first rendered frame and
obtain non-robot object masks using SAM~2 on the first reference frame.
These masks are propagated through the reference video to define the
evaluation regions.

A non-robot object is classified as moving when its maximum centroid
displacement, normalized by the square root of its initial mask area, is
at least \(0.25\). We compute
\(\mathrm{PSNR}_{\mathrm{arm}}\) over the reference robot mask and
\(\mathrm{PSNR}_{\mathrm{obj}}\) over moving non-robot object masks.
For object metrics, scores are first averaged over time for each object
and then averaged equally across objects.

\paragraph{Structural consistency.}
Structural consistency score (SCS) measures agreement between predicted
and reference region trajectories. Starting from identical first-frame
masks, SAM~2 independently propagates each region through the generated
and reference videos. We compute the intersection-over-union between
corresponding masks at each predicted frame and average over time and
regions. Higher SCS indicates greater consistency in region position and
shape.

In the whole-frame results table, SCS is averaged over all non-robot
objects. In the region-specific results,
\(\mathrm{SCS}_{\mathrm{arm}}\) evaluates the robot and
\(\mathrm{SCS}_{\mathrm{obj}}\) evaluates moving non-robot objects. The training-budget comparisons use SCS averaged over all tracked non-robot objects, as in the Tab. \ref{tab:overall-results}.

\subsection{Camera extrinsics dataset}
\label{app:camera_extrinsics}
\begin{figure}[t]
\centering
\includegraphics[width=\linewidth]{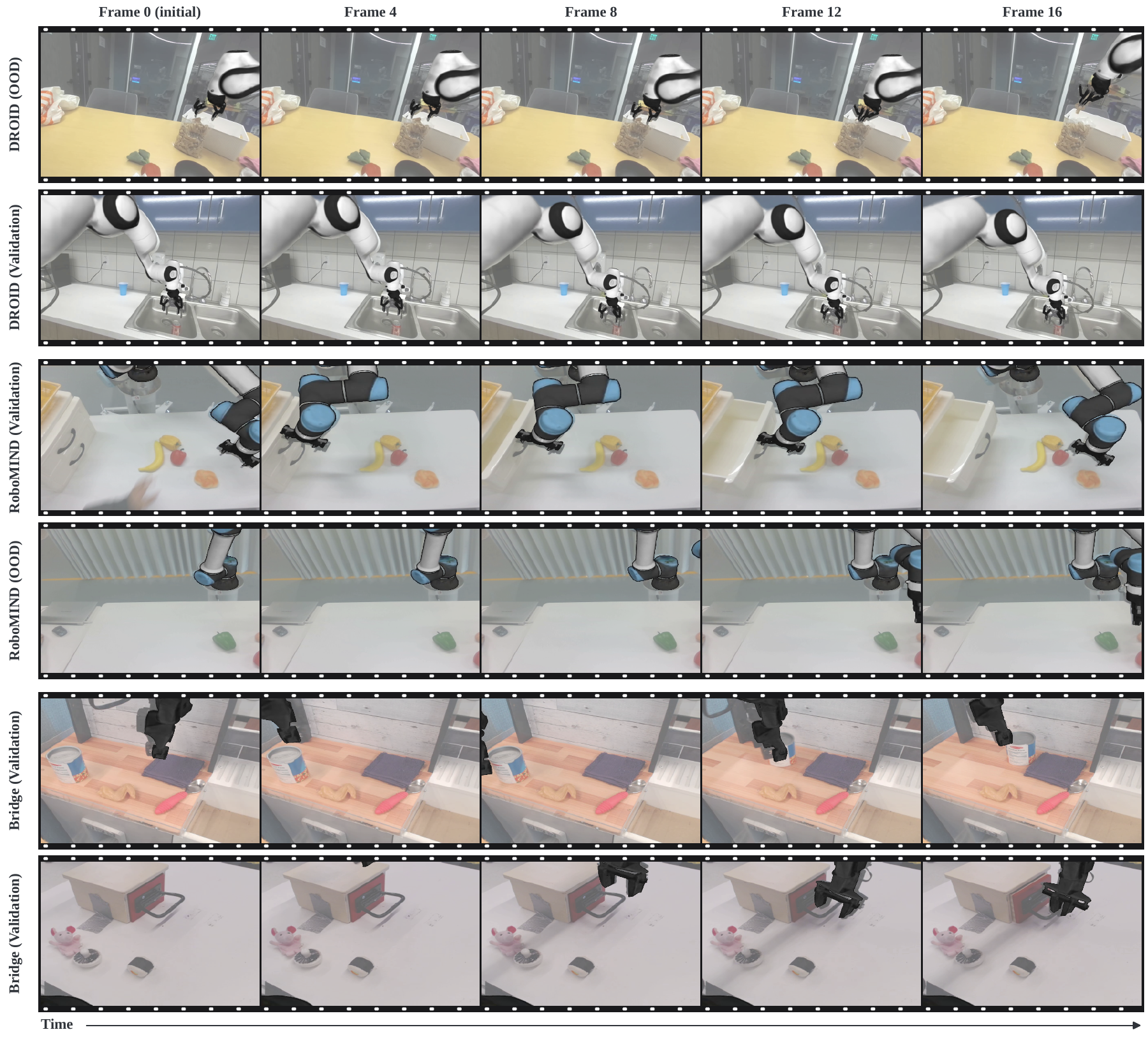}
\vspace{-0.5em}
\caption{
Overlays of camera calibration results.
}
\label{app:fig:camera_overlay}

\end{figure}

We release the camera extrinsics and intrinsics we recover for 45,345 demonstrations: \bridge{} 14,824, \robomind{} 17,588, and ABC 12,933. The other five datasets ship their own cameras, which we use and do not re-release; for GigaHands we refine the released cameras per recording session. \droid{} extrinsics and ZED intrinsics are used as provided by \citet{droid2024}. \robomind{} and \bridge{} ship no extrinsics, so we treat each camera as fixed within a recording session and recover one pose per session with the procedure of Sec.~\ref{sec:method-calib}: 49 poses for \robomind{} and 56 for \bridge{}. Intrinsics are the RealSense D435i values for \robomind{}~\citep{wu2024robomind} and MoGe-estimated~\citep{Wang_2025_CVPR} focal lengths for \bridge{}. We show sample camera-calibration overlays in Fig. \ref{app:fig:camera_overlay}. ABC's top-down cameras also ship without extrinsics and move between recordings, so we calibrate each ABC video separately by fitting the camera pose to segmentation masks of the two YAM arms and to photometric agreement with the rendering. We fit 13,518 videos, of which 12,933 (95.7\%) pass our alignment check.

\subsection{Results}
\label{app:results}

\subsubsection{Additional Qualitative Results}
\label{app:AdditionalQualitative}

\begin{figure}[t]
    \centering
    \includegraphics[width=\linewidth]{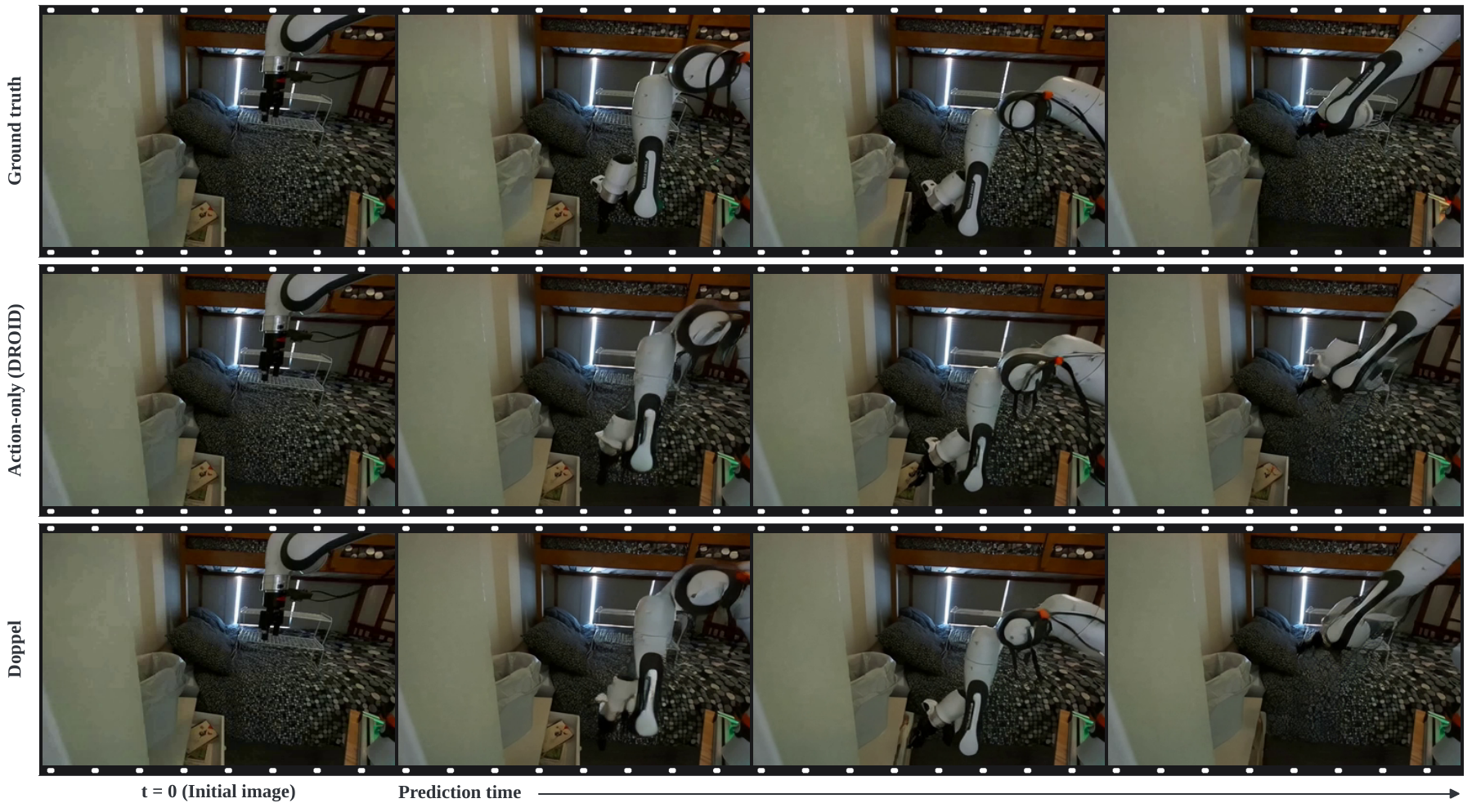}
    \caption{
    Additional trajectory-generalization example on \droid{}.
    We compare \method{} with the action-only baseline on a held-out trajectory
    from a familiar environment.
    }
    \label{fig:additional-droid-trajectory}
\end{figure}

We provide additional qualitative results across all our experiments. Please see \ref{app:website} for the complete set of results. We also include an example of trajectory generalization, comparing \method{} with the baseline in Fig.~\ref{fig:additional-droid-trajectory}.

\subsubsection{From three to eight embodiments}
\label{app:three-to-eight}
Table~\ref{tab:three-to-eight} compares \methodthree{} with \methodeight{} on the same clips. Both models see a similar amount of each robot dataset (Table~\ref{tab:overall-results}); \methodeight{} additionally trains on the five other datasets. Table~\ref{tab:matched-51} repeats the comparison of Table~\ref{tab:overall-results} at matched training exposure, with every model at its saved checkpoint closest to 51 passes over each robot dataset. Table~\ref{tab:gap-over-training} follows the same comparison over training.

\begin{table}[t]
  \centering
  \caption{\methodthree{} versus \methodeight{}, paired on the same 200 clips per set (final checkpoints). $\Delta$ is \methodeight{} minus \methodthree{} with its standard error; \emph{Better} is the share of clips where \methodeight{} has higher PSNR; arm and moving-object $\Delta$ use the region metrics of Appendix~\ref{app:metrics}.}
  \label{tab:three-to-eight}
  \small
  \setlength{\tabcolsep}{4pt}
  \resizebox{\columnwidth}{!}{%
  \begin{tabular}{@{}llrrrrrr@{}}
    \toprule
    Dataset & Split & \methodthree{} & \methodeight{} & $\Delta$ PSNR & Better & $\Delta$ arm & $\Delta$ moving obj. \\
    \midrule
    \droid{} & Val. & 24.30 & 24.82 & $+0.52 \pm 0.03$ & 92\% & $+0.70$ & $+0.69$ \\
     & OOD & 21.25 & 21.46 & $+0.21 \pm 0.03$ & 73\% & $+0.34$ & $+0.13$ \\
    \bridge{} & Val. & 26.55 & 26.54 & $-0.01 \pm 0.06$ & 57\% & $+0.09$ & $+0.12$ \\
     & OOD & 22.47 & 22.24 & $-0.24 \pm 0.05$ & 38\% & $-0.01$ & $-0.08$ \\
    \robomind{} & Val. & 31.07 & 31.30 & $+0.24 \pm 0.04$ & 75\% & $+0.42$ & $+0.14$ \\
     & OOD & 30.44 & 30.66 & $+0.22 \pm 0.04$ & 72\% & $+0.34$ & $+0.16$ \\
    \bottomrule
  \end{tabular}%
  }
\end{table}

\begin{table}[t]
  \centering
  \caption{PSNR of \methodeight{} minus \methodthree{}, on the same 200 clips per set, at three points in training.}
  \label{tab:gap-over-training}
  \small
  \setlength{\tabcolsep}{5pt}
  \begin{tabular}{@{}llrrrr@{}}
    \toprule
    & & \methodthree{} & \multicolumn{3}{c}{$\Delta$ PSNR (\methodeight{} $-$ \methodthree{})} \\
    \cmidrule(lr){4-6}
    Dataset & Split & passes & ${\approx}30$ passes & 51 passes & final \\
    \midrule
    \droid{} & Val. & 24.8 & $+0.70$ & $+0.45$ & $+0.52$ \\
     & OOD & 24.8 & $+0.51$ & $+0.17$ & $+0.21$ \\
    \midrule
    \bridge{} & Val. & 33.4 & $+0.25$ & $+0.13$ & $-0.01$ \\
     & OOD & 33.4 & $-0.10$ & $-0.05$ & $-0.24$ \\
    \midrule
    \robomind{} & Val. & 27.9 & $+0.52$ & $+0.21$ & $+0.24$ \\
     & OOD & 27.9 & $+0.25$ & $+0.13$ & $+0.22$ \\
    \bottomrule
  \end{tabular}
\end{table}

\begin{table*}[t]
\centering
\caption{Prediction quality on the three robot datasets at matched training exposure: every model is evaluated at its saved checkpoint closest to 51 passes over each robot dataset.}
\label{tab:matched-51}
\small
\setlength{\tabcolsep}{4pt}
\resizebox{\textwidth}{!}{%
\begin{tabular}{@{}llllrrrrrrrr@{}}
\toprule
& & & & \multicolumn{4}{c}{Validation: unseen trajectories} & \multicolumn{4}{c}{OOD test} \\
\cmidrule(lr){5-8} \cmidrule(lr){9-12}
Dataset & Method & Input & Trained on & PSNR \(\uparrow\) & SSIM \(\uparrow\) & LPIPS \(\downarrow\) & SCS \(\uparrow\) & PSNR \(\uparrow\) & SSIM \(\uparrow\) & LPIPS \(\downarrow\) & SCS \(\uparrow\) \\
\midrule
\droid{} & Action-only & Actions & This dataset & 22.847 & 0.841 & 0.079 & 0.927 & 20.047 & 0.780 & 0.131 & 0.873 \\
 & \method{} & Render + actions & This dataset & \underline{24.322} & \underline{0.864} & \underline{0.065} & \underline{0.940} & \underline{21.278} & 0.802 & 0.113 & \textbf{0.903} \\
 & \methodthree{} & Render + actions & 3 robot datasets & 24.297 & 0.862 & 0.066 & \underline{0.940} & 21.250 & \underline{0.803} & \underline{0.111} & 0.899 \\
 & \methodeight{} & Render + actions & All 8 datasets & \textbf{24.749} & \textbf{0.869} & \textbf{0.063} & \textbf{0.943} & \textbf{21.421} & \textbf{0.805} & \textbf{0.109} & \underline{0.901} \\
\midrule
\bridge{} & Action-only & Actions & This dataset & 25.625 & 0.864 & 0.054 & 0.941 & 20.665 & 0.756 & 0.129 & 0.869 \\
 & \method{} & Render + actions & This dataset & 26.311 & \underline{0.871} & \underline{0.050} & \underline{0.947} & 21.824 & 0.773 & \underline{0.109} & \textbf{0.905} \\
 & \methodthree{} & Render + actions & 3 robot datasets & \underline{26.389} & \underline{0.871} & \underline{0.050} & \underline{0.947} & \textbf{22.348} & \underline{0.784} & \textbf{0.099} & \textbf{0.905} \\
 & \methodeight{} & Render + actions & All 8 datasets & \textbf{26.521} & \textbf{0.873} & \textbf{0.049} & \textbf{0.948} & \underline{22.295} & \textbf{0.785} & \textbf{0.099} & \underline{0.900} \\
\midrule
\robomind{} & Action-only & Actions & This dataset & 29.992 & 0.919 & 0.040 & 0.938 & 29.805 & 0.915 & 0.040 & 0.930 \\
 & \method{} & Render + actions & This dataset & 31.007 & \underline{0.928} & \textbf{0.037} & \textbf{0.942} & 30.387 & \underline{0.920} & \underline{0.038} & 0.932 \\
 & \methodthree{} & Render + actions & 3 robot datasets & \underline{31.013} & 0.927 & \underline{0.038} & \underline{0.940} & \underline{30.530} & \underline{0.920} & \textbf{0.037} & \textbf{0.939} \\
 & \methodeight{} & Render + actions & All 8 datasets & \textbf{31.221} & \textbf{0.929} & \textbf{0.037} & \textbf{0.942} & \textbf{30.662} & \textbf{0.921} & \underline{0.038} & \underline{0.936} \\
\bottomrule
\end{tabular}%
}
\end{table*}

\subsubsection{Numerical results for the training curves}
\label{app:scaling-numbers}

We report the numerical values underlying both training-curve comparisons.
Each entry averages 200 held-out clips for the corresponding dataset and
split. PSNR excludes the conditioning frame and averages over the 16
predicted frames before averaging across clips; SCS averages all tracked
non-robot objects. All values are reported to three decimal places.
Training draws include repeated examples. The two comparisons use
different budget conventions, explicitly indicated in the tables.
The uncertainty bars in the figures use 5,000 source-episode bootstrap
resamples; the tables below report the corresponding point estimates.

\begin{table*}[t]
\centering
\caption{
Prediction quality by image region on the three robot datasets, for the same models, checkpoints and clips as Table~\ref{tab:overall-results}. \emph{Arm} is the robot; \emph{Obj.} is the objects that move during the clip (non-robot regions segmented in the first frame with SAM~2 and tracked through the clip). PSNR is computed inside each region and averaged over the 16 predicted frames; SCS is structural consistency of the same regions.
}
\label{tab:region-results}
\small
\setlength{\tabcolsep}{4pt}
\resizebox{\textwidth}{!}{%
\begin{tabular}{@{}llllrrrrrrrr@{}}
\toprule
& & & & \multicolumn{4}{c}{Validation: unseen trajectories} & \multicolumn{4}{c}{OOD test} \\
\cmidrule(lr){5-8} \cmidrule(lr){9-12}
Dataset & Method & Input & Trained on & \(\mathrm{PSNR}_{\mathrm{arm}}\) & \(\mathrm{PSNR}_{\mathrm{obj}}\) & \(\mathrm{SCS}_{\mathrm{arm}}\) & \(\mathrm{SCS}_{\mathrm{obj}}\) & \(\mathrm{PSNR}_{\mathrm{arm}}\) & \(\mathrm{PSNR}_{\mathrm{obj}}\) & \(\mathrm{SCS}_{\mathrm{arm}}\) & \(\mathrm{SCS}_{\mathrm{obj}}\) \\
\midrule
\droid{} & Action-only & Actions & This dataset & 18.74 & 27.26 & 0.808 & 0.839 & 17.64 & 24.67 & 0.712 & 0.746 \\
 & OSCAR (fine-tuned) & Skeleton & This dataset & 16.54 & 22.92 & 0.778 & 0.801 & 16.05 & 22.08 & 0.712 & 0.728 \\
 & \method{} & Render + actions & This dataset & 20.88 & \underline{28.63} & 0.843 & 0.867 & \underline{19.33} & \underline{26.50} & 0.773 & \textbf{0.805} \\
 & \methodthree{} & Render + actions & 3 robot datasets & \underline{20.89} & 28.56 & \underline{0.846} & \underline{0.869} & 19.23 & 26.41 & \underline{0.777} & \underline{0.800} \\
 & \methodeight{} & Render + actions & All 8 datasets & \textbf{21.59} & \textbf{29.25} & \textbf{0.853} & \textbf{0.876} & \textbf{19.57} & \textbf{26.54} & \textbf{0.795} & \textbf{0.805} \\
\midrule
\bridge{} & Action-only & Actions & This dataset & 21.15 & 27.65 & 0.776 & 0.894 & 17.74 & 21.50 & 0.591 & 0.778 \\
 & OSCAR (fine-tuned) & Skeleton & This dataset & 18.25 & 22.78 & 0.702 & 0.840 & 17.50 & 20.62 & 0.591 & 0.788 \\
 & \method{} & Render + actions & This dataset & \underline{21.52} & \textbf{28.04} & 0.787 & \textbf{0.901} & 18.72 & 23.64 & \underline{0.689} & \underline{0.838} \\
 & \methodthree{} & Render + actions & 3 robot datasets & 21.49 & 27.84 & \underline{0.791} & \underline{0.900} & \textbf{19.05} & \textbf{24.93} & \textbf{0.692} & \textbf{0.843} \\
 & \methodeight{} & Render + actions & All 8 datasets & \textbf{21.57} & \underline{27.96} & \textbf{0.793} & \textbf{0.901} & \underline{19.04} & \underline{24.85} & 0.685 & 0.831 \\
\midrule
\robomind{} & Action-only & Actions & This dataset & 25.64 & 29.16 & 0.909 & 0.866 & 25.03 & 28.70 & 0.896 & 0.826 \\
 & OSCAR (fine-tuned) & Skeleton & This dataset & 22.96 & 24.96 & 0.882 & 0.809 & 22.89 & 25.34 & 0.876 & 0.798 \\
 & \method{} & Render + actions & This dataset & 26.71 & \underline{29.53} & 0.916 & \underline{0.869} & 25.12 & 28.78 & 0.902 & 0.826 \\
 & \methodthree{} & Render + actions & 3 robot datasets & \underline{26.81} & 29.46 & \underline{0.917} & \underline{0.869} & \underline{25.30} & \underline{28.91} & \underline{0.903} & \textbf{0.844} \\
 & \methodeight{} & Render + actions & All 8 datasets & \textbf{27.23} & \textbf{29.60} & \textbf{0.919} & \textbf{0.873} & \textbf{25.64} & \textbf{29.07} & \textbf{0.905} & \underline{0.841} \\
\bottomrule
\end{tabular}%
}
\vspace{2pt}
\parbox{\textwidth}{\scriptsize Best in \textbf{bold}, second-best \underline{underlined}; ties at displayed precision share emphasis. Higher is better for all metrics. 200 clips per split, the same clips for every method.}
\end{table*}

\paragraph{Single-dataset experts versus \methodthree{}.}
Tables~\ref{tab:scaling-expert-psnr} and~\ref{tab:scaling-expert-scs_objects}
give the four-checkpoint curves for the separately trained action-only
experts and \methodthree{} (batch 256). The training budgets
are measured per dataset. The expert rows combine the corresponding
single-dataset model for each dataset.

\begin{table*}[t]
\centering
\caption{PSNR at every checkpoint in the expert-versus-shared-model scaling comparison. Action-only experts are trained separately on each dataset; \method{} is trained jointly on all three.}
\label{tab:scaling-expert-psnr}
\small
\setlength{\tabcolsep}{4pt}
\resizebox{\textwidth}{!}{%
\begin{tabular}{@{}lrrrrrrr@{}}
\toprule
& & \multicolumn{2}{c}{\droid{}} & \multicolumn{2}{c}{\bridge{}} & \multicolumn{2}{c}{\robomind{}} \\
\cmidrule(lr){3-4}\cmidrule(lr){5-6}\cmidrule(lr){7-8}
Method & \shortstack{Draws/dataset\\(M)} & Val & OOD & Val & OOD & Val & OOD \\
\midrule
Action-only expert & \(\sim 0.5\) & 19.194 & 18.058 & 22.136 & 21.119 & 25.199 & 25.848 \\
\methodthree{} & \(\sim 0.5\) & 20.815 & 19.505 & 24.253 & 22.314 & 28.560 & 28.634 \\
\midrule
Action-only expert & \(\sim 1.0\) & 19.583 & 18.110 & 23.374 & 21.279 & 26.449 & 27.016 \\
\methodthree{} & \(\sim 1.0\) & 22.768 & 20.545 & 24.935 & 22.907 & 29.386 & 29.589 \\
\midrule
Action-only expert & \(\sim 2.0\) & 20.963 & 18.982 & 24.805 & 20.552 & 27.501 & 28.039 \\
\methodthree{} & \(\sim 2.0\) & 23.664 & 20.965 & 25.809 & 22.472 & 30.387 & 30.172 \\
\midrule
Action-only expert & \(\sim 3.4\) & 22.228 & 19.767 & 25.869 & 20.555 & 29.629\rlap{$^\dagger$} & 29.629\rlap{$^\dagger$}  \\
\methodthree{} & \(\sim 3.4\)  & 24.220 & 21.147 & 26.458 & 22.438 & 30.964 & 30.436 \\
\bottomrule

\end{tabular}%

}
\vspace{2pt}
\parbox{\textwidth}{\scriptsize $^\dagger$Val.\ and OOD PSNR differ only beyond 3 decimals.}

\end{table*}

\begin{table*}[t]
\centering
\caption{SCS at every checkpoint in the expert-versus-shared-model scaling comparison. Action-only experts are trained separately on each dataset; \method{} is trained jointly on all three.}
\label{tab:scaling-expert-scs_objects}
\small
\setlength{\tabcolsep}{4pt}
\resizebox{\textwidth}{!}{%
\begin{tabular}{@{}lrrrrrrr@{}}
\toprule
& & \multicolumn{2}{c}{\droid{}} & \multicolumn{2}{c}{\bridge{}} & \multicolumn{2}{c}{\robomind{}} \\
\cmidrule(lr){3-4}\cmidrule(lr){5-6}\cmidrule(lr){7-8}
Method & \shortstack{Draws/dataset\\(M)} & Val & OOD & Val & OOD & Val & OOD \\
\midrule
Action-only expert & \(\sim 0.5\) & 0.868 & 0.826 & 0.901 & 0.875 & 0.897 & 0.899 \\
\methodthree{} & \(\sim 0.5\) & 0.914 & 0.876 & 0.927 & 0.917 & 0.927 & 0.926 \\
\midrule
Action-only expert & \(\sim 1.0\) & 0.877 & 0.826 & 0.916 & 0.883 & 0.911 & 0.909 \\
\methodthree{} & \(\sim 1.0\) & 0.926 & 0.888 & 0.936 & 0.915 & 0.933 & 0.931 \\
\midrule
Action-only expert &\(\sim 2.0\) & 0.905 & 0.852 & 0.933 & 0.868 & 0.919 & 0.919 \\
\methodthree{} & \(\sim 2.0\) & 0.933 & 0.896 & 0.942 & 0.907 & 0.937 & 0.932 \\
\midrule
Action-only expert & \(\sim 3.4\) & 0.921 & 0.869 & 0.943 & 0.863 & 0.937 & 0.929 \\
\methodthree{} & \(\sim 3.4\) & 0.940 & 0.897 & 0.948 & 0.904 & 0.940 & 0.937 \\
\bottomrule
\end{tabular}%
}
\end{table*}

\paragraph{Shared action-only versus \methodthree{} (batch 64).}
Tables~\ref{tab:scaling-shared-psnr} and~\ref{tab:scaling-shared-scs_objects}
give the three-checkpoint curves for the matched batch-64 shared models.
Here, training budgets count total draws across all three datasets.
The two models use the same training exposure and optimization schedule
at each checkpoint. This \methodthree{} (batch 64) run differs from the batch-256 \methodthree{}
run in the expert comparison above.

\begin{table*}[t]
\centering
\caption{PSNR at every checkpoint in the matched shared-model scaling comparison. Both models are trained jointly on all three datasets.}
\label{tab:scaling-shared-psnr}
\small
\setlength{\tabcolsep}{4pt}
\resizebox{\textwidth}{!}{%
\begin{tabular}{@{}lrrrrrrr@{}}
\toprule
& & \multicolumn{2}{c}{\droid{}} & \multicolumn{2}{c}{\bridge{}} & \multicolumn{2}{c}{\robomind{}} \\
\cmidrule(lr){3-4}\cmidrule(lr){5-6}\cmidrule(lr){7-8}
Method & \shortstack{Total draws\\(M)} & Val & OOD & Val & OOD & Val & OOD \\
\midrule
Action-only (multi) &  \(\sim 1.2\) & 19.126 & 18.029 & 21.703 & 20.762 & 24.520 & 24.958 \\
\methodthree{} (batch 64) &  \(\sim 1.2\)  & 22.070 & 20.251 & 24.098 & 22.781 & 28.504 & 29.082 \\
\midrule
Action-only (multi) &  \(\sim 2.4\)  & 20.381 & 18.762 & 23.196 & 21.722 & 26.575 & 26.965 \\
\methodthree{} (batch 64) & \(\sim 2.4\) & 22.832 & 20.697 & 24.968 & 23.156 & 29.382 & 29.670 \\
\midrule
Action-only (multi) & \(\sim 3.4\) & 20.959 & 19.036 & 24.070 & 21.873 & 27.564 & 28.200 \\
\methodthree{} (batch 64) & \(\sim 3.4\) & 23.188 & 20.862 & 25.431 & 22.726 & 29.937 & 29.818 \\
\bottomrule
\end{tabular}%
}
\end{table*}

\begin{table*}[t]
\centering
\caption{SCS at every checkpoint in the matched shared-model scaling comparison. Both models are trained jointly on all three datasets.}
\label{tab:scaling-shared-scs_objects}
\small
\setlength{\tabcolsep}{4pt}
\resizebox{\textwidth}{!}{%
\begin{tabular}{@{}lrrrrrrr@{}}
\toprule
& & \multicolumn{2}{c}{\droid{}} & \multicolumn{2}{c}{\bridge{}} & \multicolumn{2}{c}{\robomind{}} \\
\cmidrule(lr){3-4}\cmidrule(lr){5-6}\cmidrule(lr){7-8}
Method & \shortstack{Total draws\\(M)} & Val & OOD & Val & OOD & Val & OOD \\
\midrule
Action-only (multi) & \(\sim 1.2\) & 0.864 & 0.824 & 0.892 & 0.859 & 0.886 & 0.886 \\
\methodthree{} (batch 64) & \(\sim 1.2\) & 0.919 & 0.883 & 0.927 & 0.914 & 0.927 & 0.932 \\
\midrule
Action-only (multi) & \(\sim 2.4\)   & 0.893 & 0.839 & 0.914 & 0.886 & 0.910 & 0.910 \\
\methodthree{} (batch 64) & \(\sim 2.4\)   & 0.927 & 0.892 & 0.936 & 0.919 & 0.930 & 0.934 \\
\midrule
Action-only (multi) & \(\sim 3.4\) & 0.901 & 0.846 & 0.925 & 0.890 & 0.920 & 0.916 \\
\methodthree{} (batch 64) & \(\sim 3.4\) & 0.930 & 0.896 & 0.940 & 0.912 & 0.933 & 0.930 \\
\bottomrule
\end{tabular}%
}
\end{table*}

\end{document}